\documentclass[default,iicol]{sn-jnl}
\usepackage{amsmath}
\usepackage{diagbox}
\usepackage{graphics}
\usepackage{epsfig}
\usepackage{threeparttable}
\usepackage{xspace}
\usepackage{siunitx}
\usepackage{subfigure}
\usepackage{graphicx}
\usepackage{amssymb}
\usepackage{threeparttable}
\usepackage{color}
\usepackage[normalem]{ulem}
\usepackage{multirow}
\usepackage{float}
\usepackage{amsfonts}

\usepackage{bm}
\usepackage{array}
\PassOptionsToPackage{table}{xcolor} 
\definecolor{mygray}{RGB}{200,200,200}
\definecolor{reda}{RGB}{255,0,0}
\definecolor{redb}{RGB}{217,148,143}
\definecolor{myyellow}{RGB}{190,144,0}
\definecolor{mygreen}{RGB}{0,136,51}
\definecolor{myblue}{RGB}{0,102,204}
\usepackage{colortbl}
\newcommand{\tabincell}[2]{\begin{tabular}{@{}#1@{}}#2\end{tabular}}
\usepackage{pifont}

\newcommand{\yes}{\text{\ding{51}}}
\newcommand{\no}{\text{\ding{55}}}

\makeatletter
\newcommand{\thickhline}{%
    \noalign {\ifnum 0=`}\fi \hrule height 1pt
    \futurelet \reserved@a \@xhline
}
\makeatother

\jyear{2021}%

\theoremstyle{thmstyleone}%
\theoremstyle{thmstyletwo}%

\theoremstyle{thmstylethree}%

\begin{document}

\title[Article Title]{Towards Diverse Binary Segmentation via  A Simple yet General Gated Network}

\author[1]{\fnm{Xiaoqi} \sur{Zhao}}\email{zxq@mail.dlut.edu.cn}
\author[1]{\fnm{Youwei} \sur{Pang}}\email{lartpang@mail.dlut.edu.cn}
\author*[1]{\fnm{Lihe} \sur{Zhang}}\email{zhanglihe@dlut.edu.cn}
\author[1,2]{\fnm{Huchuan} \sur{Lu}}\email{lhchuan@dlut.edu.cn}
\author[3,4]{\fnm{Lei} \sur{Zhang}}\email{cslzhang@comp.polyu.edu.hk}

\affil*[1]{\orgdiv{Dalian University of Technology, China}}

\affil[2]{\orgdiv{Peng Cheng Laboratory, China}}

\affil[3]{\orgdiv{Dept. of Computing, The Hong Kong Polytechnic University, China}}
\affil[4]{\orgdiv{OPPO Research, China}}


\abstract{In many binary segmentation tasks, most CNNs-based methods use a U-shape encoder-decoder network as their basic structure. They ignore two key problems when the encoder exchanges information with the decoder: one is the lack of interference control mechanism between them, the other is without considering the disparity of the contributions from different encoder levels. In this work, we propose a simple yet general gated network (GateNet) to tackle them all at once. With the help of multi-level gate units, the valuable context information from the encoder can be selectively transmitted to the decoder. In addition, we design a gated dual branch structure to build the cooperation among the features of different levels and improve the discrimination ability of the network. Furthermore, we introduce a “Fold” operation to improve the atrous convolution and form a novel folded atrous convolution, which can be flexibly embedded in ASPP or DenseASPP to accurately localize foreground objects of various scales. GateNet can be easily generalized to many binary segmentation tasks, including general and specific object segmentation and multi-modal segmentation. Without bells and whistles, our network consistently performs favorably against the state-of-the-art methods under $10$ metrics on $33$ datasets of $10$ binary segmentation tasks.}




\maketitle
\section{Introduction}\label{sec:introduction}
\begin{figure}[!htb]
  	\centering
  	\includegraphics[width=0.9\linewidth]{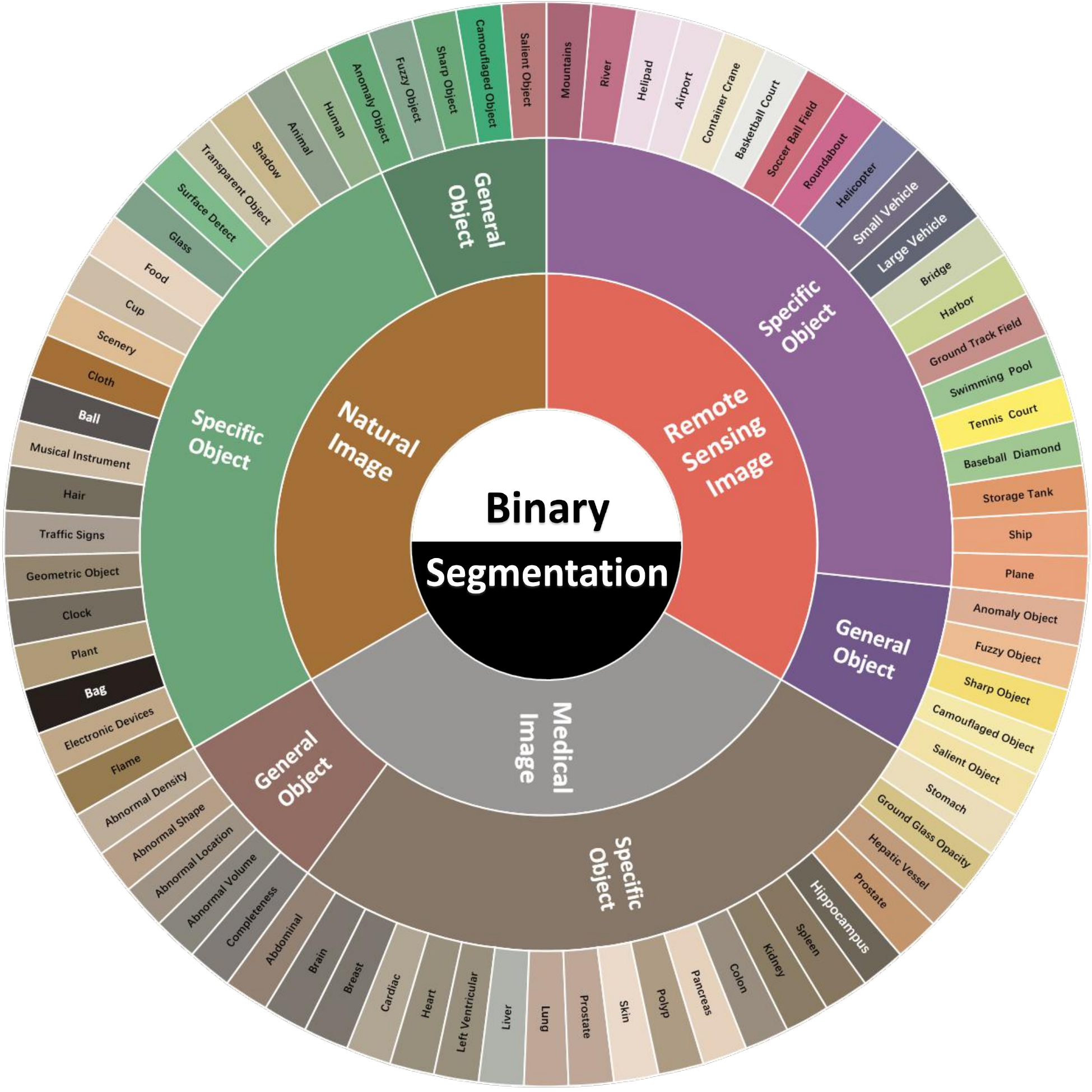}
  	\caption{Some meaningful binary segmentation tasks.
  	}
  	\label{fig:diverse_bs}
  \end{figure}

Image segmentation
 is the process of dividing a digital image into segments that simplify and/or change the representation of the image to something more meaningful and easier to analyze.
 From the perspective of pixel-level classification, image segmentation can be specifically divided into binary segmentation, semantic segmentation, instance segmentation and panoramic segmentation. Compared with the others, segmentation problems considered in binary segmentation are more pure and focused, that is, accurately distinguishing the foreground and background. 
 As shown in Fig.~\ref{fig:diverse_bs}, binary segmentation has a wide range of applications in military, industrial, medical, etc. 
   
Rich foreground definitions prompt binary segmentation with numerous branches, such as salient object detection, camouflaged object detection, shadow detection and transparent object detection. 
In recent years, with the development of deep learning, there are many effective methods proposed and achieve good performance.
Although each branch of binary segmentation  is thriving and show a gratifying state, almost all methods focus on researching single one branch and ignore cross-branch comparison in experiments and techniques. 
As we know, each branch belongs to the binary segmentation trunk because they have a same mathematical definition. They face many same challenges in segmentation techniques.
However, these task branches have become more and more independent, which will impede the development of the entire binary segmentation field.
To this end, it is urgent to provide a general method for diverse binary segmentation branches.

There are three challenges in accurate binary segmentation: 
\textbf{Firstly}, most methods~\cite{BASNet,MINet,DANet_RGBDSOD,SINet_COD,Amulet,DGRL,BMPM,CPD,DHS,DSS} tend to adopt U-shape~\cite{Unet,FPN} as the baseline and then combine multi-level features in either the encoder~\cite{Amulet,DGRL,BMPM,BASNet,CPD} or the decoder~\cite{DHS,DSS,PAGRN,CPD,UNet++} to gradually reconstruct the high-resolution feature maps. In each convolutional block, they separately formulate the relationships of internal features during forward update. It is well known that the high-quality segmentation predicted in the decoder relies heavily on the effective features provided by the encoder. 
Nevertheless, these methods directly use an all-pass skip-layer structure to concatenate the features of the encoder to the decoder in the isolated~
\cite{Unet,BMPM,SINet_COD,DANet_RGBDSOD,PraNet_Polyp,VST} or nested~\cite{U2Net,UNet++,MINet,CPFP_RGBDSOD,SINet_COD,MSNet_Polyp} manner. 
The effectiveness of feature aggregation at different levels is not quantified. This not only introduces misleading context information into the decoder but also causes that the typically useful features can not be adequately utilized. In cognitive science, Yang \textit{et al.}~\cite{NM} show that inhibitory neurons play an important role in how the human brain chooses to process the most important information from all the information presented to us. And inhibitory neurons ensure that humans respond appropriately to external stimuli by inhibiting other neurons and balancing excitatory neurons that stimulate neuronal activity. Inspired by this work, we think that it is necessary to set up an information screening unit between each pair of encoder and decoder blocks in binary prediction. It will help distinguish the most task-aware features of foreground regions and suppress background interference.
\textbf{Secondly}, due to the limited receptive field, a single-scale convolutional kernel is difficult to capture context information of size-varying objects. This motivates many efforts~\cite{R3Net,BMPM,UCNet_RGBDSOD,S2MA_RGBDSOD,DMRA_RGBDSOD,CoNet_RGBDSOD,BBSNet_RGBDSOD,JLDCF_RGBDSOD,PoolNet} to investigate multi-scale feature extraction. These methods directly equip an atrous spatial pyramid pooling module~\cite{ASPP} (ASPP) or DenseASPP~\cite{DenseASPP} in their networks. However, when using a convolution with a large dilation rate, the information under the kernel seriously lacks correlation due to inserting too many zeros. This may be detrimental to the discrimination of subtle image structures.
\textbf{Thirdly}, both body and boundary of the foreground need to accurately segmented. Most existing models either use progressive decoder~\cite{PraNet_Polyp,MSNet_Polyp,PDNet_Mirror,GDNet_Glass,ADNet_Shadow,DENets_DBD,SG_DBD,UGTR_COD} or parallel decoder~\cite{R3Net,PFA,SCRN,BANet,CSNet,CoNet_RGBDSOD,DeFusionNet_DBD,R2MRF_DBD}. The progressive structure begins with the top layer and gradually utilizes the output of the higher layer as prior knowledge to fuse the encoder features. This mechanism is not conducive to the recovery of details because the high-level features lack fine information. While the parallel structure easily results in inaccurate localization of objects since the low-level features without semantic information directly interfere with the capture of global structure cues. 

In this paper, we propose a simple yet general gated network (GateNet) for binary segmentation. Firstly, based on the feature pyramid network (FPN), we construct multi-level gate units to combine the features from the decoder and the encoder. We use convolution operation and nonlinear functions to calculate the correlations among features and assign gate values to different blocks. In this process, a partnership is established between different blocks by using weight distribution and the decoder can obtain more efficient information from the encoder and pay more attention to the target-aware regions. Secondly, we construct a folded atrous spatial pyramid pooling (Fold-ASPP) module to gather multi-scale high-level foreground cues. With the ``Fold'' operation, the atrous convolution is implemented on a group of local neighborhoods rather than a group of isolated sampling points, which can help generate more stable features and more adequately depict finer structure. Thirdly, we design a mix feature aggregation decoder that a parallel branch by concatenating the output of the progressive branch and the features of the gated encoder, so that the residual information complementary to the progressive branch is supplemented to generate the final prediction. 

Our main contributions can be summarized as follows.
\begin{itemize}
	\item We provide a unified perspective of binary segmentation by comprehensively analyzing many binary segmentation tasks.
	\item We propose a simple gated network to adaptively control the amount of information that flows into the decoder from each encoder block. With multi-level gate units, the network can balance the contribution of each encoder block to the the decoder block and suppress the features of background regions.
	\item We design a novel folded atrous convolution that can transfer existing multi-scale modules into our Fold style and enjoy more effective feature representation.
	\item We build a dual branch architecture. They form a residual structure, complement each other through the gated processing and generate better results. 
	\item We construct both single-stream and two-stream gated networks to adapt the binary segmentation required one or two input sources. 
	\item Extensive comparisons with $42$ state-of-the-art methods on $33$ challenging datasets of $10$ binary segmentation tasks, including RGB, RGB-D and optical remote sensing image salient object detection, camouflaged object detection, defocus blur detection, shadow detection, transparent detection, glass detection, mirror detection and polyp segmentation in medical images,  show that our method performs much better than other competitors under $10$ metrics and possess strong generalization. Hence, it can be seen a strong baseline for the binary segmentation field.
\end{itemize}

\textit{Compared with the ECCV version~\cite{GateNet} (Oral) of this work, the following extensions are made. 
\textbf{\uppercase\expandafter{\romannumeral1})} We conduct a survey on the field of binary segmentation, covering $10$ popular branches and $141$ fully supervised methods, evaluation metrics and datasets.
\textbf{\uppercase\expandafter{\romannumeral2})} Deeper theoretical explanations of the proposed gate unit design are added and we improve the previous gate unit into a stronger version.
\textbf{\uppercase\expandafter{\romannumeral3})} Based on the overall structure of the original single-source input GateNet, we expand a two-stream version of GateNet suitable for two-source input tasks. Meanwhile, our multi-level gate units can further carry forward the spirit of suppress and balance between different sources.
\textbf{\uppercase\expandafter{\romannumeral4})} We report much more extensive experimental results that demonstrate the superiority of both single-stream and dual-stream GateNet in $10$ popular binary segmentation tasks.
\textbf{\uppercase\expandafter{\romannumeral5})} We further provide more implementation details and thorough ablation studies at qualitative and quantitative aspects.
\textbf{\uppercase\expandafter{\romannumeral6})} We perform in-depth analyses and discussion for our gate unit.}

 \section{Retrospect}\label{sec:2}

 \subsection{Diverse Binary Segmentation Tasks (DBS)}\label{sec:2.1}
As shown in Fig.~\ref{fig:diverse_bs}, there are many kinds of binary segmentation  in real life. 
We select $10$ currently well-developed and hot tasks that cover the requirements of general and specific object segmentation in natural images, remote sensing images, and medical images.
According to the rapid development of deep learning technology, we only review the research progress in recent five years in order to provide the latest and comprehensive content.
\subsubsection{General Object Segmentation}
\noindent$\bullet$~\textbf{RGB Salient Object Detection.}
Salient object detection (SOD) aims to segment the most salient (judged by different consciousnesses) regions or objects in various scenes with or without the engineered cues, such as visual cues, geodesic cues, temporal cues, and human attention cues. Usually, it is adopted as a pre-processing step in many computer vision applications, such as scene classification~\cite{classification}, person re-identification~\cite{Reid} and image captioning~\cite{Imagecaption}. 

\noindent$\bullet$~\textbf{RGB-D Salient Object Detection.}
Although RGB SOD methods can achieve satisfactory performance in segmenting visually salient objects, some complex scenarios are still open to be resolved. For example, salient objects share similar appearance to the background or the other similar trivial objects. In recent years, various depth-assisted salient object detection (RGB-D SOD) methods~\cite{PCA_RGBDSOD,CPFP_RGBDSOD,DMRA_RGBDSOD} have been proposed, in which absorbing geodesic cues from the depth map is the hardcore.

\noindent$\bullet$~\textbf{Remote Sensing Image Salient Object Detection.}
Remote sensing images (RSIs) are usually captured by sensors on anairplane as an aerial view under various viewing angle conditions. 
Although recent decades have witnessed the remarkable success of SOD for natural scene images, there is only a limited amount of researches focusing on SOD for optical remote sensing images (RSIs). 
Typically, optical RSIs cover a wide scope with complicated background and diverse noise interference. 

\noindent$\bullet$~\textbf{Camouflaged Object Detection.}
The study of camouflage has a long history in biology, and more details can be found in~\cite{camouflage}. In the field of computer vision, research on camouflaged object detection (COD) is often associated with salient object detection task. In general, saliency models are designed for finding visually salient objects. They are not suitable for finding hidden objects. The local features of the camouflaged object are usually slightly different from the surrounding background. Recently, Fan \textit{et al.}~\cite{SINet_COD} make some attempts towards this direction. They first build the largest COD dataset, which contains $10,000$ images covering $78$ camouflaged object categories.

\subsubsection{Specific Object Segmentation}
\noindent$\bullet$~\textbf{Defocus Blur Detection.}
Defocus blur is a blurring degradation caused by defocusing and inappropriate depth of focus. 
Defocus blur is a common phenomenon in real life when the scene is beyond the focal distance of the camera.
Defocus blur detection can be potentially used to many vision tasks (\textit{e.g.}, autofocus, depth estimation).

\noindent$\bullet$~\textbf{Shadow Detection.}
Shadow is the light effect caused by surface occlusion and are almost ubiquitous in our daily lives. 
One one hand, shadow can be used as auxiliary information due to rich depth and geometry visual cues.
On the other hand, some important details of the object may be hidden when overlapping with shadows.
Hence, shadow detection is important for shadow removal~\cite{DAS_shadow_removal}, scene geometry~\cite{Shadow_scene_geometry} and camera parameters~\cite{camera_parameters_shadow}.

\noindent$\bullet$~\textbf{Glass and Transparent Detection.}
Transparent objects are widely present in the real world, such as glass, vitrines, and bottles. 
And most of them appear in indoor scenes, especially glass-like objects with brittle and smooth properties.
Smart robot operates tasks in living rooms or offices, it needs to avoid fragile objects.
Hence, it is essential for vision systems to be able to detect and segment transparent objects from input images.

\noindent$\bullet$~\textbf{Mirror Detection.}
As a very important object in daily life, mirrors are ubiquitous. They can not only reflect light, but also present a similar mirror image of surrounding objects or scenes. 
As a result, once the computer vision system or robot encounters a scene with a mirror, the performance will drop significantly. To avoid this problem, it requires these systems to be able to detect and segment mirrors.

\noindent$\bullet$~\textbf{Polyp Detection.}
According to GLOBOCAN 2020 data, colorectal cancer is the third most common cancer worldwide and the second most common cause of death. 
It usually begins as small, noncancerous (benign) clumps of cells called polyps that form on the inside of the colon.  
Over time some of these polyps can become colon cancers. 
Therefore, the best way of preventing colon cancer is to identify and remove polyps before they turn into cancer.

\begin{table*}
	\centering
	\caption{
		{\textbf{Summary of essential characteristics for reviewed fully-supervised binary segmentation methods.}  
			The superscript \textbf{``$*$''} in the fifth column (code link) regards this repository does not provide pre-trained weights for re-evaluating performance publicly and \textbf{``N/A''} represents that the code is not available. 
			\textbf{STL} is single task learning 
			and \textbf{MTL} is multi-task learning.}
	}
	\vspace{-5pt}
	\label{table:Survey_DBS_methods_1}
	\begin{threeparttable}
		\resizebox{1\textwidth}{!}{

			 \renewcommand\tabcolsep{1.pt} 
			\renewcommand\arraystretch{0.94}
			
\begin{tabular}{|c|c|r|c|c||c|c|c|c|}
    \hline\thickhline
    \rowcolor{mygray}
    
    No.&Year&Methods&Publication &\tabincell{c}{Code\\Link}  &Backbone
    &\tabincell{c}{Learning\\Paradigm} &Training Dataset &\#Training  \\
    \hline
    \hline
    \multicolumn{9}{|c|}{\multirow{2}{*}{\large\textbf{RGB Salient Object Detection}}}
    \\
    \multicolumn{9}{|c|}{}		
    \\
    
    \hline
    \hline
    1&	\multirow{7}{*}{\rotatebox{90}{2018}} &R3Net\cite{R3Net} &IJCAI &\href{https://github.com/zijundeng/R3Net}{Pytorch} &ResNeXt-101~\cite{ResNeXt} &STL  &MSRA10K~\cite{MSRA10K} &10,000\\
    2&	 &SFCN\cite{SFCN} &IJCAI & \href{https://github.com/Pchank/caffe-sal/blob/master/IIAU2018.md}{Caffe} &VGG-16~\cite{VGG} &STL  &MSRA10K~\cite{MSRA10K} &10,000\\
    3&	 &BMPM\cite{BMPM} &CVPR &\href{https://github.com/zhangludl/A-bi-directional-message-passing-model-for-salient-object-detection}{TensorFlow} &VGG-16~\cite{VGG} &STL  &DUTS~\cite{DUTS} &10,553\\
    4&	 &PiCANet\cite{PICANet} &CVPR &\href{https://github.com/Ugness/PiCANet-Implementation}{Pytorch} &ResNet-50~\cite{Resnet}/VGG-16~\cite{VGG} &STL  &DUTS~\cite{DUTS} &10,553 \\
    5&	 &PAGRN\cite{PAGRN} &CVPR &N/A &VGG-19~\cite{VGG} &STL  &DUTS~\cite{DUTS} &10,553 \\
\hline
    
\end{tabular}%
			
		}
	\end{threeparttable}
\end{table*}

\subsection{Fully Supervised Binary Segmentation Models}\label{sec:2.4}
We first formulate the image-based binary segmentation problem. 
Formally, let \bm{$\mathcal{X}$} and \bm{$\mathcal{Y}$} denote the input space and output segmentation space, respectively. 
Fully supervised learning-based models generally seek to learn an \textit{ideal} image-to-segment mapping $f^{*\!}:\bm{\mathcal{X}}\mapsto\bm{\mathcal{Y}}$ through directly utilizing ground truth masks as supervision signal. 

In Tab.~\ref{table:Survey_DBS_methods_1} and Tab.~\ref{table:Survey_DBS_methods_3}, we categorize recent fully supervised  models.
Through the analyses of $141$ methods in $10$ branches, we summarize some instructive findings:
\textbf{\uppercase\expandafter{\romannumeral1})} Single task learning (STL) is still the main learning paradigm in binary segmentation. Compared with STL, the proportion of MTL-based methods is only $34/141$ and they finish MTL via cooperating with boundary prediction or depth estimation usually. It is worth noting that MTL-based RGB SOD methods have reached the $6/9$ scale in 2021. We believe that the potential of MTL is huge, and more effective and richer strategies will emerge in the future under the continuous efforts of researchers.
\textbf{\uppercase\expandafter{\romannumeral2})} Our GateNet is the only one mixes both progress and parallel structures among $141$ methods, thereby enjoying the advantages of both. 
Most methods still adopt the  single progressive style.
\textbf{\uppercase\expandafter{\romannumeral3})} Conditional random field (CRF) gradually disappear in many models. Only some shadow and  mirror detection methods adopt the CRF as post-processing.
\textbf{\uppercase\expandafter{\romannumeral4})} Deep supervision becomes a popular supervision approach. 
$73/141$ methods build the network with side outs to perform deep supervision. On one hand, deep supervision~\cite{Deepsupervision} is originally designed to speed up network convergence. On the other hand, it may bring extra performance gain for most models.  
\textbf{\uppercase\expandafter{\romannumeral5})} Targeted loss function is conducive to performance improvement. $72/141$ models directly adopt previous or re-design a new targeted loss, such as the hybrid loss~\cite{BASNet}, consistency-enhanced loss~\cite{MINet}, pixel position aware loss~\cite{F3Net}, etc. It is clear that there is increasing competition in the research of targeted loss.
\subsection{Multi-scale Feature Extraction}\label{sec:2.5}
The multi-scale paradigm is mainly inspired by the scale-space theory that has been widely validated as an effective and theoretically sound framework. 
It is well suited for addressing naturally occurring scale variations.
Common forms in the field of computer vision mainly include the image pyramid~\cite{ImagePyramid} and the feature pyramid~\cite{Unet,FPN}.
Although the former has shown good performance, its application is limited by high computational and latency costs associated with the multi-input parallel processing paradigm, which makes it gradually give way to the more efficient latter in the era of deep learning.
The feature pyramid can be roughly divided into two categories according to the form, namely, the inter-layer pyramid and the intra-layer pyramid.
The former is based on features with different scales extracted by the feature encoder, such as the U-shape~\cite{Unet,FPN,U2Net,LargeKernel,PAGRN,MINet} architecture. In this way, the internal cross-layer information propagation path progressively integrates semantic context and texture details from diverse scale representations.
The intra-layer pyramid~\cite{PPM,ASPP,DenseASPP,BMPM,DANet_RGBDSOD} can enhance the diversity of semantic content by constructing the multi-path structure within a layer to obtain a rich combination of receptive fields.
Its good pluggability has also made it an important component in the architecture design of modern segmentation methods.
Recently, the atrous spatial pyramid pooling module (ASPP)~\cite{ASPP} and its variants~\cite{DenseASPP,DilatedConvolution-SDC,DilatedConvolution-ESDC,DilatedConvolution-C3Convolution,DilatedConvolution-ESPNet,DilatedConvolution-TKCN,DilatedConvolution-D3DNet,DANet_RGBDSOD}, which typify this structure, are widely applied in many segmentation tasks and networks. 
Some methods~\cite{BMPM,UCNet_RGBDSOD,PFNet_COD,BDRAR_Shadow,AFFPN_Shadow,RCARP_Glass} insert several ASPP modules into the encoder/decoder blocks of different levels, while some ones~\cite{R3Net,AFNet,DMRA_RGBDSOD,CoNet_RGBDSOD,Rank-Net_COD,EBLNet_Glass} install it on the highest-level encoder block.
As a basic component of ASPP, atrous convolution has the advantage of enlarging the receptive field to obtain large-scale features without increasing the computational cost compared to the vanilla convolution.
Nevertheless, the repeated stride and pooling operations already make the top-layer features lose much fine information. 
With the increase of atrous rate, the correlation of sampling points further degrades, which leads to difficulties in capturing the changes of image details (\textit{e.g.}, lathy background regions between adjacent objects or spindly parts of objects). 
In this work, we  propose a folded atrous convolution to alleviate these issues and achieve a \textsl{local-in-local} effect. 
The folded atrous convolution can seamlessly replace the original atrous convolution in ASPP and other variants (\textit{e.g.}, DenseASPP~\cite{DenseASPP}, PAFE~\cite{DANet_RGBDSOD}), thus significantly improving performance.

\subsection{Gated Mechanisms}
\label{sec:2.6}
The gated mechanism plays an important role in controlling the flow of information and is widely used in the long short term memory (LSTM).
In~\cite{gatedlabel}, the gate unit combines two consecutive feature maps of different resolutions from the encoder to generate rich contextual information.
And the gated mechanism is also integrated into the block feedback mechanism to bridge multiple iterations in the recurrent architecture~\cite{Gate-RIGNet}.
Zhang \textit{et al.}~\cite{BMPM} adopt gate function to control the message passing when combining feature maps at all levels of the encoder.
Chen \textit{et al.}~\cite{DPANet_RGBDSOD}  propose a gate function controller to focus on regulating the fusion rate of the cross-modal information.
Zhang \textit{et al.}~\cite{FRDT_RGBDSOD} utilize the gated select fusion module to selectively process the useful information from two modal features in low-levels.
Due to the ability to filter information, the gated mechanism can also be seen as a special kind of attention mechanism.
Wang \textit{et al.}~\cite{PASE} exploit the pyramid attention module to enhance saliency representations for each layer in the encoder and enlarge the receptive field.
Chen \textit{et al.}~\cite{GCPANet} propose a head attention  module  to  reduce  information  redundancy and enhance the top layers features by leveraging both spatial and channel-wise attention.
Zhang \textit{et al.}~\cite{PAGRN} apply both spatial and channel attention to each layer of the decoder.
Liu \textit{et al.}~\cite{S2MA_RGBDSOD} construct both self-attention and mutual-attention in a non-local~\cite{Nonlocal} style for extracting the complementary information between the different modalities. 
Zhu \textit{et al.}~\cite{BDRAR_Shadow} design the recurrent attention residual module to combine and process spatial contexts in two adjacent CNN layers.
Zhang \textit{et al.}~\cite{ACSNet_Polyp} apply both the local context attention and SE-like~\cite{SENet} channel-wise attention for context selection.
Taehun \textit{et al.}~\cite{UACANet_Polyp} propose the uncertainty augmented context attention module to  incorporate uncertain area for rich semantic feature extraction.
More description about attention-based methods can found in Tab.~\ref{table:Survey_DBS_methods_3}.
In general, the above methods unilaterally consider the information interaction between different layers or intra-layer in the encoder or decoder.
We integrate the features from the encoder and the decoder to formulate gate function, which has the function of \textbf{\textsl{block-wise attention}} and models the overall distribution of all blocks in the network from the global perspective. 
However, while previous methods utilize the block-specific features to compute dense attention weights for the corresponding block, they directly feed the encoder features into the decoder and do not consider their mutual interference. 
Our proposed gate unit can naturally balance their contributions, thereby suppressing the response of the encoder to background regions. 
Experimental results in Fig.~\ref{Fig:gatevalue} and Fig.~\ref{fig:Gate_suppress_visual_results} intuitively demonstrate the effect of multi-level gate units on the above two aspects, respectively.

\begin{figure*}
	\includegraphics[width=\linewidth]{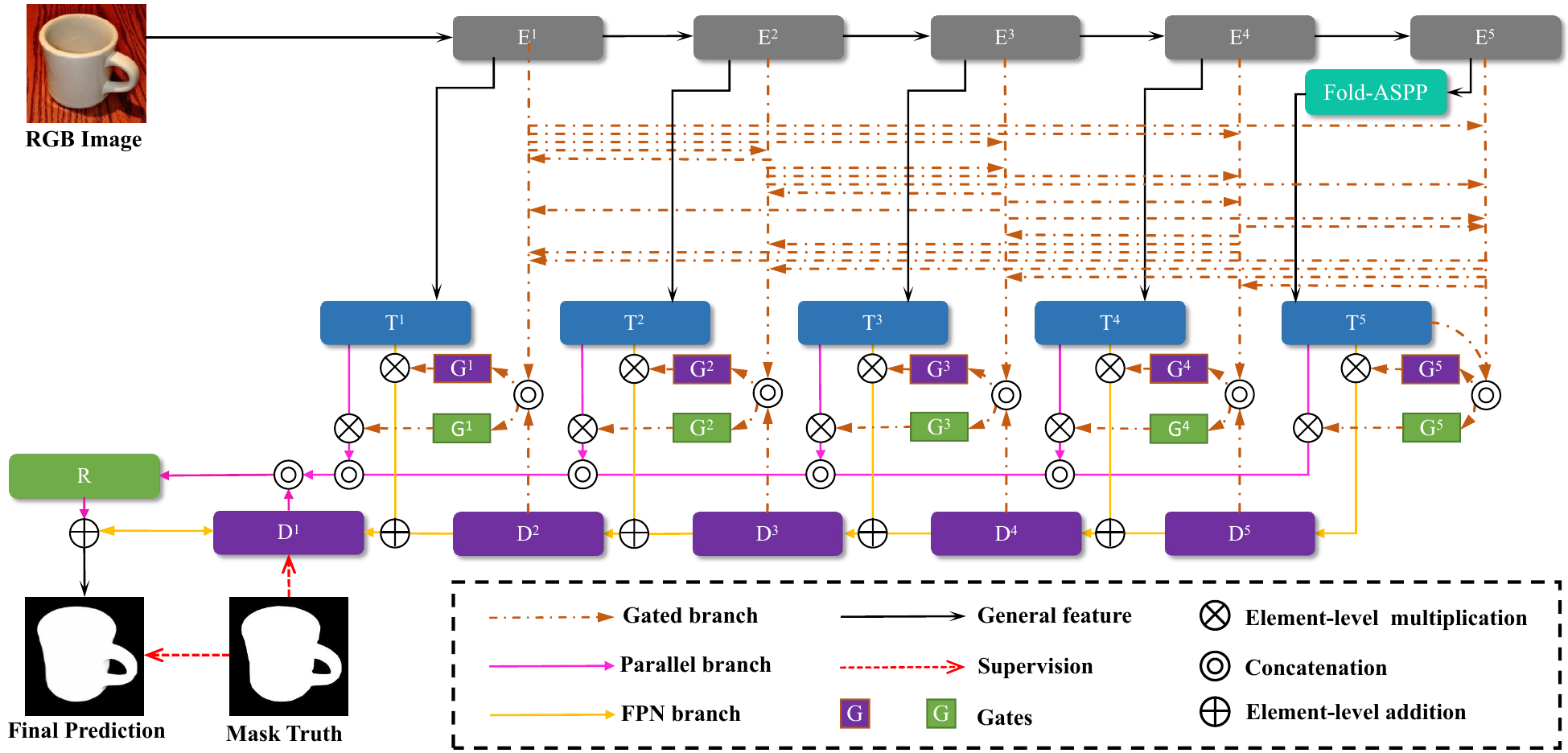}\\ 
	      \vspace{-6mm}
	\centering
	\caption{Overall architecture of the gated network. It consists of   five encoder blocks ($\mathbf{E}^1 \sim \mathbf{E}^5$), five transition layers ($\mathbf{T}^1 \sim \mathbf{T}^5$), five gate units ($\mathbf{G}^1 \sim \mathbf{G}^5$), five decoder blocks ($\mathbf{D}^1 \sim \mathbf{D}^5$) and the {Fold-ASPP} module. We employ twice supervision in this network. One acts at the end of the FPN branch ${D}^1$. The other is used to guide the fusion of the two branches.} 		
	\label{fig:GateNet}
\end{figure*}

\section{Proposed Method}
\label{sec:proposedmethod}
The gated network architecture is shown in Fig.~\ref{fig:GateNet}, in which encoder blocks, transition layers, decoder blocks and  gate units are denoted as $\mathbf{E}^i$, $\mathbf{T}^i$, $\mathbf{D}^i$ and $\mathbf{G}^i$, respectively ($i \in \left \{1, 2, 3, 4, 5 \right \}$ indexes different levels). Their output feature maps are denoted as $E^i$, $T^i$, $D^i$ and $G^i$, respectively. 
The final prediction is obtained by combining the FPN branch and the parallel branch. 
\subsection{Network Overview}\label{sec:Network}
\noindent\textbf{Encoder Network.} In our model, the encoder is based on a common pretrained backbone network, \textit{e.g.}, the VGG~\cite{VGG}, ResNet~\cite{Resnet} or ResNeXt~\cite{ResNeXt}. We take the VGG-16 network as an example, which contains thirteen Conv layers, five max-pooling layers and two fully connected layers. In order to fit saliency detection task, similar to most previous approaches~\cite{Amulet,DSS,PAGRN,BMPM}, we cast away all the fully-connected layers of the VGG-16 and remove the last pooling layer to retain details of last convolutional layer. 

\noindent\textbf{Decoder Network.} The decoder comprises three main components: i) the FPN branch, which continually fuses different level features from ${T}^1 \sim{T}^5$ by element-wise addition; ii) the parallel branch, which combines the saliency map of the FPN branch and the feature maps of transition layers by cross-channel concatenation (At the same time, multi-level gate units ($\mathbf{G}^1 \sim \mathbf{G}^5$) are inserted between the transition layer and the decoder layer); iii) the Fold-ASPP module, which improves the original atrous spatial pyramid pooling (ASPP) by using a ``Fold'' operation. It can take advantage of semantic features learned from ${E}^5$ to provide multi-scale information to the decoder.
\subsection{Gated Dual Branch}\label{sec:Gated Dual Branch}
\begin{figure}[t]
	\centering
	\includegraphics[width=\linewidth]{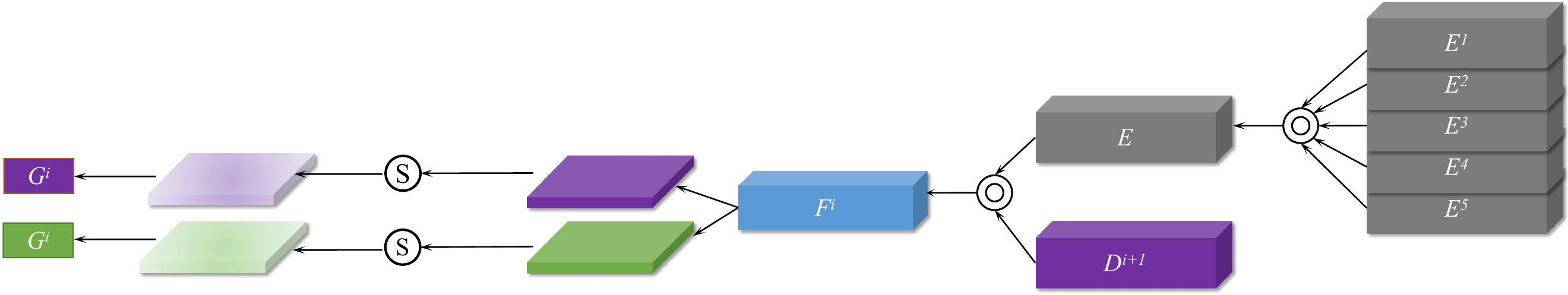}
	\caption{Detailed illustration of the gate unit. ${D}^{i+1}$ indicates feature maps of the previous decoder block. \textcircled{\scriptsize S} is sigmoid function.}\label{fig:Gate_Unit}
	\label{fig:GateUnit_v2}
\end{figure}
\begin{figure}[t]
	\centering
	\includegraphics[width=\linewidth]{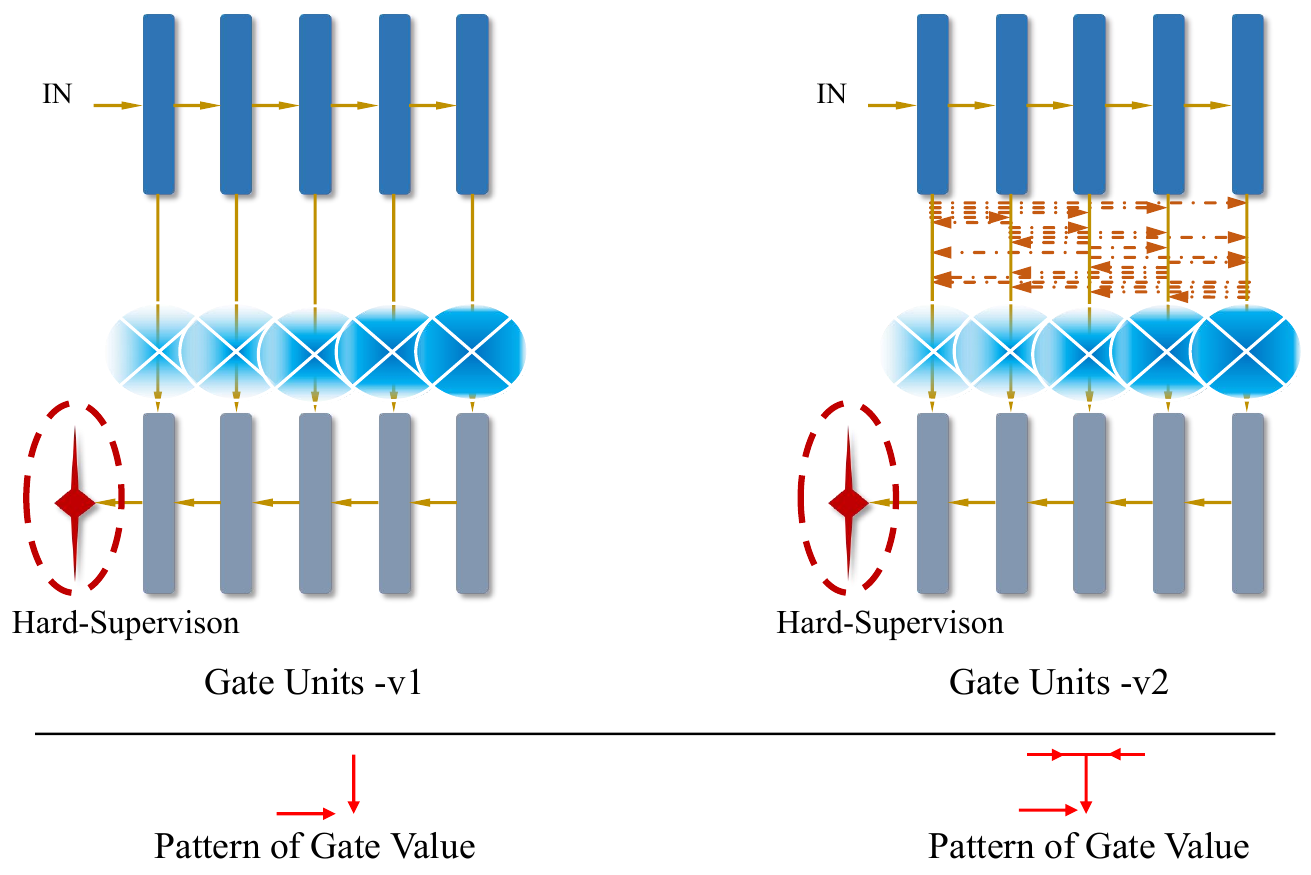}
	\caption{Architecture comparison between the Gated FPN with gate units-v1 and gate units-v2.}
	\label{fig:GateUnit_v1_v2_comparison}
\end{figure}
The gate unit can control the message passing between scale-matching encoder and decoder blocks. By combining the feature maps of the previous decoder block, the gate value also characterizes the contribution that the current block of the encoder can provide. Fig.~\ref{fig:Gate_Unit} shows the internal structure of the proposed gate unit. In particular, the aggregated encoder feature ${E}$ and decoder feature ${D}^{i+1}$ are integrated to obtain feature ${F}^i$, and then the output is fed into two branches, which includes a series of convolution, activation and pooling operations, to compute a pair of gate values ${G}^i$. The entire gated process can be formulated as,
\begin{equation}\label{equ:3}
E = Conv(Cat(E^1,E^2,E^3,E^4,E^5)),
\end{equation}
\begin{equation}\label{equ:gate_value}
{G}^i
= \left\{\begin{matrix}
P(S(Conv(Cat(E, D^{i+1})))) & \text{ if } i=1, 2, 3, 4\\
P(S(Conv(Cat(E, T^i)))), & \text{ if } i=5
\end{matrix}\right.
\end{equation}
where $Cat(\cdot)$ is the concatenation operation among channel axis, $Conv(\cdot)$ is the convolution layer, $S(\cdot)$ is the element-wise sigmoid function, and $P(\cdot)$ is the global average pooling. The output channel of $Conv(\cdot)$ in Eq.~\ref{equ:gate_value} is 2. The resulted gate vector ${G}^i$ has two different elements which correspond to two gate values in Fig.~\ref{fig:GateUnit_v2}. Given the gate values, we can apply them to the FPN branch and the parallel branch to weight the transition-layer features ${T}^1 \sim {T}^5$, which are generated by exploiting $3 \times 3$ convolution to reduce the dimension of ${E}^1 \sim {E}^4$ and the Fold-ASPP to finely process ${E}^5$ (See Fig.~\ref{fig:GateNet} for details). Through multi-level gate units, we can suppress and balance the information flowing from different encoder blocks to the decoder. 

Compared to the ECCV version~\cite{GateNet} of Gate Units-v1, we modify the input feature maps of the current encoder block  ${E}^{i}$  to the all-level aggregated feature maps ${E}$. As shown in Fig.~\ref{fig:GateUnit_v1_v2_comparison}, the Gated FPN with Gate Units-v2 enjoys bidirectional soft supervision, which motivates the gating values of each layer to consider their corresponding contributions from a global perspective, rather than the local perspective in Gate Unit-v1. In this way, the cooperation among the various layers is closer, thereby, making the optimization of the network more efficient.

 \begin{figure}[t]
	\centering
	\includegraphics[width=\linewidth]{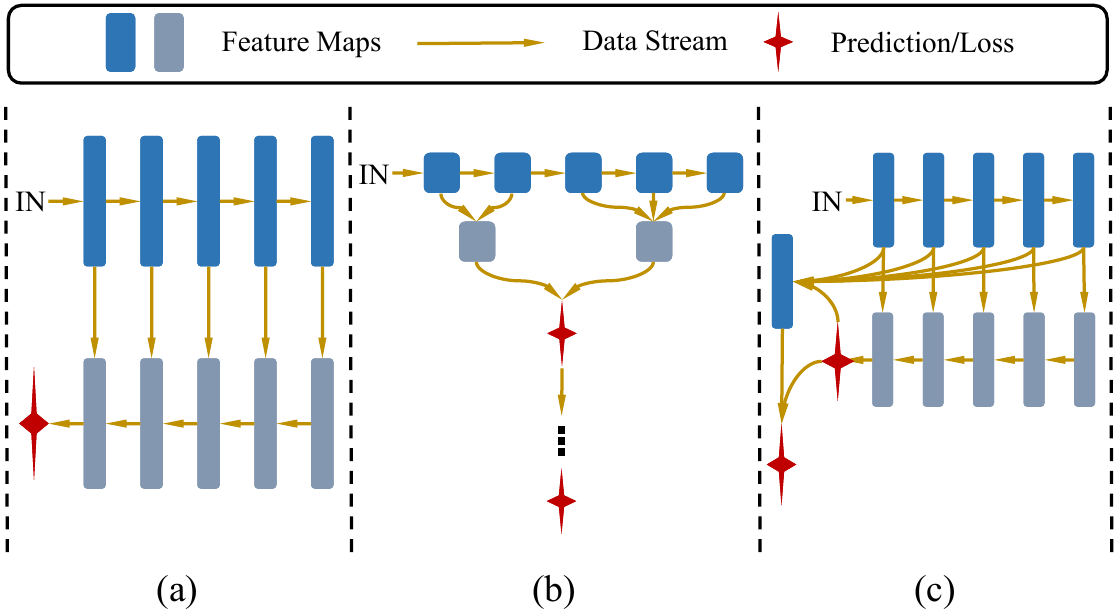}\\
	\caption{ Illustration of different decoder architectures. (a) Progressive structure. (b) Parallel structure. (c) Dual branch structure.}
	\label{fig:Dual_branch}
\end{figure} 
Generally, binary segmentation methods usually adopt the progressive or parallel structure as decoder architecture, as shown in Fig.~\ref{fig:Dual_branch}(a, b). Progressive style is more conducive to the localization of the objects through the high-level feature guidance, while the parallel style is easier to restore details by making full use of low-level features.
As can be seen from Tab.~\ref{table:Survey_DBS_methods_3}, previous methods only adopt either progressive or parallel mode and ignore the advantages brought by the other.  
In this work, we mix the two structures to build a dual branch decoder to overcome the above restrictions.
We briefly describe the FPN branch. Taking $D^i$ as an example, we firstly apply bilinear interpolation to upsample the higher-level feature $D^{i+1}$ to the same size as ${T}^{i}$. Next, to decrease the number of parameters, ${T}^{i}$ is reduced to $32$ channels and fed into gate unit $G^{i}$. Lastly, the gated feature is fused with the upsampled feature of $D^{i+1}$ by element-wise addition and convolutional layers. This process can be formulated as:
\begin{equation}\label{equ:fpn}
D^i = \left\{\begin{matrix}
Conv(G_1^i \cdot T^i + Up(D^{i+1})) & \text{ if } i=1, 2, 3, 4\\
Conv(G_1^i \cdot T^i), & \text{ if } i=5,
\end{matrix}\right.
\end{equation}
where $D^{1}$ is a single-channel feature map with the same size as the input image. 

In the parallel branch, we firstly upsample ${T}^1 \sim {T}^5$ to the same size of $D^1$. Next, the multi-level gate units are followed to weight the corresponding transition-layer features. Lastly, we combine $D^1$ and the gated features by cross-channel concatenation. The whole process is written as:
\begin{equation}\label{equ:1}
\begin{split}
F_{Cat} = Cat(&D^1, Up(G_2^1 \cdot T^1), Up(G_2^2 \cdot T^2),\\
&  Up(G_2^3 \cdot T^3), Up(G_2^4 \cdot T^4), Up(G_2^5 \cdot T^5)) .
\end{split} 
\end{equation}

The final saliency map $S^F$ is generated by integrating the predictions of the two branches with a residual connection as shown in Fig.~\ref{fig:Dual_branch}(c),
\begin{equation}\label{equ:2}
\begin{split}
S^F = S(Conv(F_{Cat}) + D^1)),
\end{split}
\end{equation}
where $S(\cdot)$ is the element-wise sigmoid function.
\begin{figure}[t]
	\centering
	\includegraphics[width=\linewidth]{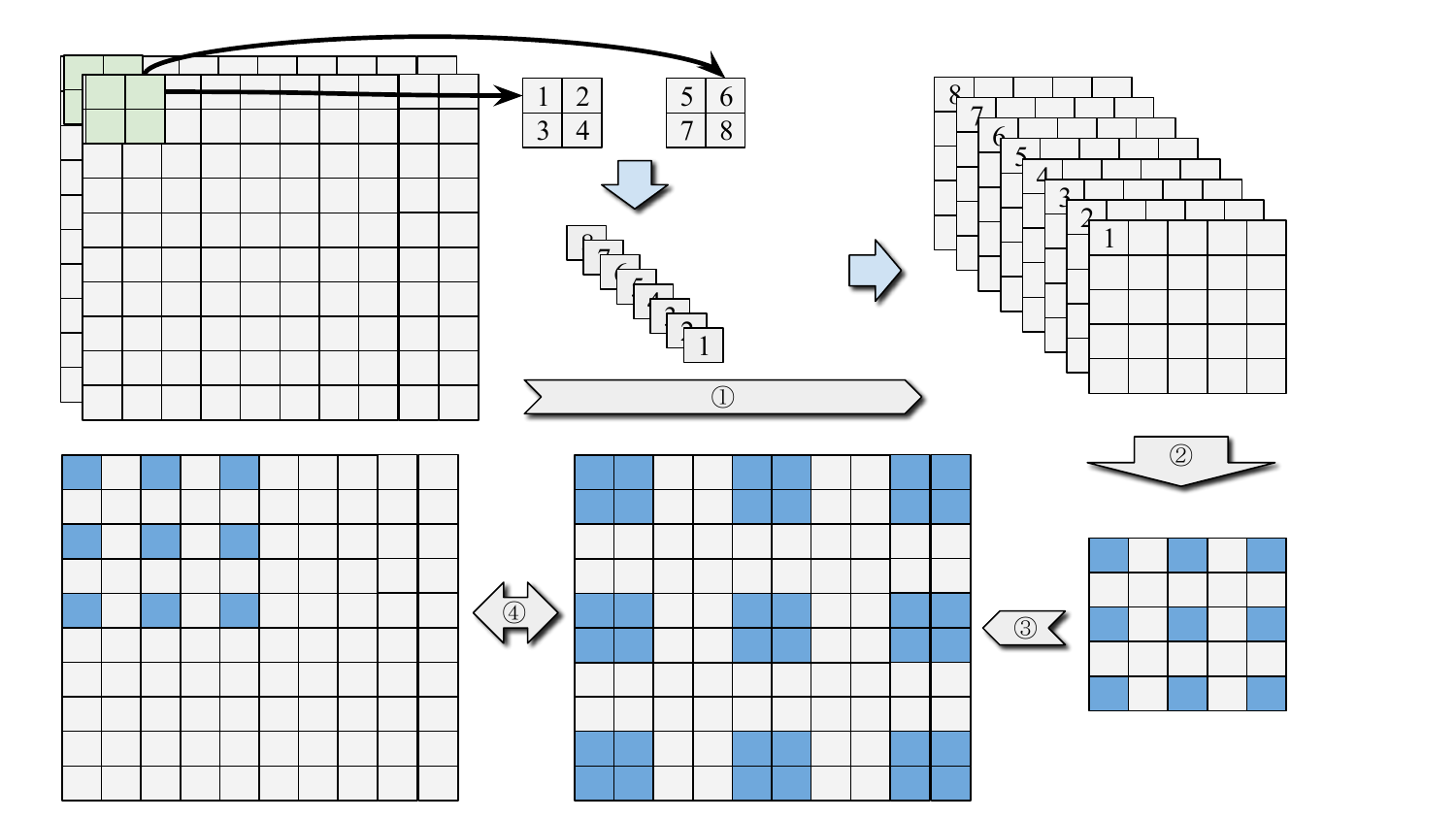}\\
	\caption{ Illustration of the folded atrous convolution. We use \textcircled{\scriptsize 1}, \textcircled{\scriptsize 2} and \textcircled{\scriptsize 3} to respectively indicate ``Fold'' operation, atrous convolution and ``Unfold'' operation. \textcircled{\scriptsize 4} shows the comparison between atrous convolution (Left) and the folded atrous convolution (Right).}
	\label{fig:FoldConv}
\end{figure}
 
 \begin{figure*}[!t]
	\centering
	\includegraphics[width=\linewidth]{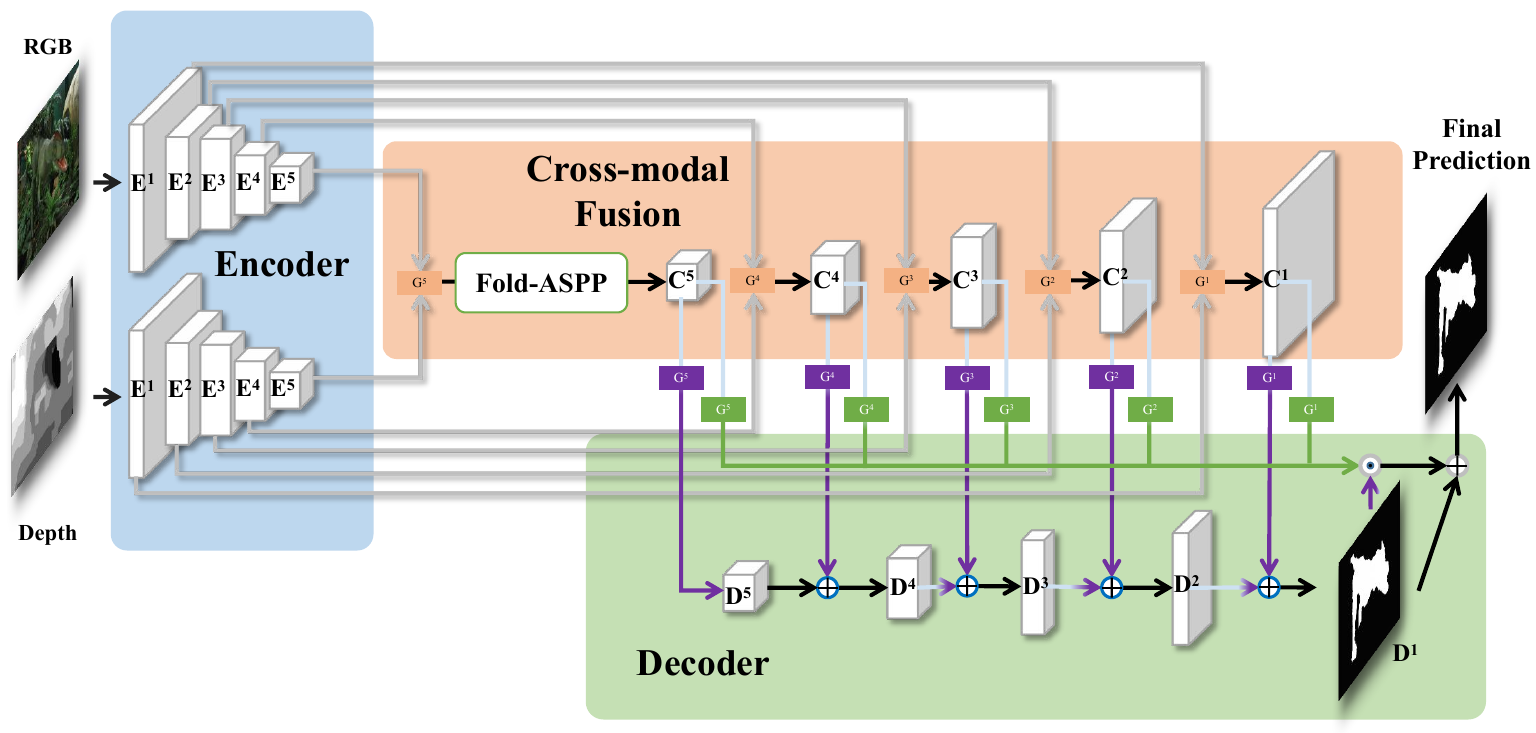}\\
	\caption{Overall pipeline of the two-stream gated network. Firstly, we use two independent encoders to extract features for each modality separately. 
	Fold-ASPP is followed and embedded in the top layer of the encoder. And then, we utilize multi-level gate units to control both cross-modal fusion and fused information transmitted to the decoder. The final prediction is yielded by the gated dual branch.}
	\label{fig:Twostream_GateNet}
\end{figure*}
 
\subsection{Folded Atrous Convolution}\label{sec:Fold}
In order to obtain robust segmentation results by integrating multi-scale information, atrous spatial pyramid pooling (ASPP) is proposed in Deeplab~\cite{ASPP}. And many works~\cite{BMPM,UCNet_RGBDSOD,PFNet_COD,BDRAR_Shadow,AFFPN_Shadow,RCARP_Glass,DMRA_RGBDSOD,CoNet_RGBDSOD,Rank-Net_COD,EBLNet_Glass} also show its effectiveness in different binary segmentation branches. The ASPP uses multiple  parallel  atrous  convolutional layers  with  different  dilation  rates. 
The sparsity of atrous convolution kernel, especially when using a large dilation rate, results in that the association relationships among sampling points are too weak to extract stable features. 
In this paper, we apply a simple ``Fold'' operation to effectively relieve this issue. We visualize the folded atrous convolution structure in Fig.~\ref{fig:FoldConv}, which not only further enlarges the receptive field but also extends each valid sampling position from an isolate point to a $2\times2$ connected region.

Let $\mathbf{X}$ represent feature maps with the size of $N \times N \times C$ (C is the channel number). We slide a $2 \times 2$ window on $\mathbf{X}$ in stride $2$ and then conduct atrous convolution with kernel size  $K \times K$ in different dilation rates. Fig.~\ref{fig:FoldConv} shows the computational process when $K = 3$ and dilation rate is $2$. Firstly, we collect  $2 \times 2 \times C$ feature points in each window from  $\mathbf{X}$ and then it is stacked by channel direction, we call this operation ``Fold'', which is shown in Fig.~\ref{fig:FoldConv}\textcircled{\scriptsize 1}. After the fold operation, we can get new feature maps with the size of $N/2 \times N/2 \times 4C$. A point on the new feature maps corresponds to a $2 \times 2$ area on the original feature maps. Secondly, we adopt an atrous convolution with a kernel size of $3 \times 3$ and dilation rate is $2$. Followed by the reverse process of ``Fold'' which is  called ``Unfold'' operation, the final feature maps are obtained. By using the folded atrous convolution, in the process of information transfer across convolution layers, more contexts are merged and the certain local correlation is also preserved, which provides the fault-tolerance capability for subsequent operations. 

The Fold-ASPP is only equipped on the top of the encoder network, which consists of three folded atrous convolutional layers with  dilation rates $[2, 4, 6]$ to fit the size of feature maps. Just as group convolution~\cite{ResNeXt} is a trade-off between depthwise convolution~\cite{Xception,Mobilenet} and vanilla convolution in the channel dimension, the proposed folded atrous convolution is a trade-off between atrous convolution and vanilla convolution in the spatial dimension.

\subsection{Supervision}\label{sec:supervision}
We use the pixel position aware loss $L_{ppa}$~\cite{F3Net} which have been widely adopted in segmentation tasks. 
We use the same definitions as in~\cite{SPNet_RGBDSOD,F3Net,PraNet_Polyp,MSNet_Polyp,UCNet_RGBDSOD}.  As shown in Fig.~\ref{fig:GateNet}, we apply supervision for both the intermediate prediction from the FPN branch and the final prediction from the dual branch. In the dual branch decoder, since the FPN branch gradually combines all-level gated encoding and decoding features, it has very powerful prediction ability. We expect that it can predict salient objects as accurately as possible under the supervision of ground truth. While the parallel branch only combines the gated encoding features, which is helpful to remedy the ignored details with the design of residual structure. Moreover, the supervision on $D^{1}$ can drive gate units to learn the weight of the contribution of each encoder block to the final prediction. The total loss \emph{L} could be written as:
\begin{equation}\label{equ:5}
\begin{split}
L = L_{ppa}^{s1}+L_{ppa}^{sf},
\end{split}
\end{equation}
where $L_{ppa}^{s1}$ and $L_{ppa}^{sf}$ are respectively used to regularize the output of the FPN branch and the final prediction. 

\subsection{Two-Stream Network}\label{sec:Two_Stream_Network}
To finish some two-source input tasks, \textit{e.g.}, RGB-D salient object detection, we extend the GateNet to a two-stream architecture  to further demonstrate its effectiveness. 
The proposed two-stream GateNet is shown in Fig.~\ref{fig:Twostream_GateNet}. 
Compared with the single-stream network for single source input tasks, there are two main differences: (1) We add an extra encoder to extract features of other modals. (2) We convey the output features from the encoding blocks of two modalities to the gate unit to achieve cross-modal fusion at each level. 
The motivation of embedding gate units when performing cross-modal fusion is straightforward, that is, different modalities present different characteristics in each layer of the encoder, and low-quality modal features can interfere with the other one, leading to build a poor decoder. The form and composition of all gate units are the same as Fig.~\ref{fig:GateUnit_v2} and Eq.~\ref{equ:gate_value} except that the input features are different.

\begin{table*}
	\centering
	\caption{
		{Summary of essential characteristics about popular binary segmentation datasets.}
	}
	\label{table:datasets_survey}
	\begin{threeparttable}
		\resizebox{1\textwidth}{!}{
			\setlength\tabcolsep{6pt}
			\renewcommand\arraystretch{1.02}
            	\begin{tabular}{|r||c|c|c|c|c|}
				\hline\thickhline
				\rowcolor{mygray}
	Dataset&Year&Publication &\#Image & Core Description
     &Role \\
				\hline
				\hline
		\multicolumn{6}{|c|}{\multirow{2}{*}{\large\textbf{RGB Salient Object Detection}}}
			\\
	\multicolumn{6}{|c|}{*}		\\
					\hline
						\hline
MSRA10K~\cite{MSRA10K}&2015 &TPMAI&10,000&Large scale; Multi-object  &Train\\
			
ECSSD~\cite{ECSSD}&2015 &CVPR&1,000& Semantically meaningful but structurally complex natural contents&Test\\
		
HKU-IS~\cite{HKU-IS}&2015 &CVPR&4,447& Multiple disconnected salient objects; Overlapping the image boundary&Test\\
	
PASCAL-S~\cite{PASCAL-S}&2014 &CVPR&850& Selected from the PASCAL VOC2010 val set &Test\\
		
DUT-OMRON~\cite{DUT-OMRON}&2013 &CVPR&5,168& Complicated background and diverse content  &Test\\
		
DUTS~\cite{DUTS}&2017 &CVPR&15,572& Large-scale; Complex scenarios with high-diversity contents; Most SOD models are typically trained on it   &Train\&Test\\
				\hline
				\hline
		\multicolumn{6}{|c|}{\multirow{2}{*}{\large\textbf{RGB-D Salient Object Detection}}}
			\\
						\multicolumn{6}{|c|}{*}		\\
					\hline
						\hline
NJUD~\cite{NJUD}&2014 &ICIP&1,985&Complex objects and challenging scenarios; Selected from 3D movies, the Internet, and photographs
taken by a Fuji W3 stereo camera  &Train\&Test\\

NLPR~\cite{NLPR}&2014 &ECCV&1,000&There may exist multiple
salient objects in each image; Captured by Kinect &Train\&Test\\
			
DUTLF-D~\cite{DMRA_RGBDSOD}&2019 &ICCV&1,200&Contains 800 indoor and
400 outdoor scenes paired with the depth map and binary ground truth&Trian\&Test\\
			
STERE~\cite{STERE}&2012 &CVPR&1,000&The first
stereoscopic photo collection; Images downloaded from the Internet &Test\\
			
SIP~\cite{D3Net_RGBDSOD}&2021 &TNNLS&929&High-resolution images; Contain
multiple salient persons per image &Test\\

SSD~\cite{SSD}&2017 &ICCVW&80&Collected from three stereo movies &Test\\
		
RGBD135~\cite{RGBD135}&2014 &ICIMCS&135&Consists of seven indoor scenes; Captured by Kinect &Test\\
		
LFSD~\cite{LFSD}&2014 &CVPR&100&Manly built for saliency detection on the light filed &Test\\
				\hline
			\hline
		\multicolumn{6}{|c|}{\multirow{2}{*}{\large\textbf{ORSI Salient Object Detection}}}
		\\
						\multicolumn{6}{|c|}{*}		\\
					\hline
						\hline
ORSSD~\cite{LV-Net_RSISOD}&2019 &TGRS&800 &Diverse spatial resolution; Background is
cluttered and complicated;  Type of salient objects is various; The number and size of salient objects are variable &Train\&Test\\

EORSSD~\cite{DAFNet_RSISOD}&2021 &TIP&2,000&Multiple
salient objects in one image; A
number of small objects; More abundant scenarios; Imaging interferences; Specific circumstances &Train\&Test\\

ORSI-4199~\cite{MJRBM_RSISOD}&2021 &TGRS&4,199&Large-scale; More challenges with background interference samples   &Train\&Test\\
				\hline
			\hline
		\multicolumn{6}{|c|}{\multirow{2}{*}{\large\textbf{Camouflaged Object Detection}}}\\
						\multicolumn{6}{|c|}{*}		\\
					\hline
						\hline
CHAMELEON~\cite{CHAMELEON}&2018 &Unpublished Manuscript&76 &The images were collected from the Internet via the Google search engine &Test\\

CAMO~\cite{CAMO}&2019 &CVIU&2,500 & Eight categories &Train\&Test\\

COD10K~\cite{SINet_COD}&2020&CVPR&10,000 &Large-scale; Broader size distribution; Global/Local contrast; Rich sub-classes; A large number of Full HD 1080p images  &Train\&Test\\

NC4K~\cite{Rank-Net_COD}&2021&CVPR&4,121 &Large-scale; Evaluate the generalization ability of existing models  &Test\\
				\hline
			\hline
		\multicolumn{6}{|c|}{\multirow{2}{*}{\large\textbf{Defocus Blur Detection}}}
				\\
						\multicolumn{6}{|c|}{*}		\\
					\hline
						\hline
CUHK~\cite{CUHK}&2014 & CVPR&704 &Cluttered backgrounds and various scenes&Train\&Test\\
DUT~\cite{BTBNet_DBD}&2018 & CVPR&1,100 &Multi-scale focused areas&Train\&Test\\
				\hline
			\hline
		\multicolumn{6}{|c|}{\multirow{2}{*}{\large\textbf{Shadow Detection}}}
				\\
						\multicolumn{6}{|c|}{*}		\\
					\hline
						\hline
UCF~\cite{UCF}&2010 & CVPR&245 &Cluttered backgrounds and various scenes&Test\\
ISTD~\cite{ISTD}&2018 & CVPR&1,870 &Both shadow detection and removal; Only 135 background
scenes&Train\&Test\\
SBU~\cite{SBU}&2016 & ECCV&4,727 &The largest shadow dataset
covering general scenes; A wider variety of scenes&Train\&Test\\
			\hline
			\hline
		\multicolumn{6}{|c|}{\multirow{2}{*}{\large\textbf{Transparent Object Detection}}}
				\\
						\multicolumn{6}{|c|}{*}		\\
					\hline
						\hline
Trans10K~\cite{TransLab_Transparent}&2020 & ECCV&10,428 &Two categories (stuff and things); Large-scale realworld images; Complex scenarios &Train\&Test\\
				
Trans10K-v2~\cite{Trans2Seg_Transparent}&2021 & IJCAI&10,428 &Semantic Segmentation; 11 fine-grained glass image categories with a diverse scenario and high resolution&Train\&Test\\
	\hline
			\hline
		\multicolumn{6}{|c|}{\multirow{2}{*}{\large\textbf{Glass Detection}}}
				\\
						\multicolumn{6}{|c|}{*}		\\
					\hline
						\hline
GDD~\cite{GDNet_Glass}&2020 & CVPR&3,916 &Large-scale; Both indoor and outdoor scenes; Various types of common glass; Glass located at different positions of an image &Train\&Test\\
		\hline
			\hline
		\multicolumn{6}{|c|}{\multirow{2}{*}{\large\textbf{Mirror Detection}}}
				\\
						\multicolumn{6}{|c|}{*}		\\
					\hline
						\hline
MSD~\cite{MirrorNet_Mirror}&2019 & ICCV&4,018 &Large-scale; Both indoor and outdoor scenes;  Different mirror shapes and multiple mirrors; Low global color contrast &Train\&Test\\
		\hline
			\hline
		\multicolumn{6}{|c|}{\multirow{2}{*}{\large\textbf{Polyp Segmentation}}}
				\\
						\multicolumn{6}{|c|}{*}		\\
					\hline
						\hline
CVC-ColonDB~\cite{CVC-ColonDB}&2015 & TMI&380 &Colonoscopy images each with a polyp inside; Selected from 15 short colonoscopy videos &Test\\
			
CVC-ClinicDB~\cite{CVC-ClinicDB}&2015 & CMIG&612 &Images from 31 colonoscopy clip; Images of size 576 $\times$ 768&Train$\&$Test\\
					
EndoScene~\cite{Endoscene}&2017 & JHE&912 &Images from CVC-ColonDB and CVC-ClinicDB and are reannotated; Extend the old annotations to account for lumen, and specular highlights &Train\&Test\\
				
ETIS~\cite{ETIS}&2014 & IJCARS&196 &An early established dataset for early diagnosis of colorectal cancer &Test\\
				
Kvasir~\cite{Kvasir}&2020 & MMM&1,000 &Existing largest-scale dataset &Train$\&$Test\\
						\hline
			\end{tabular}
		}
	\end{threeparttable}
\end{table*}

\section{Experiments}
\subsection{Datasets}
For the training and test dataset, we follow the settings of the most state-of-the-art methods~\cite{VST,TriTransNet_RGBDSOD,UACANet_Polyp,UGTR_COD,DENets_DBD,DSDNet_Shadow,TransLab_Transparent,GDNet_Glass,MirrorNet_Mirror,MJRBM_RSISOD} in Tab.~\ref{table:Survey_DBS_methods_1} on each binary segmentation task. 
And the details about these datasets can find in Tab.~\ref{table:datasets_survey}. 
\subsection{Evaluation Metrics}
There are ten popular metrics used in different binary segmentation branches. 
F-measure~\cite{colorcontrast_Fm} ($F_{\beta}^{max}$, $F_{\beta}^{mean}$), weighted F-measure~\cite{Fwb} ($F_{\beta}^{\omega}$), S-measure~\cite{S-m} ($S_m$), E-measure~\cite{E-m} ($E_m$) and MAE~\cite{MAE} ($\mathcal{M}$) are widely used in salient object detection, camouflaged object detection and defous blur detection tasks.
IOU and Dice scores are popular with medical image segmentation. 
BER~\cite{BER} and Pixel Accuracy (PA) are more commonly used for shadow, mirror, glass and transparent detection. 
 The lower value is better for the BER and MAE, and higher is better for others.
 
\noindent\textbullet~\textbf{Pixel Accuracy ($PA$)} 
 is calculated based on {the binarized} prediction mask and ground-truth:
 \begin{equation}\label{equation:PA}\small
   PA= \frac{TP+TN}{TP+TN+FP+FN},
 \end{equation}
 where TP, TN, FP, FN denote true-positive, true-negative, false-positive, and false-negative, respectively.

 \noindent\textbullet~\textbf{F-measure ($F_{\beta}$)~\cite{colorcontrast_Fm}}
 is a metric that comprehensively considers both precision and recall:
 \begin{equation}\label{equation:Fm}\small
   F_{\beta}= \frac{(1+\beta^2)\text{Precision}\times \text{Recall}}{\beta^2 \text{Precision} + \text{Recall}},
 \end{equation}
 \begin{equation}\label{equation:pr}\small
   \text{Precision}= \frac{{TP}}{{TP}+{FP}},~~~~\text{Recall}= \frac{{TP}}{{TP}+{FN}},
 \end{equation}
 where $\beta^2$ is set to $0.3$ as suggested in~\cite{colorcontrast_Fm} to emphasize the precision.
 Some methods report the maximum F-measure ($F_{\beta}^{max}$) across the binary maps of different thresholds or the mean F-measure ($F_{\beta}^{mean}$) score by an adaptive threshold.

 \noindent\textbullet~\textbf{weighted F-measure ($F_{\beta}^{\omega}$)~\cite{Fwb}} 
 is proposed to improve the metric F-measure. 
 It assigns different weights ($\omega$) to precision and recall across different errors at different locations, considering the neighborhood information: 
 \begin{equation}\label{equation:wFm}\small
   F_\beta^\omega= \frac{(1+\beta^2)\text{Precision}^\omega\times \text{Recall}^\omega}{\beta^2 \text{Precision}^\omega + \text{Recall}^\omega}.
 \end{equation}

 \noindent\textbullet~\textbf{S-measure ($S_m$)~\cite{S-m}} 
 evaluates the spatial structure similarity by combining the region-aware structural similarity $S _ { r }$ and the object-aware structural similarity $S _ { o}$: 
 \begin{equation}\label{equation:Sm}\small
 S_m = \alpha \times S_{ o } + ( 1 - \alpha ) \times S_{ r },
 \end{equation}
 where $\alpha$ is empirically set to $0.5$.

 \noindent\textbullet~\textbf{E-measure ($E_m$)~\cite{E-m}}
 can jointly capture image level statistics and local pixel matching information:
 \begin{equation}\label{equation:Em}\small
   Q_{\bm{S}}= \frac{1}{W\!\times\!H} \sum\nolimits_{i=1}^{W} \sum\nolimits_{j=1}^{H} \phi_{\bm{S}}(i,j),
 \end{equation}
 where $\phi_{\bm{S}}$ is the enhanced alignment matrix, reflecting the correlation between prediction $\bm{S}$ and the ground truth $\bm{G}$ after subtracting their global means, respectively.

 \noindent\textbullet~\textbf{IOU}
 is the most common metric for evaluating classification accuracy:
 \begin{equation}\label{equation:IOU}\small
 IOU  = \frac{TP}{TP+FP+FN}.
 \end{equation}

 \noindent\textbullet~\textbf{Dice}
  is a statistic used to gauge the similarity of two samples and become the most used metric in validating medical image segmentation:
 \begin{equation}\label{equation:Dice}\small
 Dice  = \frac{2TP}{FP+2TP+FN}.
 \end{equation}

 \noindent\textbullet~\textbf{Balanced error rate ($BER$)~\cite{BER}}
  is a common metric to evaluate shadow detection performance, where shadow and non-shadow regions contribute equally to the overall performance without considering their relative areas:
 \begin{equation}\label{equation:BER}\small
 BER \ = \ (1-\frac{1}{2}(\frac{TP}{TP+FN}+\frac{TN}{TN+FP})).
 \end{equation}

 \noindent\textbullet~\textbf{MAE ($\mathcal{M}$)~\cite{MAE}} 
 measures the average absolute difference between the prediction $\bm{S}\!\in\![0,1]^{W\!\times\! H}$ and the  ground truth $\bm{G}\!\in\!\{0,1\}^{W\!\times\! H}$ pixel by pixel:
 \begin{equation}\label{equation:mae}\small
   \text{MAE}= \frac{1}{W\!\times\!H} \sum\nolimits_{i=1}^{W} \sum\nolimits_{j=1}^{H} \lvert \bm{G}(i,j)-\bm{S}(i,j) \rvert.
 \end{equation}

In fact, all above metrics can be used for any binary segmentation sub-task. In this paper, we are the first to introduce all ten metrics into the quantitative comparison to provide a comprehensive performance evaluation.
\subsection{Implementation Details}
We use the PyTorch framework to implement our models on  one RTX 3090 GPU for $100$ epochs.
The input resolutions of images are resized to $352\times352$ and we employ a  general multi-scale training strategy as most methods~\cite{F3Net,GCPANet,Rank-Net_COD,SPNet_RGBDSOD,PraNet_Polyp,MSNet_Polyp}.
We adopt some image augmentation techniques to avoid overfitting, including random flipping,  rotating, and border clipping. For the optimizer, we use the Adam~\cite{Adam}. For the learning rate, initial learning rate  is  set  to $0.0001$. We adopt the ``step'' learning rate decay policy, and set the decay size as $30$ and decay rate as $0.9$.
For any sub-tasks, the above training strategy is used for all the gated network models involved in this paper.
The difference among these models is only in the mini-batch size due to adopting different backbones.
Specifically, the mini-batch size settings in the gatenet using VGG-16, ResNet-50, Res2Net-50, and ResNeXt-101 as the backbone are $8$, $24$, $24$ and $16$, respectively. 
The source code can be available at \url{https://github.com/Xiaoqi-Zhao-DLUT/GateNet-RGB-Saliency}.
\subsection{Performance}
We compare our models with state-of-the-art approaches in terms of ten metrics on all test sets corresponding for each binary segmentation task in Tab.~\ref{tab:RGBSOD_performance} - Tab.~\ref{tab:ORSISOD_performance}.
Since there are many test sets for RGB SOD, RGB-D SOD and polyp segmentation, we not only compare the performance under each metric, but also count the proportion of top $3$ and top $1$ performance to get an overall evaluation. Some quantitative analyses are as follows:
\begin{table}[!t]
\large
\centering
	\scriptsize
	\setlength{\abovecaptionskip}{2pt}
	\caption{Quantitative comparison of different RGB SOD methods.  Top $3$ and Top $1$ scores are highlighted in {\color{myblue}{\textbf{blue}}} and {\color{reda}{\textbf{red}}}, respectively.}
 \begin{threeparttable}
   \resizebox{\linewidth}{!}{
    \setlength\tabcolsep{5pt}
    \renewcommand\arraystretch{0.9}
    		\begin{tabular}{|lr||ccccccccc|}
		\hline\thickhline
		 \rowcolor{mygray}
			\multicolumn{2}{|c||}{\multirow{5}{*}{Metric}} 
			&{F3Net} &{ITSD} &{MINet} &{KRN}&{Auto-MSF} &   {LDF} &   {VST} & {CTDNet} &{GateNet}
			\\
		
			\multicolumn{2}{|l||}{}   & ~\cite{F3Net}&~\cite{ITSD}&~\cite{MINet}&~\cite{KRN}&~\cite{Auto-MSFNet}&~\cite{LDF}&~\cite{VST} & ~\cite{CTDNet}&
			\\
		\multicolumn{2}{|l||}{}   & AAAI&  CVPR&  CVPR& AAAI& ACMMM& CVPR& ICCV&ACMMM& --
			\\
		\multicolumn{2}{|l||}{}   & 2020&  2020&  2020& 2021& 2021& 2020& 2021&2021& --
			\\
			\multicolumn{2}{|l||}{}   & Res-50&  Res-50& Res-50 & Res-50& Res-50&  Res-50& T2T & Res-50&Res-50
			\\
			\hline
			\hline
			\multirow{10}{*}{\emph{\rotatebox{90}{DUTS~\cite{DUTS}}}}      
			&$PA\uparrow$   
&0.966 
&0.962 
&0.965 
&0.966 
&\color{myblue}\textbf{0.968} 
&\color{myblue}\textbf{0.968} 
&\color{myblue}\textbf{0.967} 
&\color{myblue}\textbf{0.968} 
&\color{reda}\textbf{0.972}

			\\
			&$F_{\beta}^{max}\uparrow$    
&0.891 
&0.883 
&0.884 
&0.877 
&\color{myblue}\textbf{0.898} 
&\color{myblue}\textbf{0.898} 
&0.890 
&\color{myblue}\textbf{0.897} 
&\color{reda}\textbf{0.911} 

			\\
			&$F_{\beta}^{mean}\uparrow$ 
&0.840 
&0.804 
&0.828 
&\color{myblue}\textbf{0.856} 
&0.851 
&\color{myblue}\textbf{0.855} 
&0.818 
&0.853 
&\color{reda}\textbf{0.857} 
			\\
			&$F_{\beta}^{\omega}\uparrow$	
&0.835 
&0.824 
&0.825 
&0.841 
&\color{myblue}\textbf{0.847} 
&\color{myblue}\textbf{0.845} 
&0.828 
&\color{myblue}\textbf{0.847} 
&\color{reda}\textbf{0.864} 

			\\
			&$S_m\uparrow$ 		
&0.887 
&0.883 
&0.883 
&0.876 
&\color{myblue}\textbf{0.891} 
&\color{myblue}\textbf{0.891} 
&\color{myblue}\textbf{0.895} 
&\color{myblue}\textbf{0.891} 
&\color{reda}\textbf{0.906} 

			\\
				&$E_m\uparrow$ 	
&0.918 
&0.898 
&0.917 
&\color{reda}\textbf{0.931} 
&0.926 
&\color{myblue}\textbf{0.929} 
&0.916 
&\color{myblue}\textbf{0.928} 
&\color{reda}\textbf{0.931} 
			\\
			&$IOU\uparrow$ 	
&0.793 
&0.783 
&0.782 
&0.779 
&0.799 
&0.799 
&\color{myblue}\textbf{0.802} 
&\color{myblue}\textbf{0.800} 
&\color{reda}\textbf{0.828} 

			\\
			&$Dice\uparrow$ 
&0.854 
&0.844 
&0.845 
&0.855 
&\color{myblue}\textbf{0.864} 
&\color{myblue}\textbf{0.861} 
&0.848 
&\color{myblue}\textbf{0.864} 
&\color{reda}\textbf{0.878} 

			\\
			&$BER\downarrow$	
&0.062 
&0.065 
&0.069 
&0.072 
&0.064 
&0.064 
&\color{myblue}\textbf{0.060} 
&\color{myblue}\textbf{0.061} 
&\color{reda}\textbf{0.052} 

			\\
			&$\mathcal{M}\downarrow$	
&\color{myblue}\textbf{0.035} 
&0.041 
&0.037 
&\color{myblue}\textbf{0.034} 
&\color{myblue}\textbf{0.034} 
&\color{myblue}\textbf{0.034} 
&0.037 
&\color{myblue}\textbf{0.034} 
&\color{reda}\textbf{0.030} 

			\\
		\hline
			\multirow{10}{*}{\emph{\rotatebox{90}{DUT-OMRON~\cite{DUT-OMRON}}}}      
			&$PA\uparrow$   
&0.949 
&0.942 
&0.946 
&\color{myblue}\textbf{0.951} 
&\color{reda}\textbf{0.953} 
&0.950 
&0.946 
&0.949 
&\color{myblue}\textbf{0.952}

			\\
			&$F_{\beta}^{max}\uparrow$    
&0.813 
&0.821 
&0.810 
&0.798 
&\color{reda}\textbf{0.827} 
&0.820 
&\color{myblue}\textbf{0.825} 
&\color{myblue}\textbf{0.826} 
&0.824 

			\\
			&$F_{\beta}^{mean}\uparrow$ 
&0.766 
&0.756 
&0.756 
&0.778 
&\color{reda}\textbf{0.783} 
&0.774 
&0.756 
&\color{myblue}\textbf{0.779} 
&\color{myblue}\textbf{0.781}

			\\
			&$F_{\beta}^{\omega}\uparrow$	
&0.747 
&0.750 
&0.738 
&\color{myblue}\textbf{0.757} 
&\color{reda}\textbf{0.765} 
&0.752 
&0.755 
&\color{myblue}\textbf{0.762} 
&\color{reda}\textbf{0.765} 

			\\
			&$S_m\uparrow$ 	
&0.837 
&0.839 
&0.832 
&0.831 
&\color{myblue}\textbf{0.845} 
&0.838 
&\color{reda}\textbf{0.849} 
&0.842 
&\color{myblue}\textbf{0.847} 

			\\
				&$E_m\uparrow$ 
&0.876 
&0.867 
&0.873 
&0.876 
&\color{reda}\textbf{0.889} 
&0.881 
&0.872 
&\color{myblue}\textbf{0.884} 
&\color{myblue}\textbf{0.882} 

			\\
			&$IOU\uparrow$ 	
&0.710 
&0.715 
&0.699 
&0.705 
&\color{myblue}\textbf{0.723} 
&0.711 
&\color{reda}\textbf{0.731} 
&0.720 
&\color{myblue}\textbf{0.730} 

			\\
			&$Dice\uparrow$ 
&0.772 
&0.778 
&0.764 
&0.776 
&\color{myblue}\textbf{0.786} 
&0.775 
&\color{myblue}\textbf{0.783} 
&\color{reda}\textbf{0.787} 
&\color{myblue}\textbf{0.786} 

			\\
			&$BER\downarrow$	
&0.101 
&\color{myblue}\textbf{0.091} 
&0.106 
&0.107 
&0.098 
&0.102 
&\color{reda}\textbf{0.088} 
&\color{myblue}\textbf{0.095} 
&0.097 

			\\
			&$\mathcal{M}\downarrow$
&0.053 
&0.061 
&0.056 
&\color{myblue}\textbf{0.050} 
&\color{reda}\textbf{0.049} 
&\color{myblue}\textbf{0.052} 
&0.058 
&\color{myblue}\textbf{0.052} 
&\color{myblue}\textbf{0.050} 

			\\
				\hline
			\multirow{10}{*}{\emph{\rotatebox{90}{ECSSD~\cite{ECSSD}}}}      
			&$PA\uparrow$   
&\color{myblue}\textbf{0.969} 
&\color{myblue}\textbf{0.969} 
&\color{myblue}\textbf{0.969} 
&0.967 
&0.967 
&0.968 
&\color{myblue}\textbf{0.973} 
&\color{myblue}\textbf{0.969} 
&\color{reda}\textbf{0.977}

			\\
			&$F_{\beta}^{max}\uparrow$    
&0.945 
&0.947 
&0.948 
&0.941 
&0.946 
&\color{myblue}\textbf{0.950} 
&\color{myblue}\textbf{0.951} 
&\color{myblue}\textbf{0.950} 
&\color{reda}\textbf{0.960} 

			\\
			&$F_{\beta}^{mean}\uparrow$ 
&0.925 
&0.895 
&0.924 
&\color{myblue}\textbf{0.929} 
&0.922 
&\color{myblue}\textbf{0.930} 
&0.920 
&0.927 
&\color{reda}\textbf{0.931}

			\\
			&$F_{\beta}^{\omega}\uparrow$
&0.912 
&0.911 
&0.911 
&\color{myblue}\textbf{0.916} 
&0.910 
&\color{myblue}\textbf{0.915} 
&0.910 
&\color{myblue}\textbf{0.915} 
&\color{reda}\textbf{0.931} 

			\\
			&$S_m\uparrow$ 	
&0.924 
&\color{myblue}\textbf{0.925} 
&\color{myblue}\textbf{0.925} 
&0.914 
&0.923 
&0.924 
&\color{myblue}\textbf{0.932} 
&\color{myblue}\textbf{0.925} 
&\color{reda}\textbf{0.941} 

			\\
				&$E_m\uparrow$ 	
&0.946 
&0.932 
&0.953 
&\color{myblue}\textbf{0.954} 
&0.942 
&0.951 
&\color{myblue}\textbf{0.957} 
&0.949 
&\color{reda}\textbf{0.959} 

			\\
			&$IOU\uparrow$ 	
&0.879 
&0.879 
&0.879 
&0.871 
&0.876 
&0.880 
&\color{myblue}\textbf{0.893} 
&\color{myblue}\textbf{0.881} 
&\color{reda}\textbf{0.909} 

			\\
			&$Dice\uparrow$ 
&0.921 
&0.919 
&0.922 
&0.922 
&0.919 
&\color{myblue}\textbf{0.923} 
&0.922 
&\color{myblue}\textbf{0.924} 
&\color{reda}\textbf{0.940} 

			\\
			&$BER\downarrow$	
&0.045 
&0.044 
&\color{myblue}\textbf{0.043} 
&0.048 
&0.047 
&0.045 
&\color{myblue}\textbf{0.036} 
&0.044 
&\color{reda}\textbf{0.031} 

			\\
			&$\mathcal{M}\downarrow$	
&\color{myblue}\textbf{0.033} 
&0.035 
&0.034 
&\color{myblue}\textbf{0.033} 
&0.036 
&0.034 
&\color{myblue}\textbf{0.033} 
&\color{myblue}\textbf{0.032} 
&\color{reda}\textbf{0.026} 

			\\
				\hline
			\multirow{10}{*}{\emph{\rotatebox{90}{HKU-IS~\cite{HKU-IS}}}}      
			&$PA\uparrow$   
&0.973 
&0.972 
&\color{myblue}\textbf{0.974} 
&0.973 
&\color{myblue}\textbf{0.974} 
&\color{myblue}\textbf{0.974} 
&\color{myblue}\textbf{0.976} 
&\color{myblue}\textbf{0.974} 
&\color{reda}\textbf{0.977}

			\\
			&$F_{\beta}^{max}\uparrow$    
&0.937 
&0.933 
&0.935 
&0.928 
&0.937 
&0.940 
&\color{myblue}\textbf{0.942} 
&\color{myblue}\textbf{0.941} 
&\color{reda}\textbf{0.948} 

			\\
			&$F_{\beta}^{mean}\uparrow$ 
&0.910 
&0.898 
&0.908 
&0.914 
&\color{myblue}\textbf{0.915} 
&\color{myblue}\textbf{0.915} 
&0.901 
&\color{myblue}\textbf{0.918} 
&\color{reda}\textbf{0.920}

			\\
			&$F_{\beta}^{\omega}\uparrow$	
&0.900 
&0.893 
&0.899 
&0.904 
&0.902 
&\color{myblue}\textbf{0.905} 
&0.898 
&\color{myblue}\textbf{0.908} 
&\color{reda}\textbf{0.916} 

			\\
			&$S_m\uparrow$ 	
&0.916 
&0.916 
&0.919 
&0.908 
&0.918 
&\color{myblue}\textbf{0.920} 
&\color{myblue}\textbf{0.928} 
&\color{myblue}\textbf{0.920} 
&\color{reda}\textbf{0.931} 

			\\
				&$E_m\uparrow$ 
&0.959 
&0.953 
&\color{myblue}\textbf{0.961} 
&0.960 
&0.960 
&\color{myblue}\textbf{0.962} 
&\color{myblue}\textbf{0.961} 
&\color{myblue}\textbf{0.962} 
&\color{reda}\textbf{0.965} 

			\\
			&$IOU\uparrow$ 	
&0.862 
&0.857 
&0.862 
&0.853 
&0.863 
&0.867 
&\color{myblue}\textbf{0.877} 
&\color{myblue}\textbf{0.869} 
&\color{reda}\textbf{0.886} 

			\\
			&$Dice\uparrow$ 	
&0.911 
&0.904 
&0.910 
&0.912 
&0.911 
&\color{myblue}\textbf{0.915} 
&0.911 
&\color{myblue}\textbf{0.917} 
&\color{reda}\textbf{0.925} 

			\\
			&$BER\downarrow$	
&\color{myblue}\textbf{0.045} 
&0.048 
&0.047 
&0.051 
&0.048 
&\color{myblue}\textbf{0.045} 
&\color{myblue}\textbf{0.038} 
&\color{myblue}\textbf{0.045} 
&\color{reda}\textbf{0.037} 

			\\
			&$\mathcal{M}\downarrow$	
&\color{myblue}\textbf{0.028} 
&0.031 
&\color{myblue}\textbf{0.028} 
&\color{myblue}\textbf{0.027} 
&0.029 
&\color{myblue}\textbf{0.027} 
&0.029 
&\color{myblue}\textbf{0.028} 
&\color{reda}\textbf{0.025} 

			\\
				\hline
			\multirow{10}{*}{\emph{\rotatebox{90}{PASCAL-S~\cite{PASCAL-S}}}}      
			&$PA\uparrow$   
&0.938 
&0.937 
&0.936 
&0.937 
&0.932 
&\color{myblue}\textbf{0.939} 
&\color{myblue}\textbf{0.943} 
&0.938 
&\color{reda}\textbf{0.947}

			\\
			&$F_{\beta}^{max}\uparrow$    
&0.882 
&0.882 
&0.880 
&0.874 
&0.886 
&0.887 
&\color{myblue}\textbf{0.890} 
&\color{myblue}\textbf{0.889} 
&\color{reda}\textbf{0.900} 

			\\
			&$F_{\beta}^{mean}\uparrow$ 
&0.844 
&0.797 
&0.840 
&\color{reda}\textbf{0.854} 
&0.842 
&\color{myblue}\textbf{0.853} 
&0.842 
&\color{myblue}\textbf{0.851} 
&0.848

			\\
			&$F_{\beta}^{\omega}\uparrow$	
&0.823 
&0.823 
&0.818 
&\color{myblue}\textbf{0.830} 
&0.823 
&\color{myblue}\textbf{0.829} 
&0.827 
&\color{myblue}\textbf{0.829} 
&\color{reda}\textbf{0.846} 

			\\
			&$S_m\uparrow$ 
&0.857 
&\color{myblue}\textbf{0.859} 
&0.854 
&0.849 
&0.854 
&\color{myblue}\textbf{0.859} 
&\color{myblue}\textbf{0.871} 
&\color{myblue}\textbf{0.859} 
&\color{reda}\textbf{0.875} 
 
			\\
				&$E_m\uparrow$ 	
&0.892 
&0.866 
&0.897 
&\color{myblue}\textbf{0.902} 
&0.881 
&\color{myblue}\textbf{0.903} 
&\color{reda}\textbf{0.905} 
&0.898 
&\color{myblue}\textbf{0.902} 
 
			\\
			&$IOU\uparrow$ 	
&0.780 
&0.782 
&0.773 
&0.775 
&0.773 
&\color{myblue}\textbf{0.783} 
&\color{myblue}\textbf{0.801} 
&\color{myblue}\textbf{0.783} 
&\color{reda}\textbf{0.808} 

			\\
			&$Dice\uparrow$ 	
&0.848 
&0.849 
&0.843 
&0.850 
&0.844 
&0.852 
&\color{myblue}\textbf{0.858} 
&\color{myblue}\textbf{0.853} 
&\color{reda}\textbf{0.870} 

			\\
			&$BER\downarrow$	
&0.080 
&\color{myblue}\textbf{0.078} 
&0.084 
&0.084 
&0.085 
&0.081 
&\color{myblue}\textbf{0.066} 
&\color{myblue}\textbf{0.078} 
&\color{reda}\textbf{0.065} 

			\\
			&$\mathcal{M}\downarrow$	
&0.064 
&0.066 
&0.066 
&\color{myblue}\textbf{0.063} 
&0.070 
&\color{myblue}\textbf{0.062} 
&\color{myblue}\textbf{0.062} 
&0.064 
&\color{reda}\textbf{0.055} 

			\\
			\hline\hline
	\rowcolor{mygray}			&$Top$ $3$ 
& 5/50
& 5/50
& 6/50
& 15/50
& 18/50
& 28/50
& 30/50
& 37/50
& 49/50
\\
	\rowcolor{mygray}			&$Top$ $1$  
&0/50
&0/50
&0/50
&1/50
&5/50
&0/50
&3/50
&1/50
&40/50
\\
	\hline
		\end{tabular}
	}
	\setlength{\abovecaptionskip}{2pt}
	\label{tab:RGBSOD_performance}
	\end{threeparttable}
\end{table}
\begin{table}[!t]
\large
	\scriptsize
	\centering
	\setlength{\abovecaptionskip}{2pt}
	\caption{Quantitative comparison of different polyp segmentation methods. Top $3$ and Top $1$ scores are highlighted in {\color{myblue}{\textbf{blue}}} and {\color{reda}{\textbf{red}}}, respectively.}
 \begin{threeparttable}
   \resizebox{\linewidth}{!}{
    \setlength\tabcolsep{5pt}
    \renewcommand\arraystretch{0.9}
		\begin{tabular}{|lr||cccccccc|}
		\hline\thickhline
		 \rowcolor{mygray}
			\multicolumn{2}{|c||}{\multirow{5}{*}{Metric}} 
			&{UNet} &{UNet++} &{SFA} &{PraNet}&{SANet} & {MSNet} & {UACANet} &{GateNet}
			\\
			\multicolumn{2}{|l||}{} & ~\cite{Unet}&~\cite{UNet++}&~\cite{SFA_Polyp}&~\cite{PraNet_Polyp}&~\cite{SANet_Polyp}&~\cite{MSNet_Polyp}&~\cite{UACANet_Polyp} &--
			\\
	\multicolumn{2}{|l||}{}   & MICCAI&  TMI&  MICCAI& MICCAI& MICCAI& MICCAI& ACMMM& --
			\\
		\multicolumn{2}{|l||}{}   & 2015&  2019&  2019& 2020& 2021& 2021& 2021& --
			\\
			\multicolumn{2}{|l||}{}   & Res2-50&  Res2-50& Res2-50 & Res2-50& Res2-50&  Res2-50& Res2-50 &Res2-50
			\\
			\hline
			\hline
			\multirow{10}{*}{\emph{\rotatebox{90}{Endoscene~\cite{Endoscene}}}}      
			&$PA\uparrow$   
&0.979 
&0.984 
&0.936 
&\color{myblue}\textbf{0.990} 
&\color{myblue}\textbf{0.993} 
&\color{myblue}\textbf{0.990} 
&\color{reda}\textbf{0.995} 
&\color{myblue}\textbf{0.993}

			\\
			&$F_{\beta}^{max}\uparrow$    
&0.805 
&0.817 
&0.558 
&\color{myblue}\textbf{0.905} 
&0.881 
&\color{myblue}\textbf{0.899} 
&\color{myblue}\textbf{0.889} 
&\color{reda}\textbf{0.924} 

			\\
			&$F_{\beta}^{mean}\uparrow$ 
&0.703 
&0.706 
&0.353 
&0.824 
&0.823 
&\color{myblue}\textbf{0.829} 
&\color{reda}\textbf{0.885} 
&\color{myblue}\textbf{0.867}

			\\
			&$F_{\beta}^{\omega}\uparrow$	
&0.684 
&0.687 
&0.341 
&0.843 
&\color{myblue}\textbf{0.859} 
&0.848 
&\color{reda}\textbf{0.886} 
&\color{myblue}\textbf{0.885} 

			\\
			&$S_m\uparrow$ 		
&0.842 
&0.838 
&0.640 
&0.924 
&\color{myblue}\textbf{0.927} 
&0.926 
&\color{myblue}\textbf{0.933} 
&\color{reda}\textbf{0.941}

			\\
				&$E_m\uparrow$ 	
&0.867 
&0.884 
&0.604 
&0.938 
&\color{myblue}\textbf{0.948} 
&0.942 
&\color{reda}\textbf{0.976} 
&\color{myblue}\textbf{0.961}

			\\
			&$IOU\uparrow$
&0.639 
&0.636 
&0.332 
&0.804 
&\color{myblue}\textbf{0.829} 
&0.808 
&\color{myblue}\textbf{0.840} 
&\color{reda}\textbf{0.843}

			\\
			&$Dice\uparrow$ 	
&0.717 
&0.714 
&0.465 
&0.873 
&\color{myblue}\textbf{0.891} 
&0.869 
&\color{reda}\textbf{0.906} 
&\color{myblue}\textbf{0.903} 

			\\
			&$BER\downarrow$	
&0.121 
&0.137 
&0.084 
&0.033 
&\color{reda}\textbf{0.020} 
&0.037 
&\color{myblue}\textbf{0.023} 
&\color{myblue}\textbf{0.027} 

			\\
			&$\mathcal{M}\downarrow$
&0.022 
&0.018 
&0.065 
&0.010 
&\color{myblue}\textbf{0.008} 
&0.010 
&\color{reda}\textbf{0.006} 
&\color{myblue}\textbf{0.007} 

			\\
\hline
		\multirow{10}{*}{\emph{\rotatebox{90}{CVC-ColonDB~\cite{CVC-ColonDB}}}}    
		&$PA\uparrow$   
&0.942 
&0.938 
&0.907 
&\color{myblue}\textbf{0.965} 
&0.958 
&0.959 
&\color{myblue}\textbf{0.966} 
&\color{reda}\textbf{0.968}

			\\
			&$F_{\beta}^{max}\uparrow$    
&0.625 
&0.622 
&0.565 
&0.765 
&\color{myblue}\textbf{0.808} 
&0.807 
&\color{reda}\textbf{0.836} 
&\color{myblue}\textbf{0.824} 

			\\
			&$F_{\beta}^{mean}\uparrow$ 
&0.569 
&0.560 
&0.407 
&0.718 
&0.731 
&\color{myblue}\textbf{0.759} 
&\color{reda}\textbf{0.798} 
&\color{myblue}\textbf{0.767}

			\\
			&$F_{\beta}^{\omega}\uparrow$	
&0.498 
&0.467 
&0.379 
&0.699 
&0.726 
&\color{myblue}\textbf{0.736} 
&\color{reda}\textbf{0.772} 
&\color{myblue}\textbf{0.752} 

			\\
			&$S_m\uparrow$ 		
&0.711 
&0.691 
&0.634 
&\color{myblue}\textbf{0.820} 
&\color{myblue}\textbf{0.836} 
&\color{myblue}\textbf{0.836} 
&\color{reda}\textbf{0.846} 
&\color{reda}\textbf{0.846}

			\\
				&$E_m\uparrow$ 	
&0.763 
&0.762 
&0.648 
&0.847 
&0.855 
&\color{myblue}\textbf{0.883} 
&\color{reda}\textbf{0.897} 
&\color{myblue}\textbf{0.896}

			\\
			&$IOU\uparrow$ 	
&0.449 
&0.413 
&0.351 
&0.645 
&\color{myblue}\textbf{0.678} 
&\color{myblue}\textbf{0.678} 
&\color{reda}\textbf{0.707} 
&\color{myblue}\textbf{0.700}

			\\
			&$Dice\uparrow$ 	
&0.519 
&0.490 
&0.467 
&0.716 
&0.754 
&\color{myblue}\textbf{0.755} 
&\color{reda}\textbf{0.786} 
&\color{myblue}\textbf{0.771} 

			\\
			&$BER\downarrow$	
&0.242 
&0.258 
&0.183 
&0.141 
&\color{myblue}\textbf{0.106} 
&0.118 
&\color{reda}\textbf{0.105} 
&\color{myblue}\textbf{0.107} 

			\\
			&$\mathcal{M}\downarrow$
&0.061 
&0.064 
&0.094 
&0.043 
&0.043 
&\color{myblue}\textbf{0.041} 
&\color{myblue}\textbf{0.034} 
&\color{reda}\textbf{0.033} 
 
			\\
			\hline
					\multirow{10}{*}{\emph{\rotatebox{90}{CVC-ClinicDB~\cite{CVC-ClinicDB}}}}      
			&$PA\uparrow$   
&0.982 
&0.979 
&0.960 
&0.991 
&0.989 
&\color{myblue}\textbf{0.993} 
&\color{myblue}\textbf{0.992} 
&\color{reda}\textbf{0.994}

			\\
			&$F_{\beta}^{max}\uparrow$    
&0.880 
&0.858 
&0.776 
&\color{myblue}\textbf{0.927} 
&0.924 
&\color{myblue}\textbf{0.940} 
&0.926 
&\color{reda}\textbf{0.958} 

			\\
			&$F_{\beta}^{mean}\uparrow$ 
&0.804 
&0.784 
&0.655 
&0.885 
&0.883 
&\color{myblue}\textbf{0.894} 
&\color{myblue}\textbf{0.919} 
&\color{reda}\textbf{0.920}

			\\
			&$F_{\beta}^{\omega}\uparrow$	
&0.811 
&0.785 
&0.647 
&0.896 
&0.909 
&\color{myblue}\textbf{0.913} 
&\color{myblue}\textbf{0.917} 
&\color{reda}\textbf{0.937} 

			\\
			&$S_m\uparrow$ 	
&0.889 
&0.872 
&0.793 
&0.935 
&0.935 
&\color{myblue}\textbf{0.942} 
&\color{myblue}\textbf{0.938} 
&\color{reda}\textbf{0.953}

			\\
				&$E_m\uparrow$ 	
&0.917 
&0.898 
&0.816 
&0.958 
&0.963 
&\color{myblue}\textbf{0.971} 
&\color{myblue}\textbf{0.968} 
&\color{reda}\textbf{0.982}

			\\
			&$IOU\uparrow$ 	
&0.767 
&0.741 
&0.615 
&0.858 
&0.867 
&\color{myblue}\textbf{0.880} 
&\color{myblue}\textbf{0.873} 
&\color{reda}\textbf{0.902}

			\\
			&$Dice\uparrow$ 	
&0.824 
&0.797 
&0.698 
&0.902 
&0.918 
&\color{myblue}\textbf{0.921} 
&\color{myblue}\textbf{0.919} 
&\color{reda}\textbf{0.943} 
 
			\\
			&$BER\downarrow$	
&0.085 
&0.105 
&0.108 
&0.047 
&\color{myblue}\textbf{0.031} 
&0.034 
&\color{myblue}\textbf{0.032} 
&\color{reda}\textbf{0.024} 

			\\
			&$\mathcal{M}\downarrow$
&0.019 
&0.022 
&0.042 
&\color{myblue}\textbf{0.009} 
&0.012 
&\color{myblue}\textbf{0.008} 
&\color{myblue}\textbf{0.008} 
&\color{reda}\textbf{0.006} 
 
			\\
			\hline
					\multirow{10}{*}{\emph{\rotatebox{90}{ETIS~\cite{ETIS}}}}      
			&$PA\uparrow$   
&0.968 
&0.971 
&0.892 
&\color{myblue}\textbf{0.983} 
&\color{myblue}\textbf{0.986} 
&0.982 
&0.977 
&\color{reda}\textbf{0.988}

			\\
			&$F_{\beta}^{max}\uparrow$    
&0.497 
&0.554 
&0.367 
&0.675 
&\color{myblue}\textbf{0.748} 
&\color{myblue}\textbf{0.764} 
&0.686 
&\color{reda}\textbf{0.772} 

			\\
			&$F_{\beta}^{mean}\uparrow$ 
&0.394 
&0.465 
&0.255 
&0.602 
&\color{myblue}\textbf{0.656} 
&0.653 
&\color{myblue}\textbf{0.668} 
&\color{reda}\textbf{0.697}

			\\
			&$F_{\beta}^{\omega}\uparrow$	
&0.366 
&0.390 
&0.231 
&0.600 
&\color{myblue}\textbf{0.685} 
&\color{myblue}\textbf{0.677} 
&0.650 
&\color{reda}\textbf{0.702} 

			\\
			&$S_m\uparrow$ 		
&0.682 
&0.681 
&0.557 
&0.791 
&\color{reda}\textbf{0.843} 
&\color{myblue}\textbf{0.840} 
&0.812 
&\color{myblue}\textbf{0.842}

			\\
				&$E_m\uparrow$ 	
&0.645 
&0.704 
&0.515 
&0.792 
&\color{myblue}\textbf{0.835} 
&0.828 
&\color{myblue}\textbf{0.851} 
&\color{reda}\textbf{0.880}

			\\
			&$IOU\uparrow$ 	
&0.343 
&0.342 
&0.219 
&0.576 
&\color{reda}\textbf{0.670} 
&\color{myblue}\textbf{0.666} 
&0.618 
&\color{myblue}\textbf{0.669}

			\\
			&$Dice\uparrow$ 	
&0.406 
&0.413 
&0.297 
&0.630 
&\color{myblue}\textbf{0.751} 
&\color{myblue}\textbf{0.719} 
&0.696 
&\color{myblue}\textbf{0.735} 

			\\
			&$BER\downarrow$	
&0.273 
&0.308 
&0.232 
&0.169 
&\color{reda}\textbf{0.059} 
&\color{myblue}\textbf{0.108} 
&\color{myblue}\textbf{0.094} 
&0.111 
 
			\\
			&$\mathcal{M}\downarrow$
&0.036 
&0.035 
&0.109 
&0.031 
&\color{myblue}\textbf{0.015} 
&\color{myblue}\textbf{0.020} 
&0.023 
&\color{reda}\textbf{0.012} 

			\\
	\hline
						\multirow{10}{*}{\emph{\rotatebox{90}{Kvasir~\cite{Kvasir}}}}      
			&$PA\uparrow$   
&0.947 
&0.954 
&0.926 
&0.971 
&\color{myblue}\textbf{0.972} 
&\color{myblue}\textbf{0.974} 
&\color{myblue}\textbf{0.974} 
&\color{reda}\textbf{0.976}

			\\
			&$F_{\beta}^{max}\uparrow$    
&0.876 
&0.904 
&0.801 
&0.929 
&\color{myblue}\textbf{0.931} 
&\color{reda}\textbf{0.938} 
&0.922 
&\color{myblue}\textbf{0.937} 

			\\
			&$F_{\beta}^{mean}\uparrow$ 
&0.832 
&0.853 
&0.715 
&0.897 
&\color{myblue}\textbf{0.903} 
&0.902 
&\color{myblue}\textbf{0.914} 
&\color{reda}\textbf{0.916}

			\\
			&$F_{\beta}^{\omega}\uparrow$	
&0.794 
&0.808 
&0.670 
&0.885 
&\color{myblue}\textbf{0.892} 
&\color{myblue}\textbf{0.892} 
&\color{myblue}\textbf{0.897} 
&\color{reda}\textbf{0.903} 

			\\
			&$S_m\uparrow$ 		
&0.858 
&0.862 
&0.782 
&\color{myblue}\textbf{0.915} 
&0.914 
&\color{reda}\textbf{0.923} 
&0.914 
&\color{myblue}\textbf{0.921}

			\\
				&$E_m\uparrow$ 	
&0.901 
&0.907 
&0.828 
&0.943 
&\color{myblue}\textbf{0.950} 
&0.944 
&\color{myblue}\textbf{0.951} 
&\color{reda}\textbf{0.958}

			\\
			&$IOU\uparrow$ 	
&0.756 
&0.753 
&0.619 
&0.848 
&0.853 
&\color{myblue}\textbf{0.862} 
&\color{myblue}\textbf{0.855} 
&\color{reda}\textbf{0.864}

			\\
			&$Dice\uparrow$ 
&0.821 
&0.824 
&0.725 
&0.901 
&\color{myblue}\textbf{0.907} 
&\color{myblue}\textbf{0.907} 
&\color{myblue}\textbf{0.908} 
&\color{reda}\textbf{0.912} 

			\\
			&$BER\downarrow$	
&0.087 
&0.100 
&0.113 
&\color{myblue}\textbf{0.052} 
&\color{myblue}\textbf{0.050} 
&\color{reda}\textbf{0.049} 
&\color{myblue}\textbf{0.050} 
&\color{myblue}\textbf{0.052} 

			\\
			&$\mathcal{M}\downarrow$
&0.055 
&0.048 
&0.075 
&0.030 
&\color{myblue}\textbf{0.028} 
&\color{myblue}\textbf{0.028} 
&\color{myblue}\textbf{0.026} 
&\color{reda}\textbf{0.024} 

			\\
			\hline\hline
	\rowcolor{mygray}		&$Top$ $3$  
&0/50
&0/50
&0/50
&9/50
&31/50
&34/50
&39/50
&49/50
\\
	\rowcolor{mygray}		&$Top$ $1$ 
&0/50
&0/50
&0/50
&0/50
&4/50
&3/50
&13/50
&30/50
\\
	\hline
		\end{tabular}
	}
	\setlength{\abovecaptionskip}{2pt}
	\label{tab:Polyp_performance}
	\end{threeparttable}
\end{table}
\begin{table}[!t]
	\small
\centering
	\setlength{\abovecaptionskip}{2pt}
	\caption{Quantitative comparison of different RGB-D SOD methods. Top $3$ and Top $1$ scores are highlighted in {\color{myblue}{\textbf{blue}}} and {\color{reda}{\textbf{red}}}, respectively.}
 \begin{threeparttable}
   \resizebox{0.98\linewidth}{!}{
    \setlength\tabcolsep{5pt}
    \renewcommand\arraystretch{1}
\begin{tabular}{|lr||cccccccc|}
		\hline\thickhline
		 \rowcolor{mygray}
			\multicolumn{2}{|c||}{\multirow{5}{*}{Metric}} 
			&{DCF} &{RD3D} &{UTA} &{DSNet}& {SPNet} &   {TriTrans} & {GateNet} &{GateNet}
			\\
		
			\multicolumn{2}{|l||}{}   & ~\cite{DCF_RGBDSOD}&~\cite{RD3D_RGBDSOD}&~\cite{UTA_RGBDSOD}&~\cite{DSNet_RGBDSOD}&~\cite{SPNet_RGBDSOD}&~\cite{TriTransNet_RGBDSOD} & &
			\\
		\multicolumn{2}{|l||}{}   & CVPR&  AAAI&  TIP& TIP& ICCV& ACMMM& --& --
			\\
		\multicolumn{2}{|l||}{}   & 2021&  2021&  2021& 2021& 2021& 2021& --& --
			\\
			\multicolumn{2}{|l||}{}   & Res-50& Res-50 & Res-50& Res-50&  Res2-50& Res-50 & Res-50&Res2-50
			\\
			\hline
			\hline
			\multirow{10}{*}{\emph{\rotatebox{90}{DUTLF-D~\cite{DMRA_RGBDSOD}}}}      
			&$PA\uparrow$   
& -
&0.973 
& -
& -
& -
&\color{myblue}\textbf{0.975} 
&\color{reda} \textbf{0.977} 
&\color{reda} \textbf{0.977}

			\\
					&$F_{\beta}^{max}\uparrow$   
& -
&0.946 
& -
& -
& -
&\color{myblue}\textbf{0.951} 
&\color{reda}\textbf{0.958}
&\color{myblue}\textbf{0.957}

			\\
				&$F_{\beta}^{mean}\uparrow$ 
& -
&0.924 
& -
& -
& -
&\color{myblue}\textbf{0.938} 
&\color{reda}\textbf{0.946} 
&\color{myblue}\textbf{0.944}

			\\
\hline
		\end{tabular}
	}
	\setlength{\abovecaptionskip}{2pt}
	\label{tab:RGBDSOD_performance}
	\end{threeparttable}
\end{table}
\begin{table*}[htbp]
  \centering
  \caption{Quantitative comparison of different camouflaged object methods. The best scores are highlighted in {\color{reda}{\textbf{red}}}.}
  \begin{threeparttable}
   \resizebox{0.85\linewidth}{!}{
    \setlength\tabcolsep{5pt}
    \renewcommand\arraystretch{1.05}
         \begin{tabular}{|c|r|c|c||c|c|c|c|c|c|c|c|c|c|}
     \hline\thickhline
     \rowcolor{mygray}
     Dataset&Method  & Pub. & Backbone& $PA\uparrow$ 	& $F_{\beta}^{max}\uparrow$ & $F_{\beta}^{mean}\uparrow$ & $F_{\beta}^{\omega}\uparrow$ & $S_m\uparrow$ & $E_m\uparrow$  &$IOU\uparrow$&$Dice\uparrow$  &$BER\downarrow$ &$\mathcal{M}\downarrow$ 
     \\
     				\hline
				\hline
	
      \multirow{6}{*}{CAMO~\cite{CAMO}}  &SINet~\cite{SINet_COD}
 &CVPR 2020&ResNet-50&	0.907 &	0.762 	&	0.709 &	0.606 &	0.751 &	0.835 &	0.546 &	0.639 &	0.201 &	0.100 
 
 \\
 &PFNet~\cite{PFNet_COD}&CVPR 2021 &	ResNet-50 &	0.917 	&	0.795 &	0.751 	&	0.695 &	0.782 	&	0.855 &	0.624 &	0.725 	&	0.150 &	0.085 
 
 \\
 &RankNet~\cite{Rank-Net_COD}
 &CVPR 2021&ResNet-50&	0.922 &	0.791 &	0.756 	&	0.696 	&0.787 	&	0.859 	&	0.626 &	0.722 &	0.154 	&	0.080 
 
 \\
 &MGL~\cite{MGL_COD}
 &CVPR 2021&ResNet-50&	0.914 &	0.792 	&	0.738 	&	0.673 &	0.775 	&	0.848 	&	0.605 &	0.699 &	0.172 	&	0.088

 \\
 &UGTR~\cite{UGTR_COD}
 &ICCV 2021&ResNet-50&	0.918 	&	0.800 	&	0.748 &	0.684 &0.784 &	0.858 &	0.618 	&	0.712 &	0.161 	&	0.086 
 
 \\
 &GateNet
 &--&ResNet-50& \color{reda} \textbf{0.934} &  \color{reda} \textbf{0.835} 	&	 \color{reda} \textbf{0.804} &	 \color{reda} \textbf{0.756} & \color{reda} \textbf{0.829} &	 \color{reda} \textbf{0.888} 	&	 \color{reda} \textbf{0.694} &	 \color{reda} \textbf{0.781}&	 \color{reda} \textbf{0.122} & \color{reda} \textbf{0.069} 
 \\
 \hline
 \multirow{6}{*}{CHAMELEON~\cite{CHAMELEON}}  &SINet~\cite{SINet_COD}
 &CVPR 2020&ResNet-50&	0.965 	&	0.846 	&	0.776 	&	0.740 &	0.869 &	0.899 	&	0.726 &	0.776 &	0.107 &	0.044 
 
 \\
\hline
\end{tabular}
}
\end{threeparttable}
\label{tab:COD_performance}
\end{table*}
\begin{table*}[htbp]
  \centering
  \caption{Quantitative comparison of different defocus blur detction methods. The best scores are highlighted in {\color{reda}{\textbf{red}}}.}
  \begin{threeparttable}
   \resizebox{0.85\linewidth}{!}{
    \setlength\tabcolsep{5pt}
    \renewcommand\arraystretch{1.05}
   \begin{tabular}{|c|r|c|c||c|c|c|c|c|c|c|c|c|c|}
     \hline\thickhline
     \rowcolor{mygray}
     Dataset&Method  & Pub. & Backbone& $PA\uparrow$ 	& $F_{\beta}^{max}\uparrow$ & $F_{\beta}^{mean}\uparrow$ & $F_{\beta}^{\omega}\uparrow$ & $S_m\uparrow$ & $E_m\uparrow$  &$IOU\uparrow$&$Dice\uparrow$  &$BER\downarrow$ &$\mathcal{M}\downarrow$ 
     \\
     				\hline
				\hline
     \multirow{6}{*}{CUHK~\cite{CUHK}}  &CENet~\cite{CENet_DBD}
     &CVPR 2019&VGG-16&	0.942	&	0.914&	0.873&	0.867&	0.873&	0.894&	0.804&	0.877	&	0.081	&	0.060

\\
     &Depth-Distill~\cite{Depth-Distill_DBD}&ECCV 2020&VGG-16 &	0.955	&	0.924	&	0.848	&	0.881	&	0.891	&0.898	&0.845&	0.904	&	0.052	&	0.049
	 \\
&SG~\cite{SG_DBD}
     &CVPR 2021&VGG-16&0.881	&	0.820	&	0.641&	0.737&0.762	&	0.749&	0.673&	0.788&	0.123	&	0.123
\\
&IS2CNet~\cite{IS2CNet_DBD}
     &TCSVT 2021&VGG-16&	0.939 &	0.917 	&	0.899 	&	0.862 &	0.863 &	0.909 &	0.799 &	0.872 	&	0.085 	&	0.063 
\\
&DENets~\cite{DENets_DBD}
     &TIP 2021&VGG-16&	0.953 &0.931 &	0.881 &	0.882 	&0.887 &0.909 	&	0.837 &	0.890 &	0.068&	0.055 
     \\
&GateNet
     &--&VGG-16&\color{reda} \textbf{0.963} &\color{reda} \textbf{0.935} 	&\color{reda} \textbf{0.920} &\color{reda} \textbf{0.903} &	\color{reda} \textbf{0.906} &\color{reda} \textbf{0.941} 	&\color{reda} \textbf{0.869} &\color{reda} \textbf{0.919} 	&	\color{reda} \textbf{0.046} &\color{reda} \textbf{0.040} 

\\
				\hline
     \multirow{6}{*}{DUT~\cite{BTBNet_DBD}}  &CENet~\cite{CENet_DBD}
     &CVPR 2019&VGG-16&	0.867 
	&	0.826 
&	0.767 
&	0.697 
&	0.742 
&	0.775 
&	0.609 
&	0.703 
	&	0.185 
	&	0.136

\\
\hline
    \end{tabular}
   }
  \end{threeparttable}
 	\label{tab:DBD_performance}
\end{table*}
\begin{table*}[htbp]
  \centering
  \caption{Quantitative comparison of different shadow detection methods. The best scores are highlighted in {\color{reda}{\textbf{red}}}.}
  \begin{threeparttable}
   \resizebox{0.85\linewidth}{!}{
    \setlength\tabcolsep{5pt}
    \renewcommand\arraystretch{1.05}
    \begin{tabular}{|c|r|c|c||c|c|c|c|c|c|c|c|c|c|}
     \hline\thickhline
     \rowcolor{mygray}
     Dataset&Method  & Pub. & Backbone& $PA\uparrow$ 	& $F_{\beta}^{max}\uparrow$ & $F_{\beta}^{mean}\uparrow$ & $F_{\beta}^{\omega}\uparrow$ & $S_m\uparrow$ & $E_m\uparrow$  &$IOU\uparrow$&$Dice\uparrow$  &$BER\downarrow$ &$\mathcal{M}\downarrow$ 
     \\
     				\hline
				\hline
     \multirow{5}{*}{SBU~\cite{SBU}}  &DSC~\cite{DSC_Shadow}
     &CVPR 2018&VGG-16&	0.969 &	0.914 &	0.892 &	0.861 &	0.856 &	0.939 	&	0.801 &	0.871 &	0.081 	&	0.032 
\\
     &ADNet~\cite{ADNet_Shadow}&ECCV 2018&--&	0.951 	&	0.877 	&	0.696 &	0.424 &	0.700 	&	0.811 	&	0.747 	&	0.539 &	0.074 	&	0.201 
	 \\
&BDRAR~\cite{BDRAR_Shadow}
     &ECCV 2018&ResNeXt-101&0.961 	&	0.884 &	0.830 	&	0.827 	&	0.844 &	0.930 	&	0.784 	&	0.863 &	0.052 	&	0.039 

\\
&DSD~\cite{DSDNet_Shadow}
     &CVPR 2019&ResNeXt-101&0.965 &	0.896 &	0.841&	0.835 	&	0.851 &0.933 &	0.797&	0.873 &	\color{reda} \textbf{0.046} 	&	0.036 
\\
&GateNet&--&ResNeXt-101&\color{reda} \textbf{0.978} &\color{reda} \textbf{0.937} &	\color{reda} \textbf{0.903} &	\color{reda} \textbf{0.889} &	\color{reda} \textbf{0.886} 	&	\color{reda} \textbf{0.957} &	\color{reda} \textbf{0.848} &	\color{reda} \textbf{0.899}&	{0.059} &	\color{reda} \textbf{0.025} 

\\
				\hline
    \multirow{5}{*}{UCF~\cite{UCF}}  &DSC~\cite{DSC_Shadow}
     &CVPR 2018&VGG-16&	0.947 	&	0.806 &0.772 &	0.737 	&	0.788 &0.897 &	0.675 &	0.770 &	0.121 &	0.054 

\\
     &ADNet~\cite{ADNet_Shadow}&ECCV 2018&--&	0.916 &	0.783 	&	0.575 	&	0.371 &	0.647 &	0.748 &	0.611 &	0.483 	&	0.110 &	0.229 

	 \\
\hline
    \end{tabular}
   }
  \end{threeparttable}
 	\label{tab:Shadow_performance}
\end{table*}
\begin{table*}[htbp]
  \centering
  \caption{Quantitative comparison of different transparent, glass and mirror detection methods. The best scores are highlighted in {\color{reda}{\textbf{red}}}.}
  \begin{threeparttable}
   \resizebox{0.85\linewidth}{!}{
    \setlength\tabcolsep{5pt}
    \renewcommand\arraystretch{1.05}
       \begin{tabular}{|c|r|c|c||c|c|c|c|c|c|c|c|c|c|}
     \hline\thickhline
     \rowcolor{mygray}
     Dataset&Method  & Pub. & Backbone& $PA\uparrow$ 	& $F_{\beta}^{max}\uparrow$ & $F_{\beta}^{mean}\uparrow$ & $F_{\beta}^{\omega}\uparrow$ & $S_m\uparrow$ & $E_m\uparrow$  &$IOU\uparrow$&$Dice\uparrow$  &$BER\downarrow$ &$\mathcal{M}\downarrow$ 
     \\
     				\hline
				\hline
		\multicolumn{14}{|c|}{\multirow{1}{*}{\textbf{Transparent Object Detection}}}
			\\
					\hline
						\hline
     \multirow{2}{*}{Transparent-Easy~\cite{TransLab_Transparent}}  &Translab~\cite{TransLab_Transparent}
     &ECCV 2020&ResNet-50&0.978&0.955&0.953&0.941&0.935&0.974	&0.921 &0.957 &0.027&0.022
\\
     &GateNet& -- &	ResNet-50 &	\textcolor{red}{\textbf{0.989}}	&	 \color{reda} \textbf{0.980} &	 \color{reda} \textbf{{0.979}} &	 \color{reda} \textbf{{0.974}} &	 \color{reda} \textbf{{0.963}} &	 \color{reda} \textbf{{0.988}} &	 \color{reda} \textbf{{0.962}}&	 \color{reda} \textbf{{0.980}}&	 \color{reda} \textbf{{0.014}}	&	 \color{reda} \textbf{{0.011}} 
 \\	
     \hline
      \multirow{2}{*}{Transparent-Hard~\cite{TransLab_Transparent}}  
&Translab~\cite{TransLab_Transparent}&ECCV 2020 &ResNet-50& 	0.913	&	0.843 &	0.827 	&	0.783 	&	0.798 	&0.871 	 &	0.733&	0.827 &0.110&	0.087 	 
\\
     &GateNet& -- &	ResNet-50 &	 \color{reda} \textbf{{0.947}}	&	 \color{reda} \textbf{{0.913}} 	&	 \color{reda} \textbf{{0.904}} &	 \color{reda} \textbf{{0.874}} &	 \color{reda} \textbf{{0.871}} &	 \color{reda} \textbf{{0.923}} &	 \color{reda} \textbf{{0.838}}&	 \color{reda} \textbf{{0.899}} 	&	 \color{reda} \textbf{{0.069}} 	&	 \color{reda} \textbf{{0.053}} 	
     \\
 \hline
   \multirow{2}{*}{Transparent-All~\cite{TransLab_Transparent}} 
&Translab~\cite{TransLab_Transparent}&ECCV 2020 &ResNet-50&	0.964&	0.931 	&	0.927 &	0.907 	&	0.906 	&	0.952 		&	0.881 &	0.929 &	0.044	  &	0.036 
\\
\hline
    \end{tabular}
   }
  \end{threeparttable}
	\label{tab:Glass_like_performance}
\end{table*}
\begin{table*}[htbp]
  \centering
  \caption{Quantitative comparison of different ORSI SOD methods. The best scores are highlighted in {\color{reda}{\textbf{red}}}.}
  \begin{threeparttable}
   \resizebox{\linewidth}{!}{
    \setlength\tabcolsep{5pt}
    \renewcommand\arraystretch{1}
    \begin{tabular}{|c|r|c|c||c|c|c|c|c|c|c|c|c|c|}
     \hline\thickhline
     \rowcolor{mygray}
     Dataset&Method  & Pub. & Backbone& $PA\uparrow$ 	& $F_{\beta}^{max}\uparrow$ & $F_{\beta}^{mean}\uparrow$ & $F_{\beta}^{\omega}\uparrow$ & $S_m\uparrow$ & $E_m\uparrow$  &$IOU\uparrow$&$Dice\uparrow$  &$BER\downarrow$ &$\mathcal{M}\downarrow$ 
     \\
     				\hline
				\hline
     \multirow{3}{*}{ORSSD~\cite{LV-Net_RSISOD}} 
     &DAFNet~\cite{DAFNet_RSISOD}
     &TIP 2020&VGG-16&	\color{reda} \textbf{0.990} &	0.903 &	0.788 &	0.844 &	0.912 &	0.920 	&	0.823 &	0.874 &	0.059 &	\color{reda} \textbf{0.011}

\\
     &MJRBM~\cite{MJRBM_RSISOD}&TGRS 2021&VGG-16&	0.986 &	0.893 	&	0.802 &	0.844 &	0.910 &	0.934 &	0.817 	&0.853 	&	0.066 &	0.016 
\\
&GateNet&--&VGG-16&\color{reda} \textbf{0.990} &	\color{reda} \textbf{0.914} &	\color{reda} \textbf{0.847} &	\color{reda} \textbf{0.875} 	&\color{reda} \textbf{0.925} 	&	\color{reda} \textbf{0.959} &	\color{reda} \textbf{0.839} &	\color{reda} \textbf{0.885} &	\color{reda} \textbf{0.048} 	&	\color{reda} \textbf{0.011} 
\\
				\hline
    \multirow{4}{*}{EORSSD~\cite{DAFNet_RSISOD}} 
       &DAFNet~\cite{DAFNet_RSISOD}
     &TIP 2020&VGG-16&\color{reda} \textbf{0.996} &	0.867 &	0.642 &	0.783 	&	0.883 &0.815 	&	0.800 &	0.830 &	\color{reda} \textbf{0.051} &	\color{reda} \textbf{0.006} 
\\
     &MJRBM~\cite{MJRBM_RSISOD}&TGRS 2021&VGG-16&	0.992 &	0.877 	&	0.707 &	0.813 &	0.879 &0.890 	&	0.793 	&0.822 &	0.070 &	0.010 
\\
     &RRNet~\cite{RRNet_RSISOD}&TGRS 2021&Res2Net-50&	0.994 	&	0.887 	&	0.725 	&	0.827 	&0.885 	&	0.873 	&	0.834 	&0.862 &	0.067 	&	0.008 
\\
&GateNet&--&VGG-16&{0.993} &	\color{reda} \textbf{0.896} &	\color{reda} \textbf{0.799} &	\color{reda} \textbf{0.859} 	&\color{reda} \textbf{0.894} 	&	\color{reda} \textbf{0.915} 	&	\color{reda} \textbf{0.850} 	&	\color{reda} \textbf{0.892} &	{0.057} &{0.008} 
\\
\hline
    \end{tabular}
   }
  \end{threeparttable}
	\label{tab:ORSISOD_performance}
\end{table*}

\begin{table*}
\centering
	\scriptsize
	\setlength{\abovecaptionskip}{2pt}
	\caption{Efficiency comparisons of the top-performing methods in Tab.~\ref{tab:RGBSOD_performance} - Tab.~\ref{tab:ORSISOD_performance}. The best and worst results are shown in {\color{reda}\textbf{red}} and {\color{myblue}\textbf{blue}}, respectively.}
 \begin{threeparttable}
   \resizebox{\linewidth}{!}{
    \setlength\tabcolsep{2pt}
    \renewcommand\arraystretch{1}
\begin{tabular}{|r||ccc|ccccc|cc|ccccc|cc|}
	\hline 
	 \rowcolor{mygray}
		Model Nmae  &{TriTrans$^{\dagger}$} &{SPNet$^{\dagger}$} &{GateNet$^{\dagger}$} & {EBLNet} &{GDNet}  &{MirrorNet}&{DSDNet}&{GateNet}&{UACANet}&{GateNet}&{UGTR}&{MGL}& {Translab}&{CTDNet}&{GateNet}&{MJRBM}&{GateNet}\\
		\rowcolor{mygray}
		&~\cite{TriTransNet_RGBDSOD}  &~\cite{SPNet_RGBDSOD} &&~\cite{EBLNet_Glass}&~\cite{GDNet_Glass} & ~\cite{MirrorNet_Mirror} & ~\cite{DSDNet_Shadow} &&~\cite{UACANet_Polyp} & &~\cite{UGTR_COD}&~\cite{MGL_COD}&~\cite{TransLab_Transparent}&~\cite{CTDNet}&&~\cite{MJRBM_RSISOD}& \\
		\rowcolor{mygray}
		Backbone  &{Res-50} &{Res2-50} &{Res-50} & {ResX-101} &{ResX-101}  &{ResX-101}&{ResX-101}&{ResX-101}&{Res2-50}&{Res2-50}&{Res-50}&{Res-50}& {Res-50}&{Res-50}&{Res-50}&{VGG-16}&{VGG-16} \\
		\hline
		\hline
		Model Size (MB)$\downarrow$   &559  & \color{myblue}\textbf{702}&\color{reda}\textbf{254}  &\color{myblue}\textbf{845} & 770 & 465 &414&\color{reda}\textbf{373}&\color{myblue}\textbf{278}&\color{reda}\textbf{211}&296& \color{myblue}\textbf{435}&162&\color{reda}\textbf{94}&119&\color{myblue}\textbf{175}&\color{reda}\textbf{78} 

		\\
		Parameters (MB)$\downarrow$    &139.55   & \color{myblue}\textbf{175.29} &\color{reda}\textbf{81.55} & 111.45 & \color{myblue}\textbf{201.72} & 121.77 &\color{reda}\textbf{58.16}&92.89&\color{myblue}\textbf{69.16}&\color{reda}\textbf{52.51}&48.87& \color{myblue}\textbf{63.60}&42.20&\color{reda}\textbf{21.93}&29.68&\color{myblue}\textbf{43.78}&\color{reda}\textbf{20.43}
		\\
		FLOPs (G)$\downarrow$     &\color{myblue}\textbf{680.07} & 135.86 &\color{reda}\textbf{63.30} & \color{myblue}\textbf{674.41} & 244.21 & 111.92 &\color{reda}\textbf{54.34}&94.18&\color{myblue}\textbf{119.60}&\color{reda}\textbf{74.16}&358.53&\color{myblue}\textbf{475.40}&284.68&\color{reda}\textbf{21.43}&64.24&\color{myblue}\textbf{191.46}&\color{reda}\textbf{131.01}
		\\
\hline
	\end{tabular}

	}
	\setlength{\abovecaptionskip}{2pt}
	\label{tab:modelsize_para_flops}
	\end{threeparttable}
\end{table*}

\begin{table}[htbp]
  \centering
  \caption{Ablation  experiments  for seven binary segmentation tasks. 
  M1: FPN Baseline. 
  M2: + Residual Parallel Branch. 
  M3: + Gate Units v1.
  M4: + Gate Units v2.
  M5: + Fold-ASPP.}
  \begin{threeparttable}
   \resizebox{\linewidth}{!}{
    \setlength\tabcolsep{5pt}
    \renewcommand\arraystretch{1.05}

\begin{tabular}{|c|c||c|c|c|c|c|c|c|c|c|c|>{\columncolor{mygray}}l|}
  \hline\thickhline
  \rowcolor{mygray}
  Dataset                                        & Method & $PA\uparrow$ & $F_{\beta}^{max}\uparrow$ & $F_{\beta}^{mean}\uparrow$ & $F_{\beta}^{\omega}\uparrow$ & $S_m\uparrow$ & $E_m\uparrow$ & $IOU\uparrow$ & $Dice\uparrow$ & $BER\downarrow$ & $\mathcal{M}\downarrow$ & $\Delta$ gains              \\
  \hline
  \hline
  \multirow{6}{*}{DUTS~\cite{DUTS}}              & M1     & 0.958        & 0.870                     & 0.801                      & 0.800                        & 0.870         & 0.888         & 0.763         & 0.820          & 0.102           & 0.045                   &                   \\
                                                 & M2     & 0.962        & 0.875                     & 0.810                      & 0.808                        & 0.875         & 0.894         & 0.775         & 0.830          & 0.095           & 0.043                   & $\uparrow$1.85\%  \\
                                                 & M3     & 0.967        & 0.888                     & 0.835                      & 0.830                        & 0.885         & 0.906         & 0.795         & 0.844          & 0.082           & 0.038                   & $\uparrow$5.70\%  \\
                                                 & M4     & 0.970        & 0.895                     & 0.842                      & 0.836                        & 0.892         & 0.914         & 0.807         & 0.856          & 0.071           & 0.035                   & $\uparrow$8.20\%  \\
                                                 & M5     & 0.972        & 0.911                     & 0.857                      & 0.864                        & 0.906         & 0.931         & 0.828         & 0.878          & 0.052           & 0.030                   & $\uparrow$12.91\% \\
\hline
\end{tabular}

   }
  \end{threeparttable}
	\label{tab:Ablation_study_RGBSOD}
\end{table}
 
\begin{table}[htbp]
  \centering
  \caption{Ablation  experiments  for RGB-D salient object detection. 
  M1: FPN Baseline. 
  M2: + Residual Parallel Branch. 
  M3: + Cross-modal Gate Units.
  M4: + Encoder-Decoder Gate Units. 
  M5: + Fold-ASPP.}
  \begin{threeparttable}
   \resizebox{\linewidth}{!}{
    \setlength\tabcolsep{5pt}
    \renewcommand\arraystretch{1.05}

\begin{tabular}{|c|c||c|c|c|c|c|c|c|c|c|c|>{\columncolor{mygray}}l|}
  \hline \thickhline
  \rowcolor{mygray}
  Dataset                                   & Method & $PA\uparrow$ & $F_{\beta}^{max}\uparrow$ & $F_{\beta}^{mean}\uparrow$ & $F_{\beta}^{\omega}\uparrow$ & $S_m\uparrow$ & $E_m\uparrow$ & $IOU\uparrow$ & $Dice\uparrow$ & $BER\downarrow$ & $\mathcal{M}\downarrow$ & $\Delta$ gains              \\
  \hline \hline
  \multirow{6}{*}{STERE~\cite{STERE}}       & M1     & 0.944        & 0.882                     & 0.853                      & 0.802                        & 0.860         & 0.874         & 0.817         & 0.833          & 0.057           & 0.048                   &                   \\
                                            & M2     & 0.949        & 0.888                     & 0.871                      & 0.825                        & 0.870         & 0.890         & 0.825         & 0.845          & 0.054           & 0.045                   & $\uparrow$2.31\%  \\
                                            & M3     & 0.958        & 0.905                     & 0.885                      & 0.847                        & 0.892         & 0.914         & 0.838         & 0.867          & 0.050           & 0.041                   & $\uparrow$5.53\%  \\
                                            & M4     & 0.968        & 0.920                     & 0.897                      & 0.869                        & 0.910         & 0.934         & 0.850         & 0.888          & 0.046           & 0.036                   & $\uparrow$8.80\%  \\
                                            & M5     & 0.970        & 0.929                     & 0.907                      & 0.889                        & 0.921         & 0.952         & 0.862         & 0.909          & 0.043           & 0.03                    & $\uparrow$11.42\% \\
\hline
\end{tabular}

   }
  \end{threeparttable}
  \label{tab:Ablation_study_RGBDSOD}
\end{table}
 
\begin{table}
	\small
\centering
	\setlength{\abovecaptionskip}{2pt}
	
	\caption{Evaluation of the folded atrous convolution. (x) stands for different sampling rates of atrous convolution. D-ASPP is DenseASPP~\cite{DenseASPP}.}
 \begin{threeparttable}
   \resizebox{\linewidth}{!}{
    \setlength\tabcolsep{5pt}
    \renewcommand\arraystretch{1}
		\begin{tabular}{|lr||ccc|ccc|cc|cc|}
  \hline \thickhline
  \rowcolor{mygray}
  \multicolumn{2}{|r||}{\multirow{1}{*}{\small{Metric}}}                 & {Atrous}                     & {Atrous} & {Atrous}         & {Fold}           & {Fold}           & {Fold}           & {ASPP}             & {Fold-}          & {D-ASPP}         & {Fold-}                              \\
  \rowcolor{mygray}                                                      &                              & {r=2}    & {r=4}            & {r=6}            & {r=2}            & {r=4}            & {r=6}              &                  & {ASPP}           & {}                & {D-ASPP}         \\
  \hline \hline
  \multirow{10}{*}{\emph{\rotatebox{90}{DUTS~\cite{DUTS}}}}              & $PA\uparrow$                 & 0.970    & 0.970            & 0.971            & 0.971            & 0.971            & 0.971              & 0.972            & 0.972            & 0.971             & 0.971            \\
                                                                         & $F_{\beta}^{max}\uparrow$    & 0.894    & 0.895            & 0.898            & 0.898            & 0.902            & 0.906              & 0.903            & 0.911            & 0.902             & 0.909            \\
                                                                         & $F_{\beta}^{mean}\uparrow$   & 0.843    & 0.845            & 0.848            & 0.847            & 0.851            & 0.853              & 0.852            & 0.857            & 0.854             & 0.861            \\
                                                                         & $F_{\beta}^{\omega}\uparrow$ & 0.837    & 0.841            & 0.843            & 0.841            & 0.845            & 0.852              & 0.850            & 0.864            & 0.844             & 0.858            \\
\hline
\end{tabular}

	}
	\setlength{\abovecaptionskip}{2pt}
	\label{tab:Ablation_study_Fold}
	\end{threeparttable}
\end{table}
\noindent$\bullet$ In Tab.~\ref{tab:RGBSOD_performance}, among $50$ scores of all \textit{\textbf{RGB SOD}} datasets, our GateNet achieves significant performance improvement compared to the second best method CTDNet~\cite{CTDNet} in terms of top 3 ($49/50$ vs. $37/50$) and top 1 ($40/50$ vs. $1/50$), respectively.
And, we still consistently outperform the VST~\cite{T2T} model even if it is equipped with a stronger transformer backbone T2T~\cite{T2T}. 
\\
\noindent$\bullet$ Tab.~\ref{tab:Polyp_performance} shows performance comparisons on five \textit{\textbf{polyp segmentation}} datasets. 
Our GateNet  consistently outperforms the second best approach UACANet~\cite{UACANet_Polyp} under top 3 ($49/50$ vs. $39/50$) and top 1 ($30/50$ vs. $13/50$), respectively. 
In particular, GateNet achieves a predominant performance on the CVC-ClinicDB~\cite{CVC-ClinicDB} in terms of all ten metrics.
\\
\noindent$\bullet$ For fair comparison with other \textit{\textbf{RGB-D SOD}} methods, we show the performance of GateNet with ResNet-50 and Res2Net-50 as backbone. In Tab.~\ref{tab:RGBDSOD_performance}, we can see that GateNet-Res-50 and GateNet-Res2-50 achieve the top 1 performance of $43/80$ and $32/80$  while the TriTransNet~\cite{TriTransNet_RGBDSOD} and SPNet~\cite{SPNet_RGBDSOD} only reach $11/80$ and $10/70$, respectively. Further, the comparison of GateNet-Res-50 + GateNet-Res2-50 and the TriTransNet +  SPNet is $62/80$ vs. $19/80$. 
\\
\noindent$\bullet$ Tab.~\ref{tab:COD_performance} - Tab.~\ref{tab:ORSISOD_performance} show performance comparisons with \textit{\textbf{camouflaged, defcus blur, shadow, transparent, glass, mirror and ORSI SOD}} methods, respectively. Without too much claim, our models achieve the best performance in terms of all ten metrics across $16$ out of  $17$ different datasets.
\\
\noindent$\bullet$. Tab.~\ref{tab:modelsize_para_flops} lists the \textit{\textbf{model sizes, parameters, FLOPs and speed}} of different methods with superior performance in Tab.~\ref{tab:RGBSOD_performance} - Tab.~\ref{tab:ORSISOD_performance} in detail. It can be seen that both two-stream and single stream GateNets still have obvious advantages against most state-of-the-art methods with different backbones. 
\begin{figure*}[!t]
	\subfigure[\scriptsize{RGB Salient Object Detection}]{
		\includegraphics[width=6.5cm]{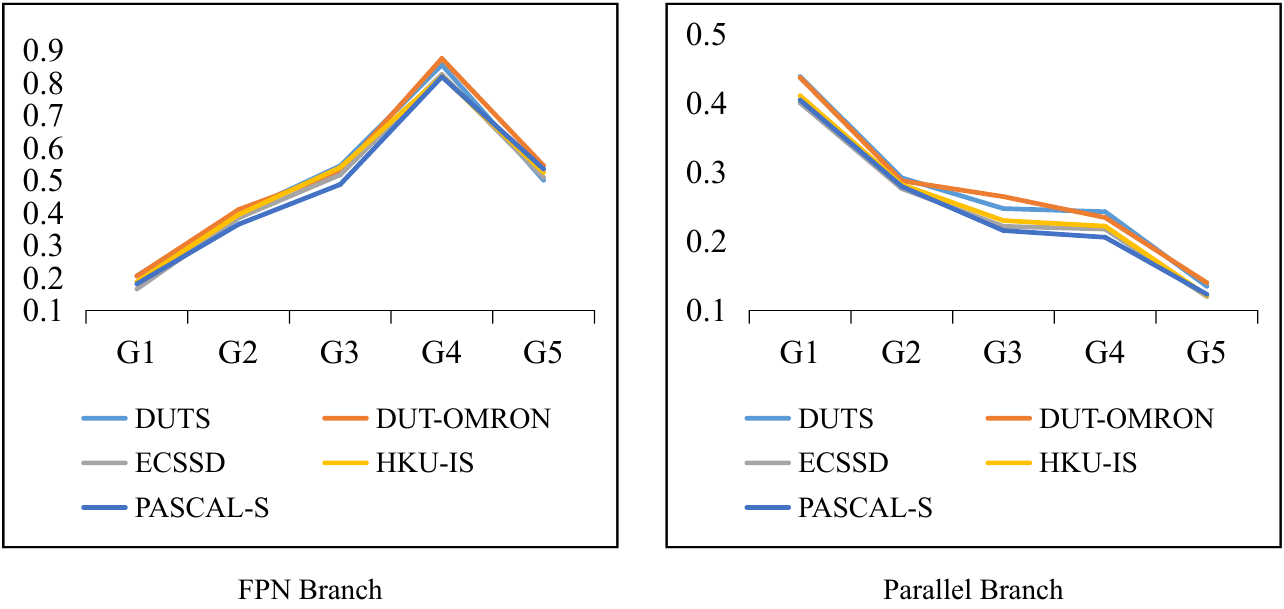}
	}	
	\subfigure[\scriptsize{RGB-D Salient Object Detection}]{
		\includegraphics[width=6.5cm]{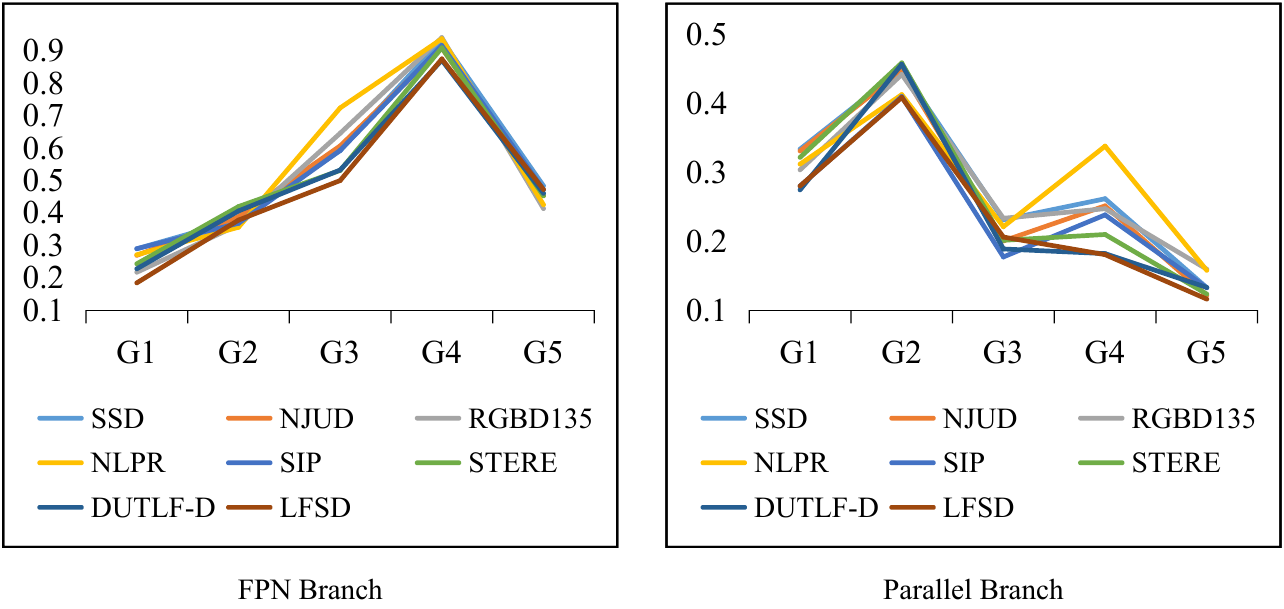}
	}	
	\subfigure[\scriptsize{ORSI Salient Object Detection}]{
		\includegraphics[width=6.5cm]{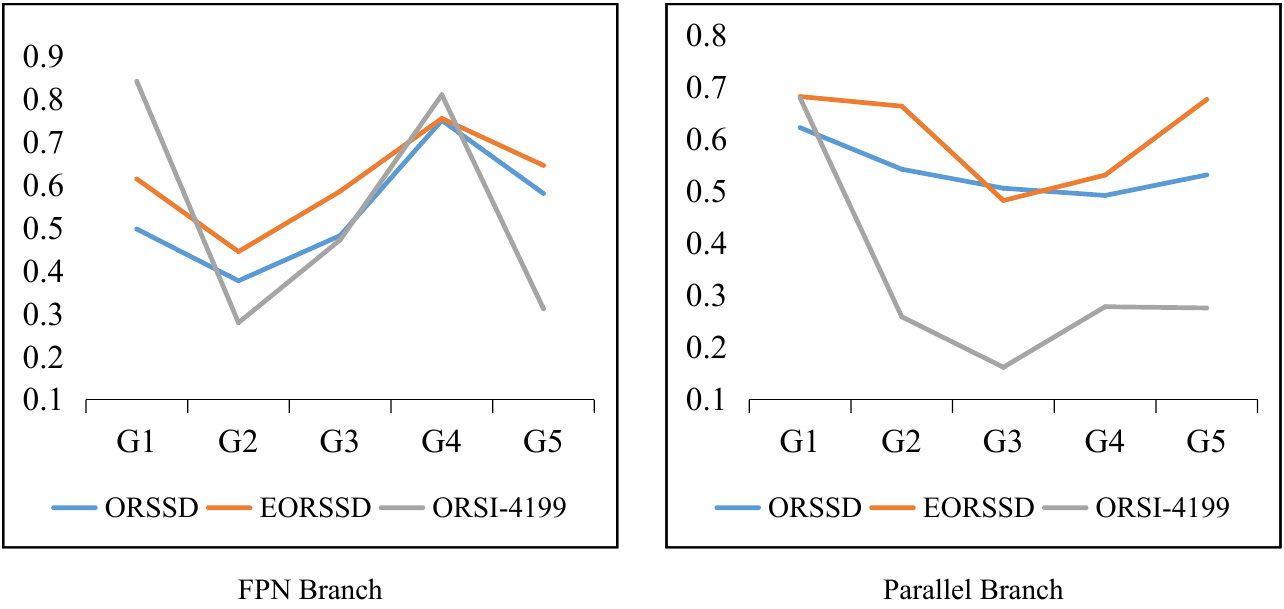}
	}	
	\subfigure[\scriptsize{Camouflaged Object Detection}]{
		\includegraphics[width=6.5cm]{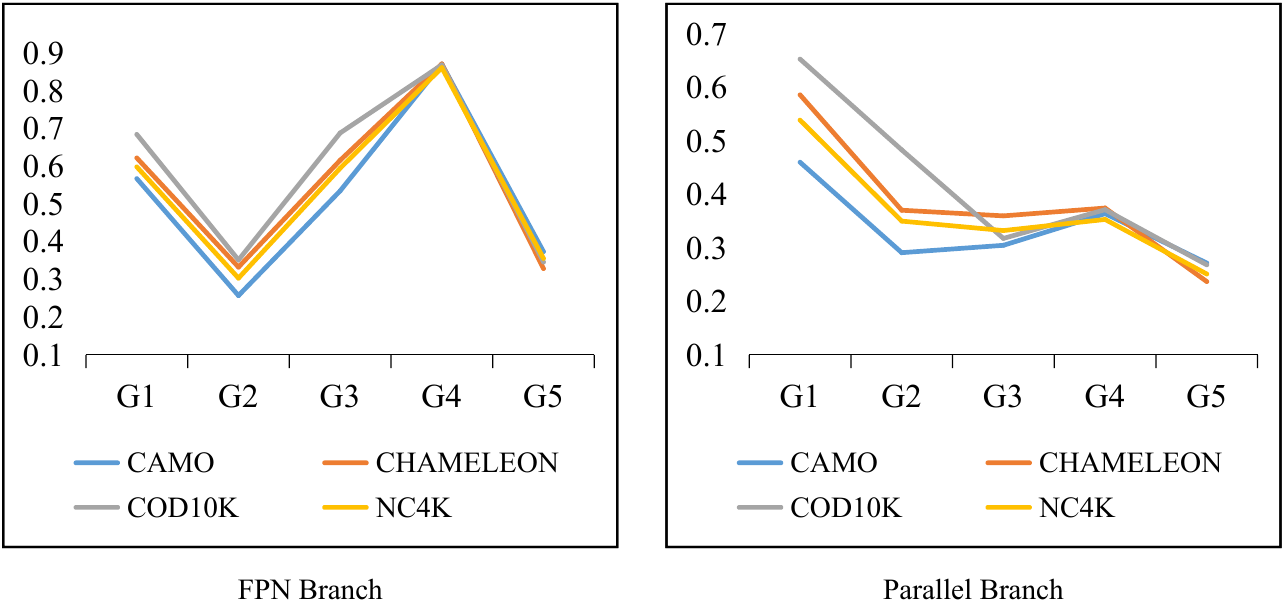}
	}	
	\subfigure[\scriptsize{Defocus Blur Detection}]{
		\includegraphics[width=6.5cm]{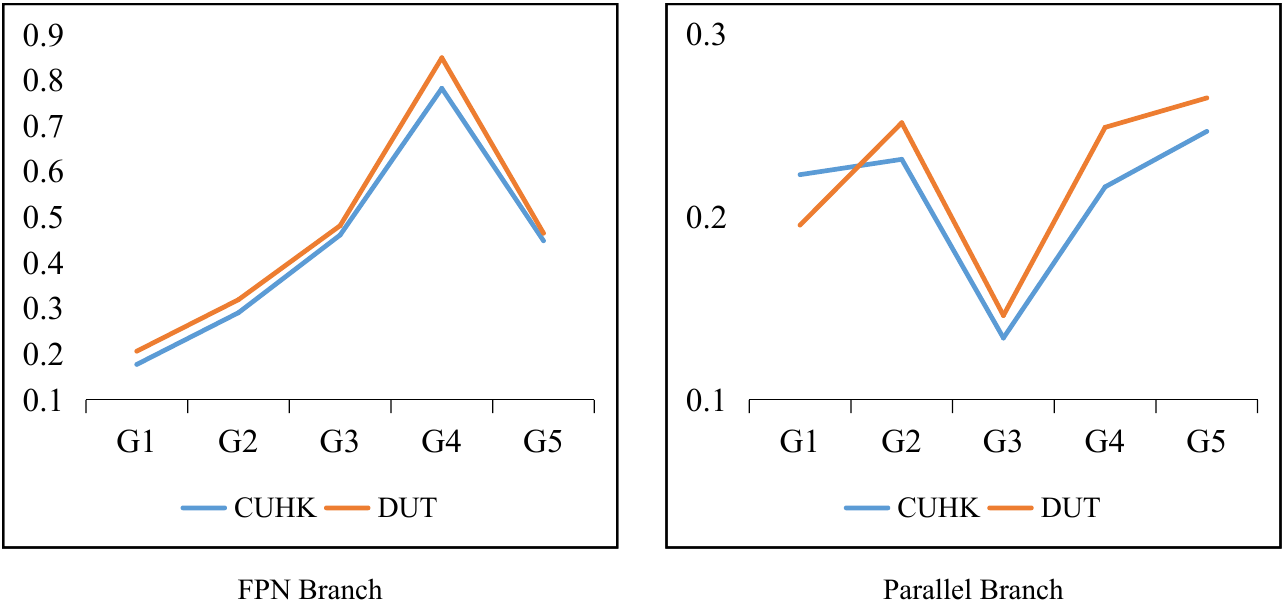}
	}	
	\subfigure[\scriptsize{Shadow Detection}]{
		\includegraphics[width=6.5cm]{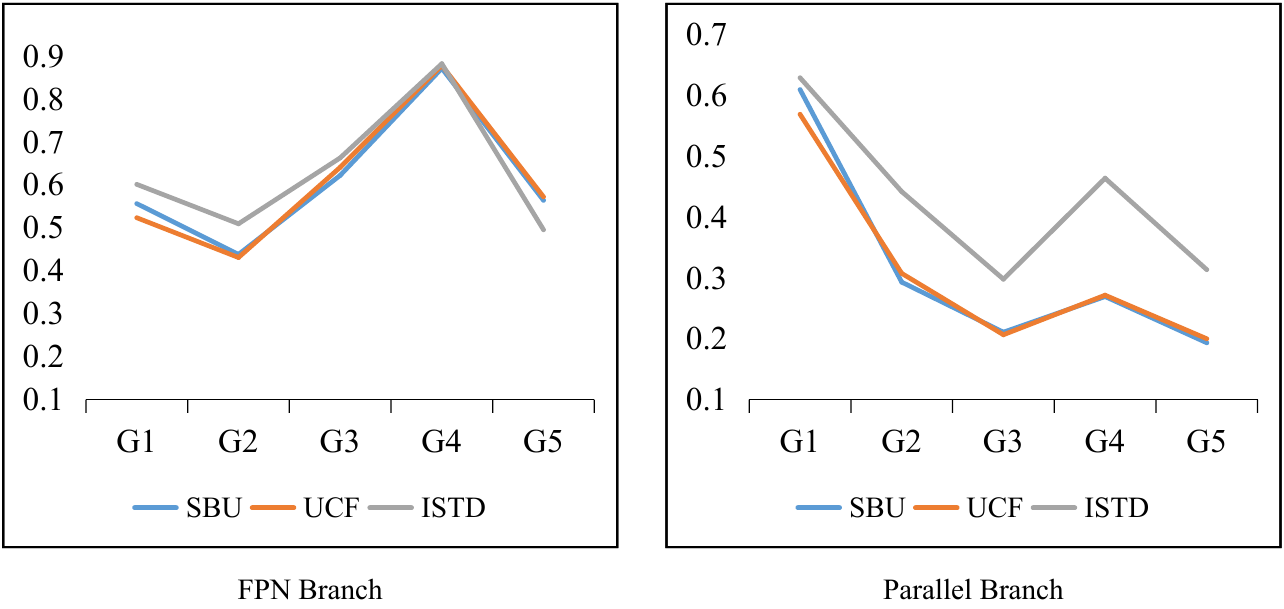}
	}	
	\subfigure[\scriptsize{Transparent, Glass, Mirror Object Detection}]{
		\includegraphics[width=6.5cm]{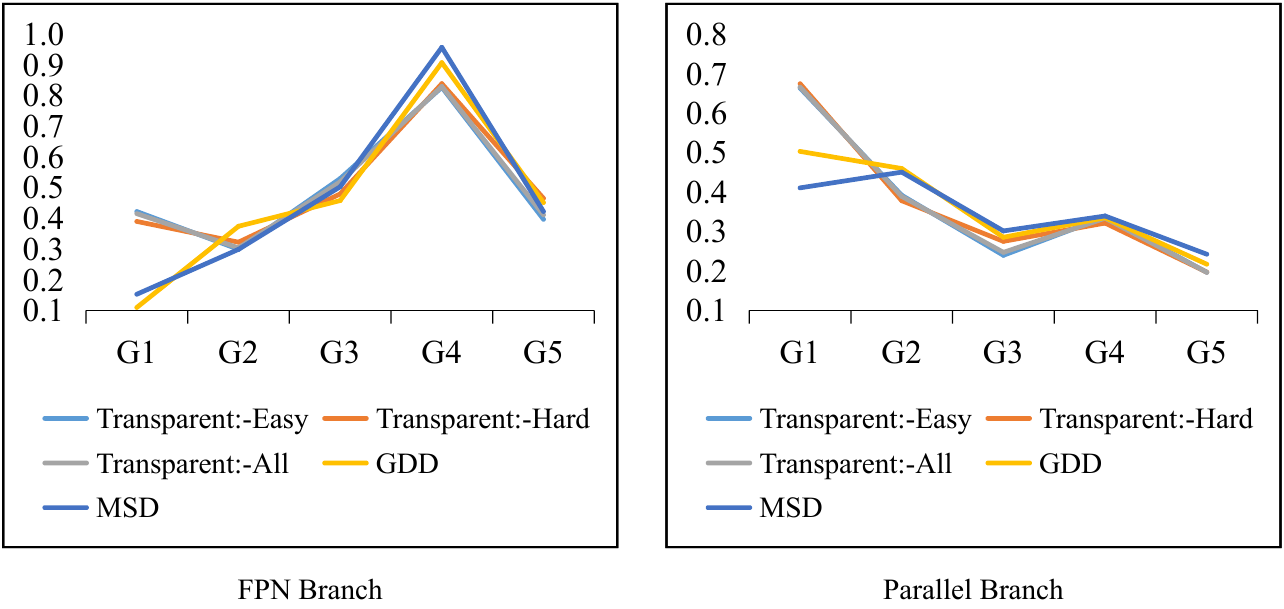}
	}	
	\subfigure[\scriptsize{Polyp Segmentation}]{
		\includegraphics[width=6.5cm]{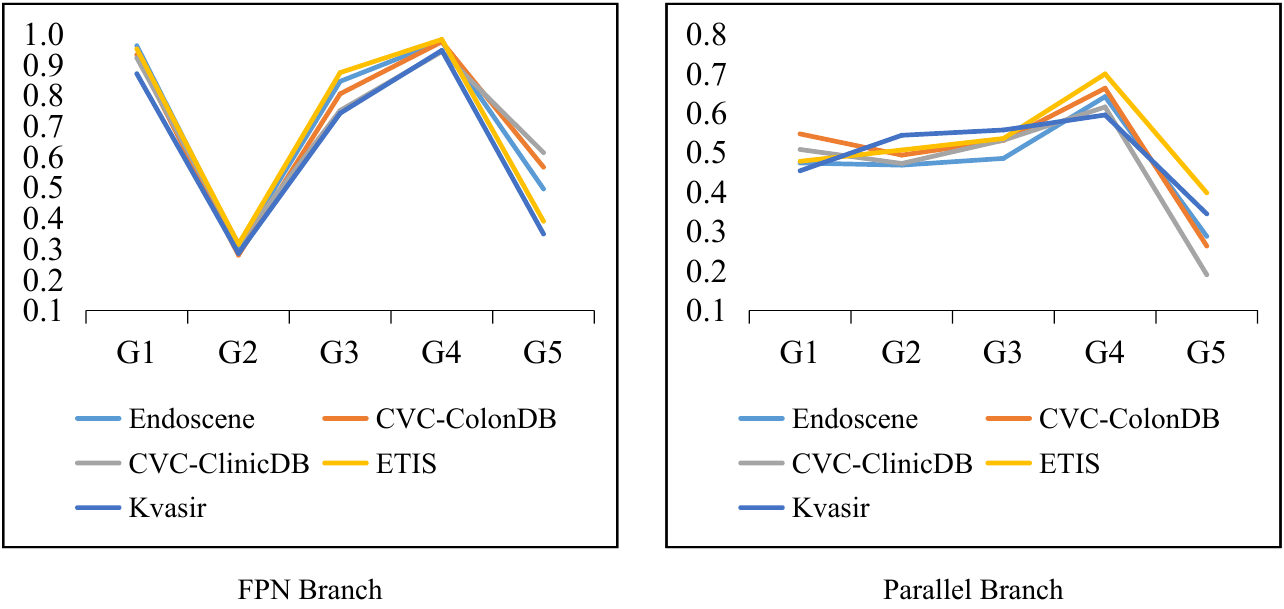}
	}
	\centering
	\caption{\small Distributions of gate weights separately presented in the FPN and parallel branch on $35$ datasets of $10$ tasks.  }
	\label{Fig:gatevalue}	
\end{figure*}
\subsection{Ablation Studies}
To reflect the general contribution of each component to the overall performance, we conduct ablation studies on the largest dataset for each sub-task individually. Tab.~\ref{tab:Ablation_study_RGBSOD} and Tab.~\ref{tab:Ablation_study_RGBDSOD} are the results for single-input tasks and the two-input task (RGB-D SOD), respectively. Tab.~\ref{tab:Ablation_study_Fold} verifies the effect of folded atrous convolution thoroughly.  
\\
\noindent\textbullet~\textbf{{Dual Branch Decoder.}} 
The baseline (M$1$) is a FPN structure with a progressive decoder. 
We add the residual  parallel branch to construct the dual branch decoder. 
We can see that M$2$ consistently outperforms M$1$ across all datasets in terms of all ten metrics.
Meanwhile, M$2$ has been able to surpass SINet~\cite{SINet_COD}, PFNet~\cite{PFNet_COD}, IS2CNet~\cite{IS2CNet_DBD} and BDRAR~\cite{BDRAR_Shadow}.
Based on this strong dual branch network, the subsequent performance gain of gate units and fold atrous convolution is more convincing.
\\
\noindent\textbullet~\textbf{{Gate Units.}} 
We embed multi-level gate units in both the FPN and parallel branches. 
In Tab.~\ref{tab:Ablation_study_RGBSOD}, the M$3$ achieves a significant improvement compared to the M$4$ indicates the necessity of designing gate units-v2 with a global information perspective. 
Further, the performance gap between M$2$ and M$4$ shows that the dual branch gated network obtains a considerable performance gain. 
In Tab.~\ref{tab:Ablation_study_RGBDSOD}, M$3$ vs. M$2$ and  M$4$ vs. M$3$ demonstrate the effectiveness of gate units in cross-modal fusion and encoder-decoder feature transition, respectively.
In addition, the curves of gate value on each dataset in ten tasks as shown in Fig.~\ref{Fig:gatevalue}. From  these gated patterns, we reveal some insightful findings:
\textbf{\uppercase\expandafter{\romannumeral1})}  
For the distribution of gate values at all levels in the FPN branch, Fig.~\ref{Fig:gatevalue}(a), (b), (e), (f), (g) present G1 and G2 are smaller than G3, G4, G5, while G1 in Fig.~\ref{Fig:gatevalue}(c), (d), (h) has the opposite trend. 
Analyzed from the visual perception, camouflaged objects, orsi object and polyps are easy to be confused with the background. The boundary information is very important to distinguish the fore/background, which drives the network to pay more attention on low-level features.
\textbf{\uppercase\expandafter{\romannumeral2})}  
For the distribution of gate values at all levels in the parallel branch, the greater contribution of G1 and G2 in Fig.~\ref{Fig:gatevalue}(d), (h) compared to the other tasks further illustrates the importance of details information in camouflaged and polyp segmentation tasks.
\textbf{\uppercase\expandafter{\romannumeral3})}
As shown in Fig.~\ref{Fig:gatevalue}(e), G4 and G5 have high values in both FPN branch and parallel branch, indicating that the accurate localization of focused regions is extremely crucial for defocus blur detection and motivate the network to consistently maintain a high pass-through pattern for high-level features.
\textbf{\uppercase\expandafter{\romannumeral4})}
Compared to other gate values in the FPN branch, G4 is the largest one and even exceeds $0.9$ for almost all tasks. 
This phenomenon is also in line with our general understanding for deep networks, i.e., level-4 features effectively can construct the main body of foreground because they not only contain stable semantic information but also have larger spatial resolution than level-5 features.
\textbf{\uppercase\expandafter{\romannumeral5})}
 Distributions of gate weights can well depict the similarities and differences among diverse binary segmentation sub-tasks.
\begin{figure}[!t]
	\begin{center}
		\includegraphics[width=\linewidth]{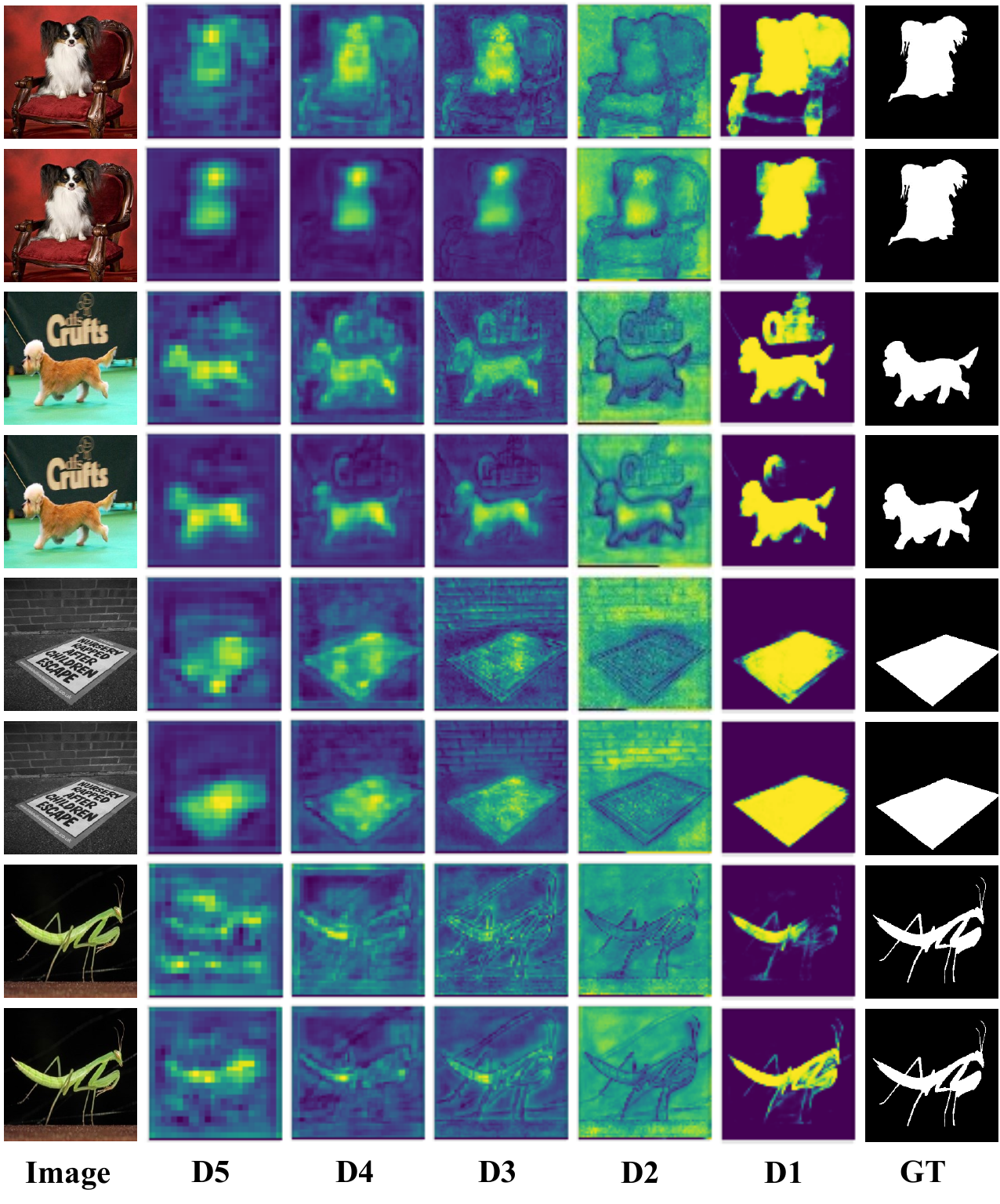}
	\end{center}
	\centering
	\caption{\small Visual comparison of feature maps for showing the effect of the multi-level gate units. D5 $\sim$ D1 represent the feature maps of each decoder block from high level to low level. Odd rows and even rows are the results of the FPN baseline without or with multi-level gate units, respectively.}\label{fig:Gate_suppress_visual_results}
\end{figure}
\begin{figure}[!t]
	\begin{center}
		\includegraphics[width=\linewidth]{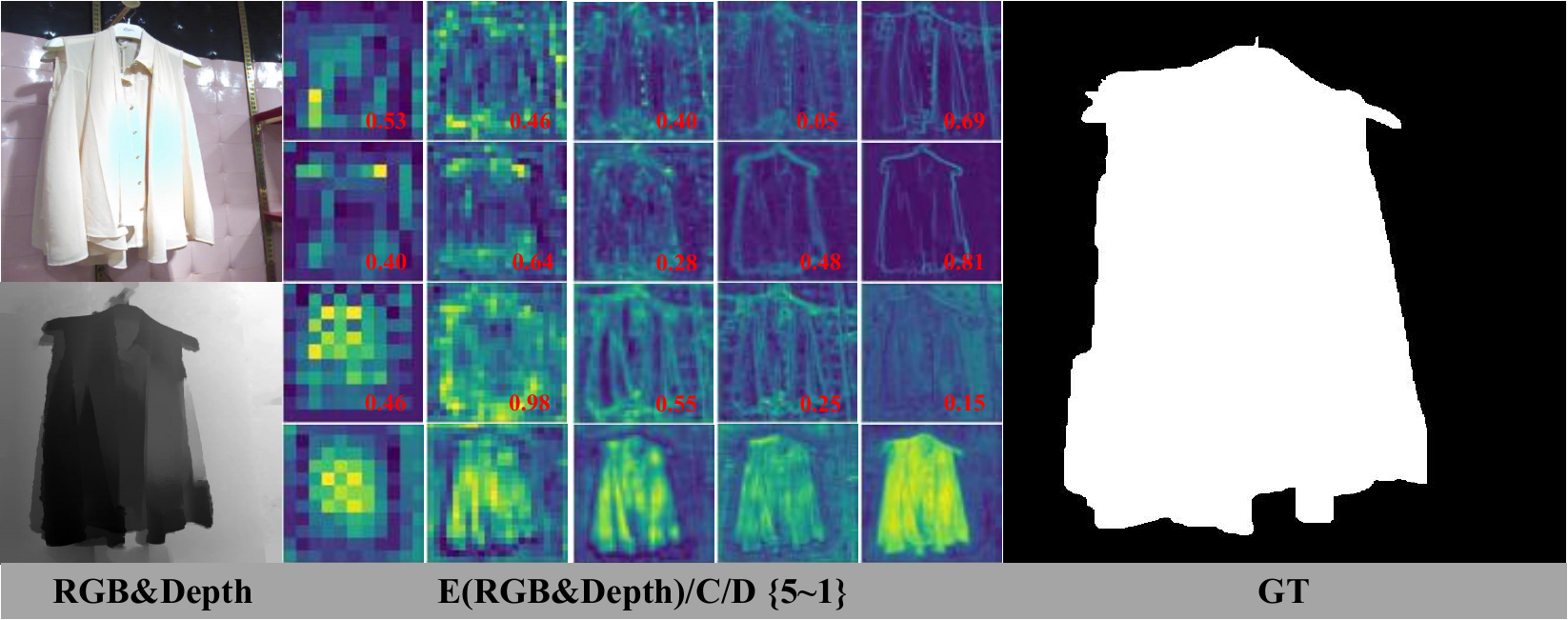}
	\end{center}
	\centering
	\caption{\small Visual results of feature maps. Each RGB-D input image corresponds to four rows of feature maps. The first two rows are RGB and Depth encoder feature maps (E5 - E1), respectively. The third row is the cross-modal fusion feature maps (C5 - C1). The last row is the decoder feature maps (D5 - D1). The naming of these feature maps is consistent with those in Fig.~\ref{fig:Twostream_GateNet}. }\label{fig:Gate_suppress_visual_results_rgbd}
\end{figure}

 To show the effect of the gate units more intuitively, we visualize the features of different levels in Fig.~\ref{fig:Gate_suppress_visual_results}. It can be observed that even if the dog has a very low contrast with the chair or the billboard (see the $1^{st}$ $\sim$ $4^{th}$ rows), through using multi-level gate units, the high contrast between the object region and the background is always maintained at each layer while the detailed information is continually regained, thereby making salient objects be effectively distinguished. 
 And, the gate units can avoid excessive suppression of the slender parts of objects (see the $5^{th}$ $\sim$ $8^{th}$ rows). The corners of the poster, the limbs and even tentacles of the mantis are retained well. 
 Besides, we show the visual results of the gate units in the two-stream network for RGB-D SOD, as shown in Fig.~\ref{fig:Gate_suppress_visual_results_rgbd}. 
 Intuitively, the depth branch has more significant and pure position and edge information about the foreground (cloth) than the RGB branch on E4, E2 and E1, thus distributes larger gate weights correspondingly in cross-modal fusion. 
\\
\noindent\textbullet~\textbf{{Folded Atrous Convolution.}} Based on the gated dual branch network, we design a series of experimental options to verify the effectiveness of the folded atrous convolution. Tab.~\ref{tab:Ablation_study_Fold} illustrates the results in detail. We adopt the atrous convolution with  dilation rates of $[2, 4, 6]$ and the same dilation rates are also applied to the folded atrous convolution. It can be observed that the folded atrous convolution consistently yields significant performance improvement at each dilation rate than the corresponding atrous convolution in terms of all ten metrics. 
And the single-layer Fold(6) already performs better than the ASPP and DenseASPP of aggregating three atrous convolution layers. The Fold-ASPP and Fold-DenseASPP  naturally outperforms the ASPP and DenseASPP, respectively. 
Our fold operation can naturally increase the receptive field. For a fair comparison, we can also see that compared with Atrous(4) with the same receptive field, Fold(2) still has an advantage under all metrics.
\begin{table*}[!t]
  \centering
  \caption{\textcolor{blue}{Accuracy and efficiency comparison with different transformer-based methods on the COD10K~\cite{SINet_COD} test set.  The best scores are highlighted in {\color{reda}{\textbf{red}}}.}}
  \begin{threeparttable}
   \resizebox{\linewidth}{!}{
    \setlength\tabcolsep{5pt}
    \renewcommand\arraystretch{1.05}
   \begin{tabular}{|r|c|c||c|c|c|c|c|c|c|}
     \hline\thickhline
     \rowcolor{mygray}
Method  &Publication&Backbone&Parameters $\downarrow$&Training Time $\downarrow$&Inference Speed $\uparrow$& Inference Memory $\downarrow$& $F_{\beta}^{\omega}\uparrow$ & $S_m\uparrow$ &$E_m\uparrow$ 
   \\
     		\hline
				\hline
OSformer~\cite{Osformer}&ECCV 2022&ResNet-50&46.6 MB&10 Hours&15 Fps&2.5 GB&0.685&0.813&0.893 \\
DTIT~\cite{DTIT}&ICPR 2022&MiT-B5&253.7 MB&23 Hours&8 Fps&4.5 GB&0.695&0.824&0.896 \\
HitNet~\cite{HitNet}&AAAI 2023&PvTv2-B2&24.4 MB&13 Hours&7 Fps&4.2 GB&0.806&0.868&0.936 \\
FSPNet~\cite{FSPNet_COD}&CVPR 2023&ViT-B&84.2 MB&64 Hours&19 Fps&3.2 GB&0.735&0.851&0.930 \\
\hline
GateNet&-&ResNet-50&29.68 MB&6 Hours&55 Fps&2.1 GB&0.742&0.846&0.901\\
GateNet&-&PvTv2-B2&\color{reda} \textbf{20.32 MB} &\color{reda} \textbf{5 Hours}&\color{reda} \textbf{58 Fps}&\color{reda} \textbf{1.9 GB}&0.813&0.876&0.942\\
GateNet&-&ViT-B&43.12 MB&8 Hours&43 Fps&2.7 GB&\color{reda} \textbf{0.828}&\color{reda} \textbf{0.888}&\color{reda} \textbf{0.947}\\
\hline
    \end{tabular}

   }
  \end{threeparttable}
 	\label{tab:transformer_methods_comparison}
\end{table*}

\begin{figure}
	\begin{center}
		\includegraphics[width=\linewidth]{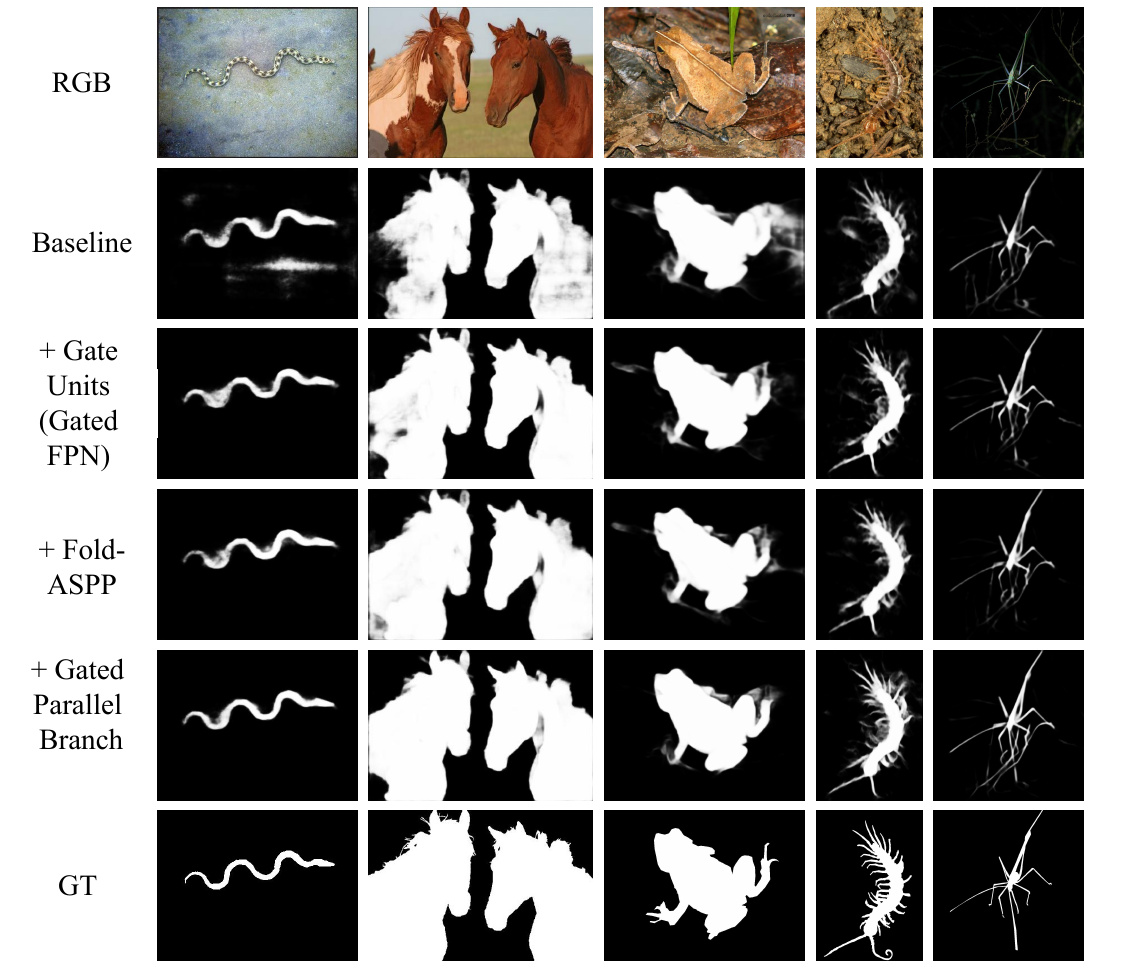}
	\end{center}
	\centering
	\caption{Illustration of the benefit of each component. 
	}\label{fig:ablation_study_visual_results}
\end{figure}

Fig.\ref{fig:ablation_study_visual_results} shows visual results of the above  ablation  studies  on some examples. It can be seen that the gated FPN model accurately determines where is the foreground object. 
With the help of Fold-ASPP, the overall integrity of the object is further captured. 
It should also be noted that the gated parallel branch can improve perceptual results greatly by highlighting the fore/back-ground difference and preserving the intra-class consistency, thereby yielding the sharpened boundary.

{
\subsection{Gate Unit Meets Transformer}
With the development of vision transformer~\cite{transformer}, 
recent binary works achieve good performance on many important benchmarks. In this section, we first analyse the advantages of GateNet compared to transformer-based methods in terms of accuracy and efficiency. Next, we quantitatively and qualitatively show the limitations of transformer-based methods in cross-branch prediction. Finally, we explore the positive impact of the gate unit on transformer. 
\\
\noindent\textbullet~\textbf{{Advantages in Accuracy and Efficiency.}}   
In Tab.~\ref{tab:transformer_methods_comparison}, we can see that the GateNet has obvious advantages in accuracy, model size, training time, inference speed and memory requirements. Compared to transformer-based architectures~\cite{Osformer,DTIT,HitNet,FSPNet_COD}, GateNet achieves a good balance between accuracy and efficiency.
\begin{figure}[!t]
	\centering
	\includegraphics[width=\linewidth]{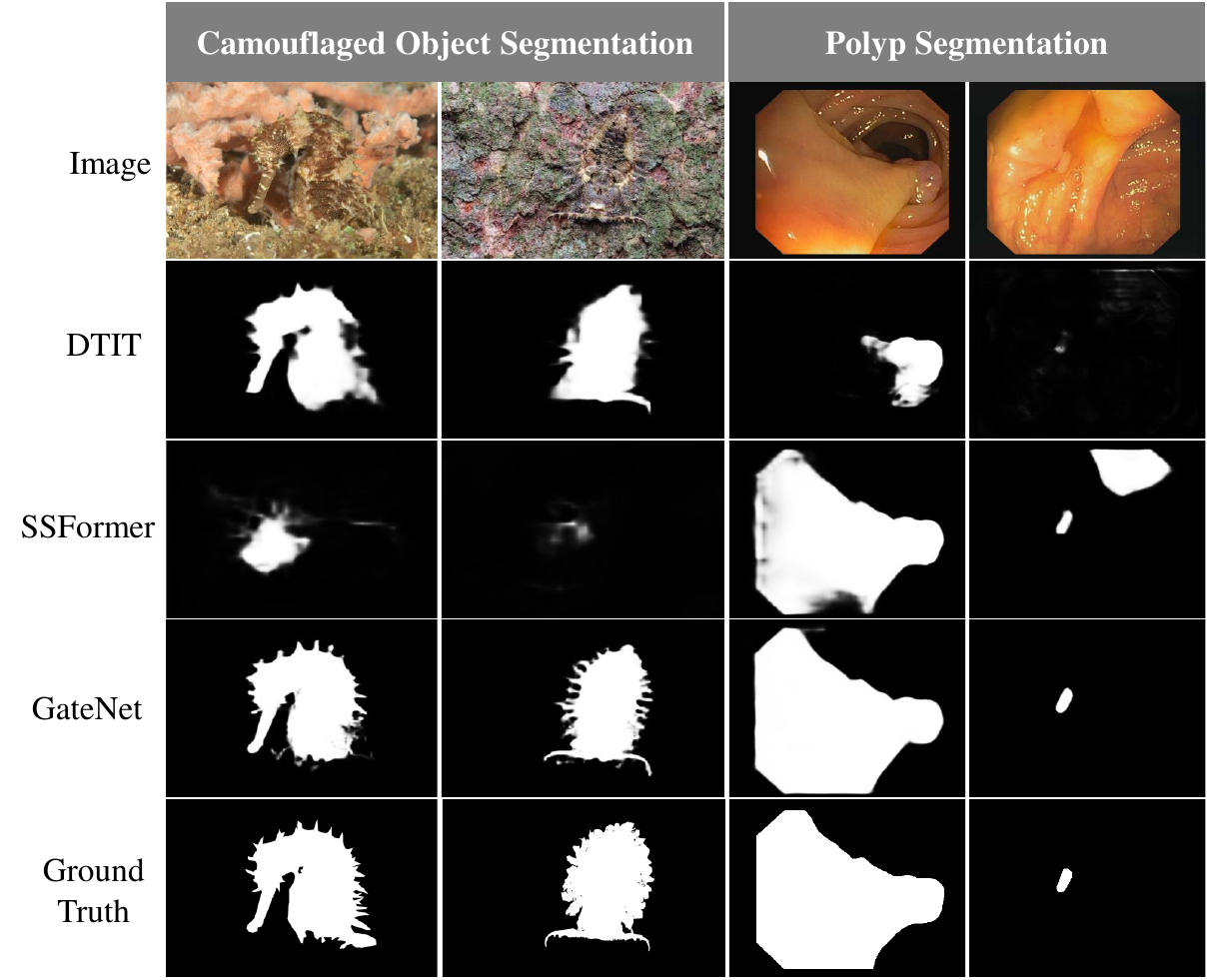}
	\caption{{Qualitative evaluation of DTIT~\cite{DTIT}, SSFormer~\cite{SSFormer} and GateNet. For DTIT and SSFormer, Polyp segmentation and camouflaged object segmentation are their cross-branch validations, respectively.} }
	\label{fig:cross-branch-visual-results}
\end{figure}
\\
\noindent\textbullet~\textbf{{Advantages in Cross-branch Prediction.}} To investigate the performance of transformer-based methods on cross-branch prediction, 
we separately select two representative transformer-based methods~\cite{SSFormer,DTIT} from camouflaged object segmentation and polyp segmentation field for cross-branch training and then conduct quantitative and qualitative evaluation. As shown in Tab.~\ref{tab:transformer_crossbranch_comparison}, both two transformer-based methods perform poorly in cross-branch comparison. We summarize some instructive reasons as follows: \textbf{\uppercase\expandafter{\romannumeral1})} Motivation of Designs. The motivation of GateNet is to solve the generalized binary segmentation challenge. We propose the gate unit, fold-aspp and residual parallel branch for suppressing background inference, perceiving multi-scale objects and restoring edge details, respectively. As a result, GateNet can be well generalized to diverse binary segmentation tasks. However, SSFormer~\cite{SSFormer} and DTIT~\cite{DTIT} focus more on specific characteristics within the sub-branch and propose expert designs for polyp segmentation and camouflaged object segmentation, respectively. SSFormer~\cite{SSFormer} introduces the multi-stage pyramid transformer architecture and proposes the progressive locality decoder  to smooth and emphasise the local features in the transformer, thereby improving the detailed information processing ability of the neural network. Authors think that the morphology of polyps is variable, but the structure of polyp images is relatively simple. Therefore, SSFormer~\cite{SSFormer} focuses on capturing the morphology and local information but ignores the background inference, which makes it perform poorly in camouflaged object segmentation where background scenes are often complex, as shown in Fig.~\ref{fig:cross-branch-visual-results}.
\begin{table}[!t]
  \centering
  \caption{{Cross-branch validation using two transformer-based methods on camouflaged object dataset  COD10K~\cite{SINet_COD} and polyp dataset CVC-ClinicDB~\cite{CVC-ClinicDB}.} }
  \begin{threeparttable}
   \resizebox{\linewidth}{!}{
    \setlength\tabcolsep{5pt}
    \renewcommand\arraystretch{1.05}
   \begin{tabular}{|r||ccc|cc|}
     \hline\thickhline
     \rowcolor{mygray}
    {Method}& \multicolumn{3}{|c|}{COD10K}&  \multicolumn{2}{|c|}{CVC-ClinicDB}  
    \\
      \rowcolor{mygray}
&$F_{\beta}^{\omega}\uparrow$ & $S_m\uparrow$ &$E_m\uparrow$ &$IOU\uparrow$ &$Dice\uparrow$
   \\
   
     		\hline
				\hline
SSFormer~\cite{SSFormer}&0.635&0.773&0.848&0.876&0.927 \\
DTIT~\cite{DTIT}&0.695&0.824&0.896&0.749&0.803 \\
GateNet&0.742&0.846&0.901&0.902&0.943\\
\hline
    \end{tabular}

   }
  \end{threeparttable}
 	\label{tab:transformer_crossbranch_comparison}
\end{table}
DTIT~\cite{DTIT} is bio-inspired by the discovery of camouflaged objects, in which the boundary feature is considered as query to improve the object detection and the object feature is taken as query to improve the boundary detection. The object and boundary detection are fully interacted by multi-head self-attention. However, DTIT~\cite{DTIT} ignores the scale varying in different objects and may produce failure prediction for tiny or large objects, as shown in Fig.~\ref{fig:cross-branch-visual-results}. \textbf{\uppercase\expandafter{\romannumeral2})} Model Complexity. Transformer-based models usually have high complexity and require a large amount of data and computational resources for training. If they are not adequately trained or the training data is not sufficient, it may lead to performance degradation when conducting cross-branch prediction. \textbf{\uppercase\expandafter{\romannumeral3})} Hyperparameter Selection. Many works~\cite{transformer_training_1,transformer_training_2,transformer_training_3} show that the transformer-based methods are very sensitive for the learning rate and optimizer settings during the training phase. Improper choice of hyperparameters may degrade the performance of the cross-branch model. Different from the transformer-based approaches, GateNet consistently achieves good results with uniform training settings for all tasks, including image size, enhancement techniques, optimizer parameters, learning rate, and the numbers of epoch.
\\
\begin{figure}[!t]
	\includegraphics[width=\linewidth]{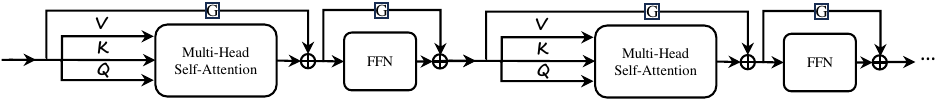}\\ 
	\centering
	\caption{{Illustration of the simple gated transformer.}} 		
	\label{fig:transformer_gate}
\end{figure} 
\begin{table}[!t]
  \centering
  \caption{{Qualitative evaluation of applying simple gated transformer and multi-level gate units to existing transformer-based COD methods on the COD10K test set.}}
  \begin{threeparttable}
   \resizebox{\linewidth}{!}{
    \setlength\tabcolsep{5pt}
    \renewcommand\arraystretch{1.05}
   \begin{tabular}{|r|c|c||c|c|c|}
     \hline\thickhline
     \rowcolor{mygray}
Method  &Publication&Backbone&$F_{\beta}^{\omega}\uparrow$ & $S_m\uparrow$ &$E_m\uparrow$ 
   \\
     		\hline
				\hline
DTIT~\cite{DTIT}&ICPR 2022&MiT-B5&0.695&0.824&0.896 \\
DTIT-Gate &ICPR 2022&MiT-B5&0.721&0.847&0.928 \\
FSPNet~\cite{FSPNet_COD}&CVPR 2023&ViT-B&0.735&0.851&0.930 \\
FSPNet-Gate&CVPR 2023&ViT-B&0.764&0.869&0.938 \\
\hline
    \end{tabular}

   }
  \end{threeparttable}
 	\label{tab:transformer_gains}
\end{table}
\noindent\textbullet~\textbf{{Gated Mechanism Powers Transformer.}} We explore the potential of improving the transformer-based methods by incorporating gate design.  The vanilla transformer uses all pass skip connection to fuse the original input features and the output features through multi-head self-attention (MHSA) and FFN. With the help of MHSA and FFN, query, key and value can generate task-specific strong attention features. However, all pass skip connection may introduce incompatible interference information and reduce the performance of the transformer. To this end, we naturally apply our gate unit to the vanilla transformer. We integrate the initial features and the output features through MHSA/FFN to generate gate values, which can adaptively control the information transition from skip connection. The internal  structure of the simple gated transformer is shown in Fig.~\ref{fig:transformer_gate}. 
We replace the vanilla transformer in the transformer-based methods~\cite{DTIT,FSPNet_COD} with the gated transformer and multi-level gate units to evaluate the effectiveness of our designs. In Tab.~\ref{tab:transformer_gains}, we can see that the gated versions consistently surpass the corresponding vanilla transformer versions. 
}

\section{Discussion}\label{sec:Discussion}

\begin{figure}[!t]
	\centering
	\includegraphics[width=\linewidth]{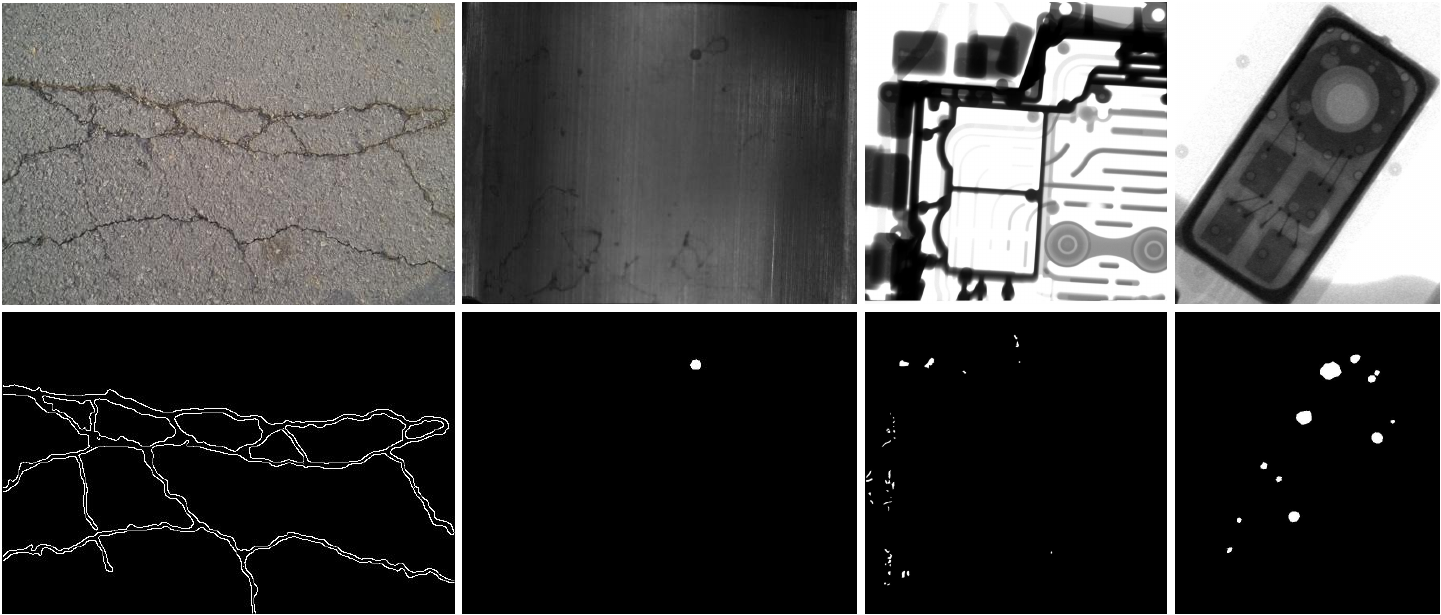}\\
	\caption{Some examples of surface defect detection (\textit{e.g.}, crack, magnetic tiles,  car parts, electronic components).  }
	\label{fig:potential_applications}
\end{figure}

In this section, we further provide deeper theoretical explanation of the multi-level gated mechanism and give some potential applications: %
\\
\noindent$\bullet$ {Prototype I: In cognitive science, inhibitory neurons, also known as interneurons, play a crucial inhibitory role in the nervous system. Inhibitory neurons play a crucial inhibitory role by balancing excitation and inhibition, improving signal quality, participating in cognitive and emotional processes, and protecting the nervous system. 
Firstly, they balance excitation and inhibition by suppressing the activity of other neurons, which is important for maintaining normal nervous system function. Secondly, they improve signal quality by reducing neuronal noise and interference, increasing the signal-to-noise ratio and making the signal clearer and more reliable, thereby enhancing the brain's information processing capabilities. In addition, they participate in various cognitive and emotional processes including working memory, long-term memory, learning, attention, and emotional regulation by regulating the excitability and inhibitory nature of neurons. Lastly, they protect the nervous system from the harm of excessive excitation or inhibition, thereby avoiding the occurrence of some neurological diseases. 
} 
\\
\noindent$\bullet$ {Prototype II: In circuit electronics analysis, a gated circuit can control the on/off state of the output signal based on the voltage of the input signal. 
For a combinational circuit with $n$ outputs, we only need add $n-1$ gates without other additional designs. In addition, gated circuits are less susceptible to external interference, which can ensure the stability and reliability of the circuit. Due to their fast state transitions, gated circuits are well suited for applications that require high-speed digital control. In terms of achieving adaptive regulation and balance of circuit output, multi-level gated unit circuits combined with feedback control mechanisms play a crucial role. 
For example, resistors and capacitors can be used to adjust the impedance and phase of the circuit to balance different sections. Furthermore, feedback control and adaptive control methods can be employed to dynamically adjust the work state of the circuit to achieve adaptive balance.} 
\begin{table}[!t]
  \centering
  \caption{{Quantitative comparison of different semantic segmentation methods on the Cityscapes~\cite{Cityscapes} val set. GateNet-JT and GateNet-ST refer to models trained separately or jointly for each category.  The best scores are highlighted in {\color{reda}{\textbf{red}}}.}}
  \begin{threeparttable}
   \resizebox{\linewidth}{!}{
    \setlength\tabcolsep{5pt}
    \renewcommand\arraystretch{1.05}
   \begin{tabular}{|r|c|c||c|}
     \hline\thickhline
     \rowcolor{mygray}
Method  & Publication & Backbone&mIoU $\uparrow$
     \\
     		\hline
				\hline
FCN~\cite{FCN}&CVPR 2015&ResNet-101&	76.6\\
EncNet~\cite{EncNet}&CVPR 2018&ResNet-101&	76.9\\
PSPNet~\cite{PSPNet}&CVPR 2017&ResNet-101&	78.5\\
CCNet~\cite{CCNet}&ICCV 2019&ResNet-101&	80.2\\
DeeplabV3+~\cite{Deeplabv3+}&ECCV 2018&ResNet-101&	80.9\\
SETR~\cite{SETR}&CVPR 2021&ViT-Large&	82.2\\
SegFormer~\cite{SegFormer}&NeurIPS 2021&MiT-B5&	84.0\\
Mask2Former~\cite{Mask2Former}&CVPR 2022&Swin-Large&	84.3\\
\hline
GateNetv2-JT&-&ResNet-101&	78.3\\
GateNetv2-JT&-&Swin-Large&	80.4\\
GateNetv2-JT&-&ConNext-Large&	80.6\\
GateNetv2-ST&-&ResNet-101&	82.8\\
GateNetv2-ST&-&Swin-Large&	84.9\\
GateNetv2-ST&-&ConNext-Large&	\color{reda} \textbf{85.2}\\
\hline
    \end{tabular}
   }
  \end{threeparttable}
 	\label{tab:Semantic_segmentation}
\end{table}
\\
\noindent$\bullet$ {Modeling guidance: Prototype I provides the basic principles of biological neural networks for introducing gated  mechanisms into artificial neural networks. 
Prototype II provides our GateNet with modular functional guidance.
On the one hand, our gate units suppress both channel-wise and spatial feature response. In this way, the network actually learns adaptive thresholding. The area, in which feature values are below this threshold, has a lower response in the prediction, while the feature values above this threshold correspond to the task-specific activation area. It helps the decoder to gradually filter out the region with strong feature response. Our gate unit achieves the same function as the gated circuit in controlling the on/off state of the output signal based on the voltage of the input signal.  
On the other hand, our GateNet only inserts several gate units between encoder blocks and decoder blocks of the FPN baseline. It has the same convenience of design as the gated circuit. Finally, the backward propagation in the neural networks has the same function as the feedback controlling mechanism in circuit electronics. Therefore, our GateNet has the same adaptive balance function as the gated circuit.}
\\
\noindent$\bullet$ Limitations: The proposed gated mechanism is unsuitable for the multi-class semantic segmentation task. Because this task needs to treat all pixels of the whole image equally importantly and all categories have the same importance, the information suppression design is out of place. {We apply GateNet separately for semantic segmentation of each class on the popular Cityscapes dataset~\cite{Cityscapes}. In Tab.~\ref{tab:Semantic_segmentation}, we can see that GateNet-ST outperforms other models, but GateNet-JT performs poorly. Therefore, the proposed gated mechanism has advantages in binary segmentation focusing on a single class, rather than the semantic segmentation task that require balancing multiple classes. This also means the GateNet has wider applicability to binary segmentation problems. }
\\
\noindent$\bullet$ Application: In this paper, we have given detailed experimental analyses in ten popular binary segmentation tasks. Besides, GateNet has potential application in the field of industry with complex scenes as shown in Fig~\ref{fig:potential_applications}. We hope that this study can provide deep insights into the underlying design for more binary segmentation tasks and spark novel ideas.   

\section{Conclusions}
    As far as we know, this is the first work to comprehensively review recent progress in binary segmentation, which summarizes more than $140$ fully supervised models according to task settings, technique contributions, and learning strategies. 
    To unify all the sub-branches and establish a fair model benchmark to promote the prosperous development of the binary segmentation field, we propose a novel yet general gated network architecture. 
	We first adopt multi-level gate units to balance the contribution of each encoder block and suppress the activation of the features of non-task-aware regions, which can provide useful context information for the decoder while minimizing interference. 
	We quantitatively reveal the role played by features at all levels of the encoder for different segmentation tasks, which provides a new perspective on the interpretability of deep learning.
	Next, we use the Fold-ASPP to gather multi-scale semantic information for the decoder. By the folded operation, the atrous convolution achieves a local-in-local effect, which not only expands the receptive field  but also retains the correlation among local sampling points.
	Finally, to further supplement the details, we combine all encoder features in parallel and construct a residual structure. 
	Experimental results on $33$ benchmark datasets towards $10$ binary segmentation tasks demonstrate that the proposed model outperforms $42$ state-of-the-art methods under $10$ evaluation metrics.

{
\small
\bibliographystyle{plain}
\bibliography{sn-bibliography}
}

\begin{appendices}
\section{Qualitative Evaluation}\label{secA1}
\begin{figure*}[!t]
  	\centering
  	\includegraphics[width=0.98\linewidth]{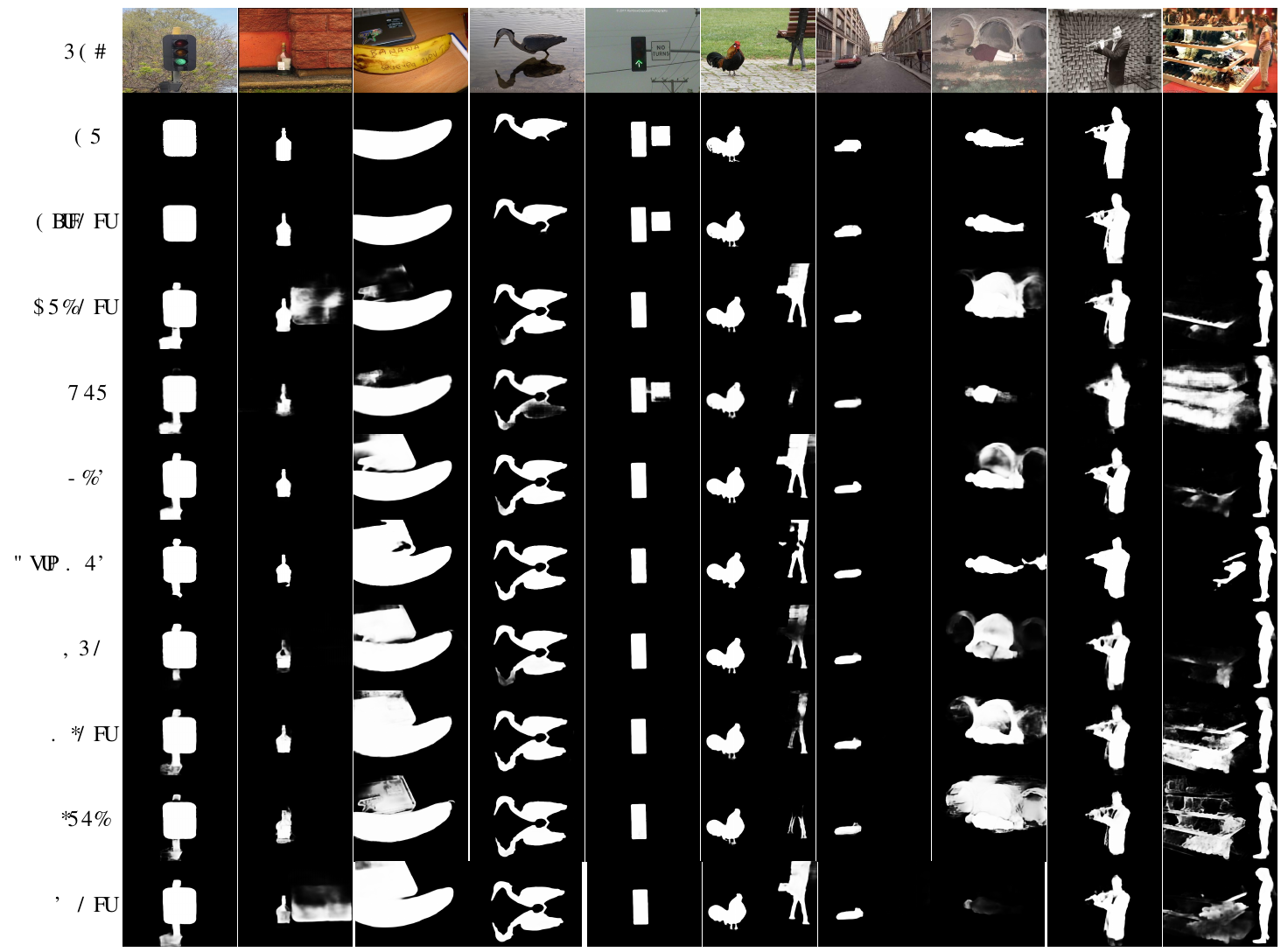}
  	\caption{Visual comparison between our GateNet results and the state-of-the-art methods (CTDNet~\cite{CTDNet}, VST~\cite{VST}, LDF~\cite{LDF}, Auto-MSF~\cite{Auto-MSFNet}, KRN~\cite{KRN}, MINet~\cite{MINet}, ITSD~\cite{ITSD}, F3Net~\cite{F3Net}) on \textbf{RGB SOD} datasets. }
  	\label{fig:RGB_SOD}
  \end{figure*}

\begin{figure*}[!t]
  	\centering
  	\includegraphics[width=0.98\linewidth]{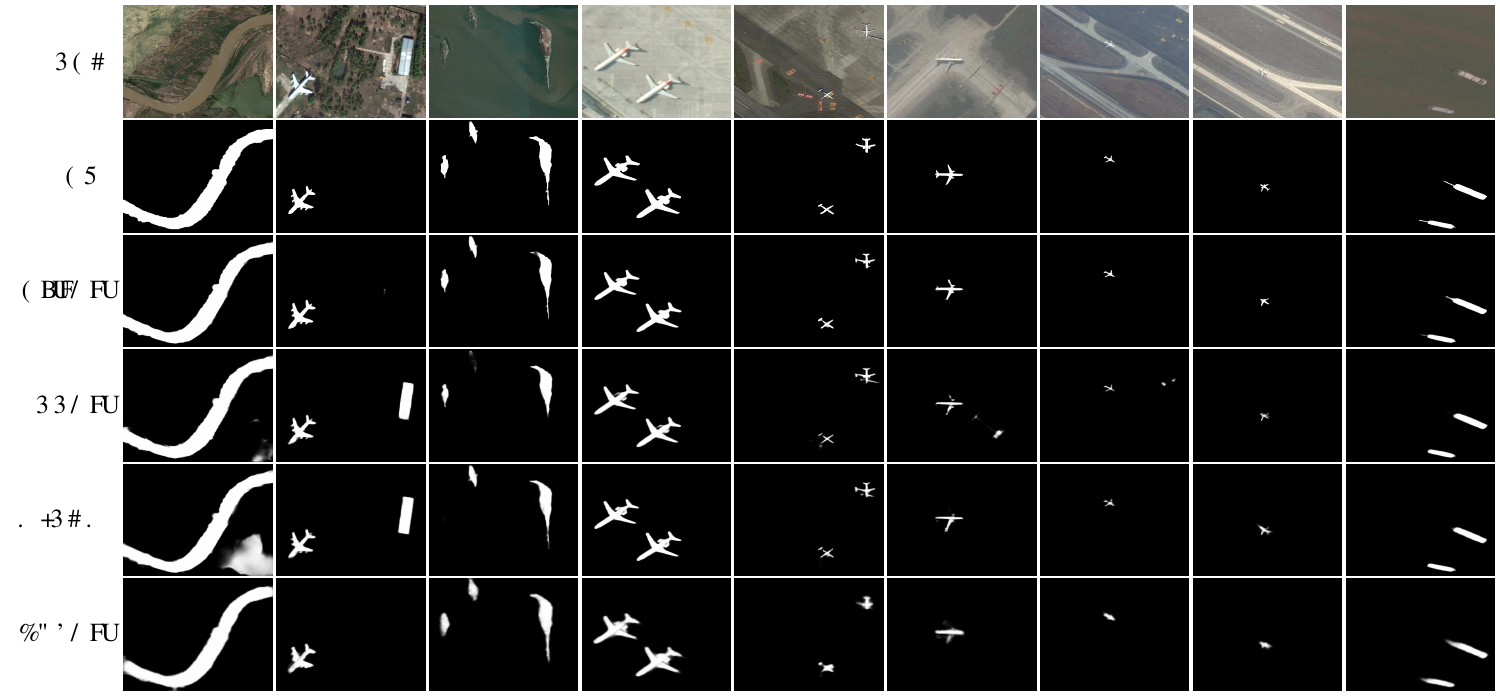}
  	\caption{Visual comparison between our GateNet results and the state-of-the-art methods (RRNet~\cite{RRNet_RSISOD}, MJRBM~\cite{MJRBM_RSISOD}, DAFNet~\cite{DAFNet_RSISOD}) on \textbf{ORSI SOD} datasets. 
  	}
  	\label{fig:eorssd}
  \end{figure*}

\begin{figure*}[!t]
  	\centering
  	\includegraphics[width=\linewidth]{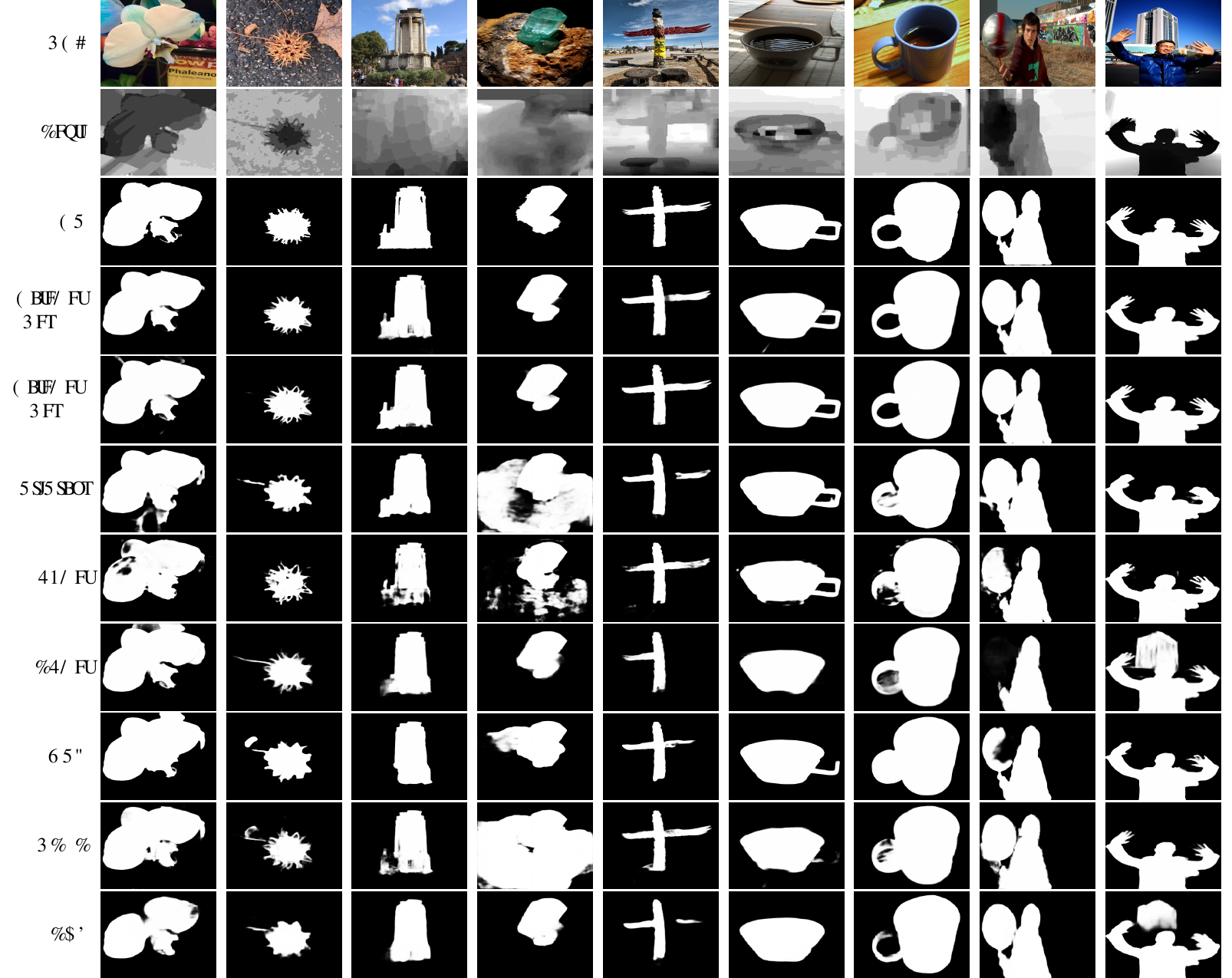}
  	\caption{Visual comparison between our GateNet results and the state-of-the-art methods (TriTransNet~\cite{TriTransNet_RGBDSOD}, SPNet~\cite{SPNet_RGBDSOD}, DSNet~\cite{DSNet_RGBDSOD}, UTA~\cite{UTA_RGBDSOD}, RD3D~\cite{RD3D_RGBDSOD}, DCF~\cite{DCF_RGBDSOD}) on \textbf{RGB-D SOD} datasets.
  	}
  	\label{fig:rgbd_sod}
  \end{figure*}

\begin{figure*}[!t]
  	\centering
  	\includegraphics[width=\linewidth]{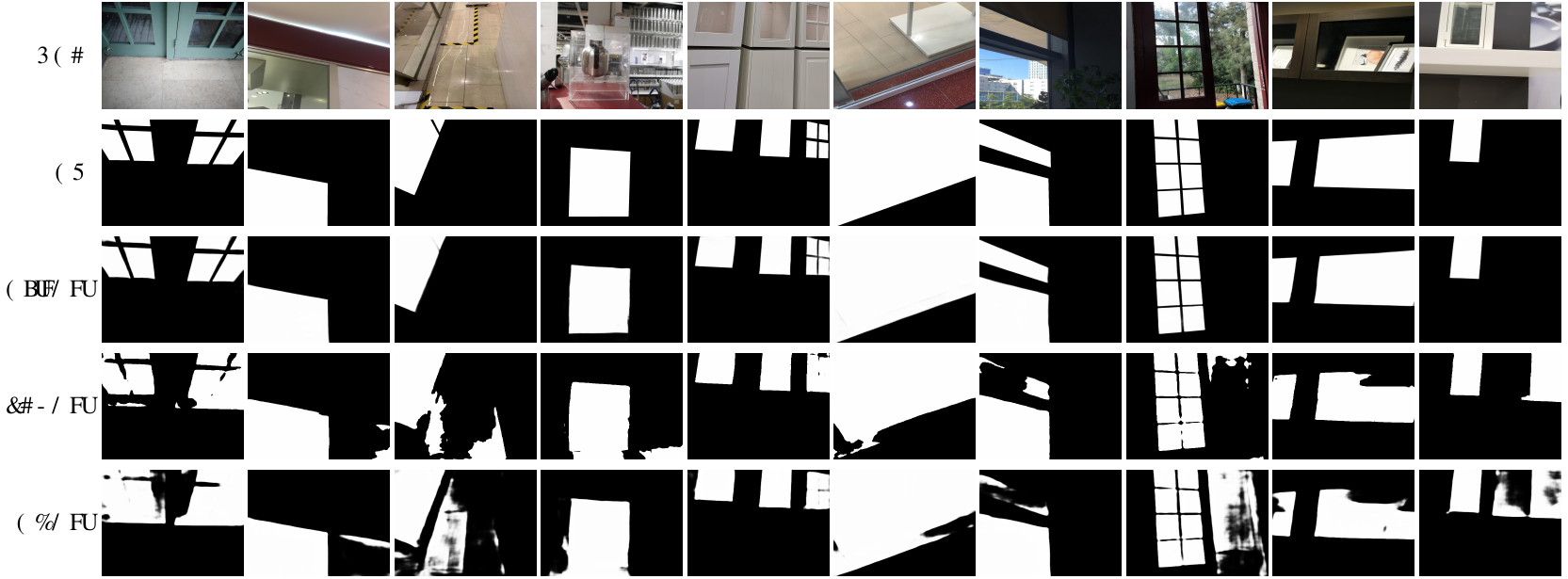}
  	\caption{Visual comparison between our GateNet results and the state-of-the-art methods (EBLNet~\cite{EBLNet_Glass}, GDNet~\cite{GDNet_Glass}) on \textbf{Glass Object Detection} datasets. 
  	}
  	\label{fig:glass}
  \end{figure*}

\begin{figure*}[!t]
  	\centering
  	\includegraphics[width=0.98\linewidth]{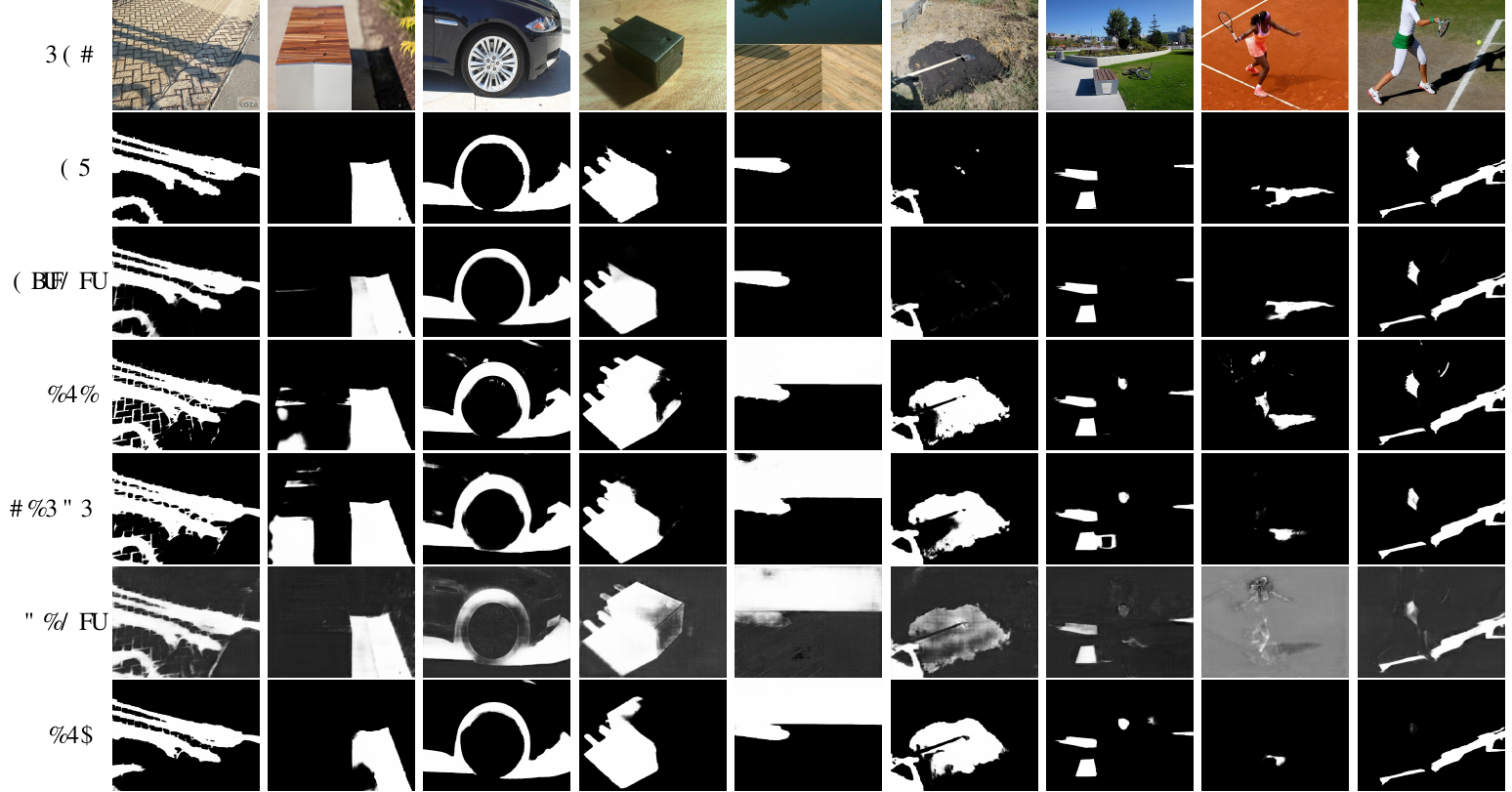}
  	\caption{Visual comparison between our GateNet results and the state-of-the-art methods (DSD~\cite{DSDNet_Shadow}, BDRAR~\cite{BDRAR_Shadow}, ADNet~\cite{ADNet_Shadow}, DSC~\cite{DSC_Shadow}) on \textbf{Shadow Detection} datasets. 
  	}
  	\label{fig:shadow}
  \end{figure*}

\begin{figure*}[!t]
  	\centering
  	\includegraphics[width=0.98\linewidth]{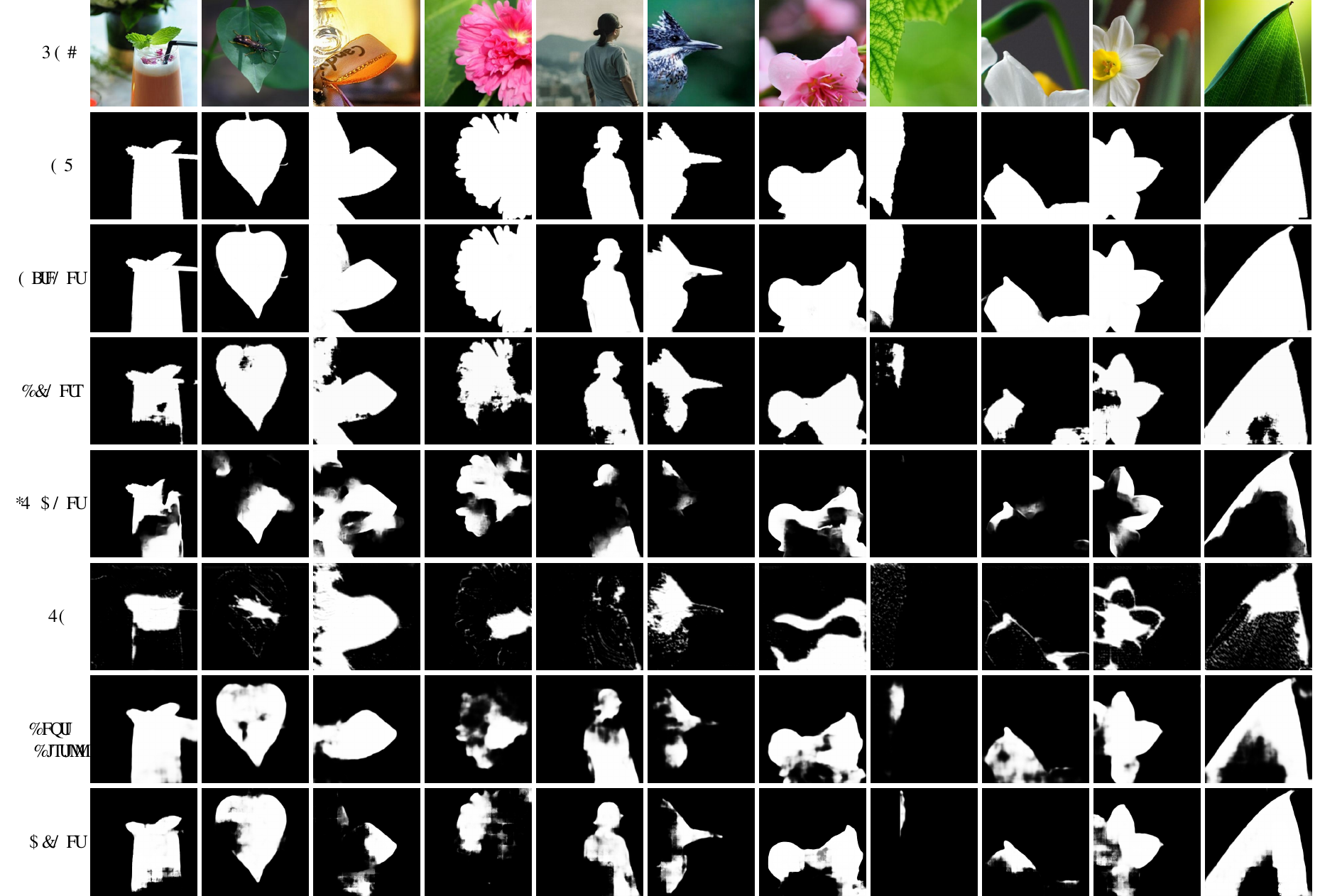}
  	\caption{Visual comparison between our GateNet results and the state-of-the-art methods (DENets~\cite{DENets_DBD}, IS2CNet~\cite{IS2CNet_DBD}, SG~\cite{SG_DBD}, Depth-Distill~\cite{Depth-Distill_DBD}, CENet~\cite{CENet_DBD}) on \textbf{Defocus Blur Detection} datasets. 
  	}
  	\label{fig:DBD}
  \end{figure*}

\begin{figure*}[!t]
  	\centering
  	\includegraphics[width=0.95\linewidth]{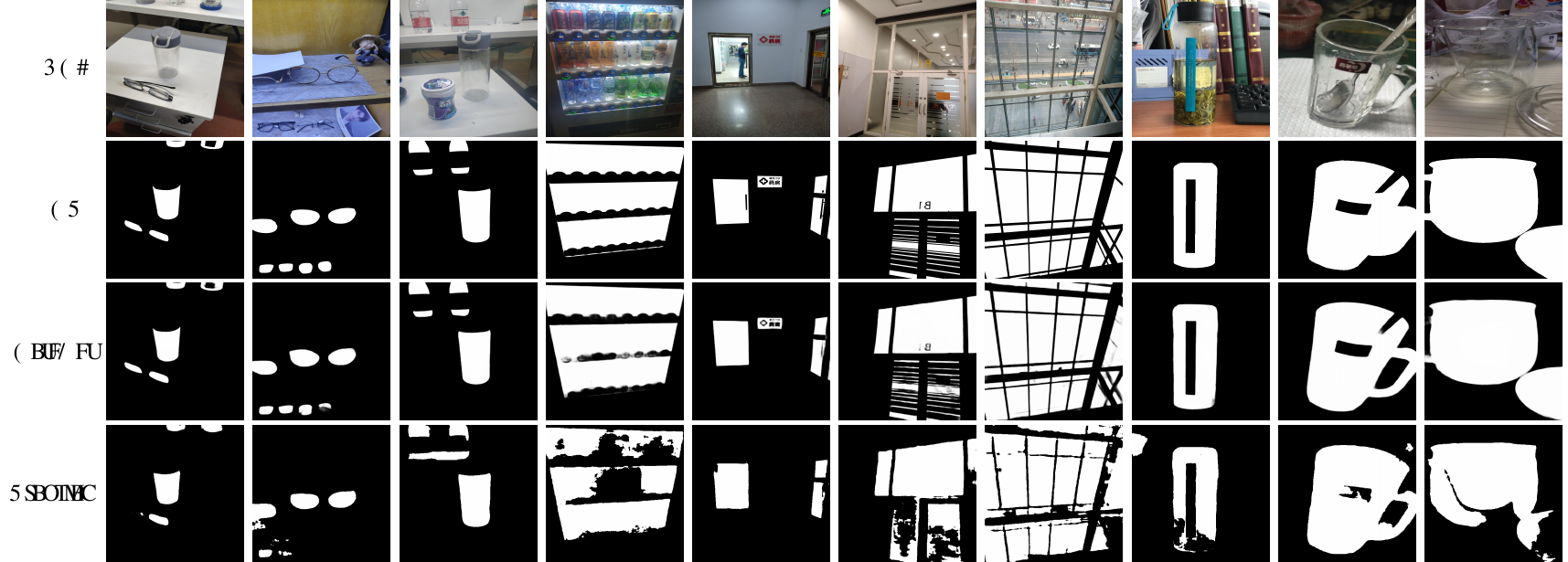}
  	\caption{Visual comparison between our GateNet results and the state-of-the-art method (Translab~\cite{TransLab_Transparent}) on \textbf{Transparent Object Detection} datasets. 
  	}
  	\label{fig:transparent}
  \end{figure*}

\begin{figure*}[!t]
  	\centering
  	\includegraphics[width=\linewidth]{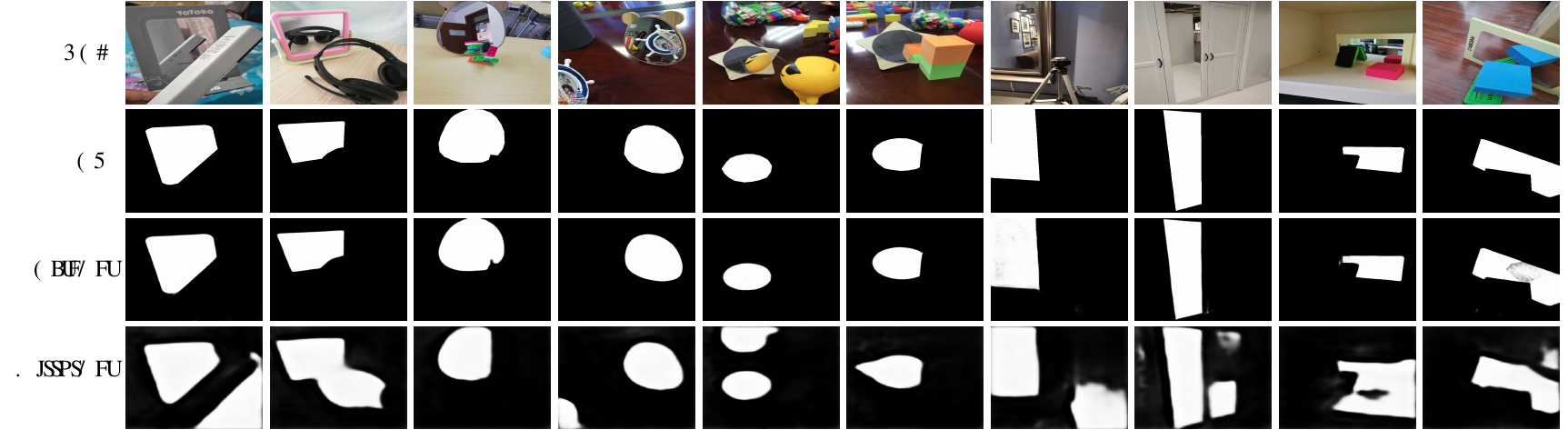}
  	\caption{Visual comparison between our GateNet results and the state-of-the-art method (MirrorNet~\cite{MirrorNet_Mirror}) on \textbf{Mirror Detection} datasets. 
  	}
  	\label{fig:mirror}
  \end{figure*}

\begin{figure*}[!t]
  	\centering
  	\includegraphics[width=\linewidth]{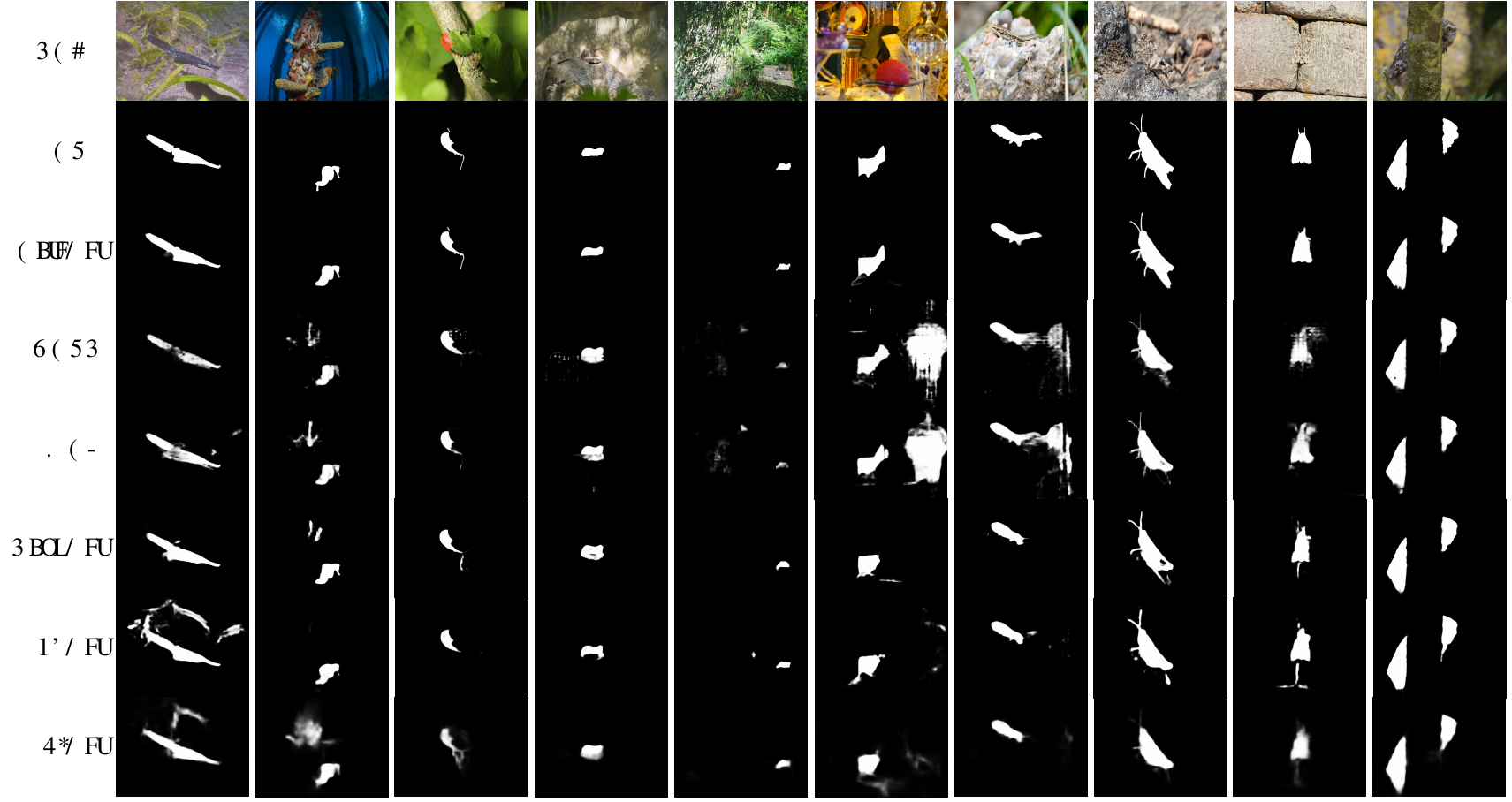}
  	\caption{Visual comparison between our GateNet results and the state-of-the-art methods (UGTR~\cite{UGTR_COD}, IS2CNet~\cite{MGL_COD}, RankNet~\cite{Rank-Net_COD}, PFNet~\cite{PFNet_COD}, SINet~\cite{SINet_COD}) on \textbf{Camouflaged Object Detection } datasets. 
  	}
  	\label{fig:cod}
  \end{figure*}

\begin{figure*}[!t]
  	\centering
  	\includegraphics[width=\linewidth]{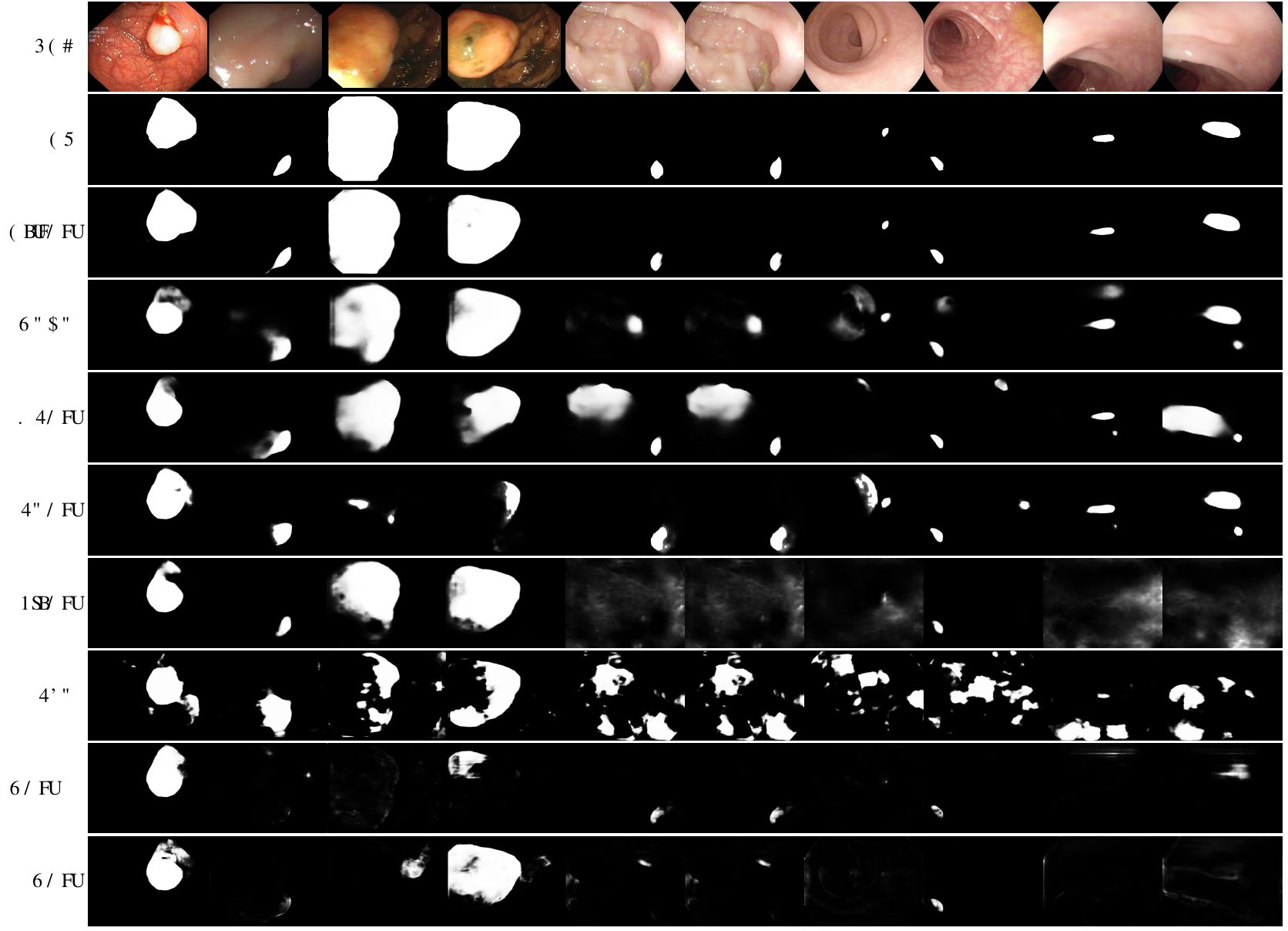}
  	\caption{Visual comparison between our GateNet results and the state-of-the-art methods (UACA~\cite{UACANet_Polyp}, MSNet~\cite{MSNet_Polyp}, SANet~\cite{SANet_Polyp}, PraNet~\cite{PraNet_Polyp}, SFA~\cite{SFA_Polyp}, UNet++~\cite{UNet++}, UNet~\cite{Unet}) on \textbf{Polyp Segmentation} datasets. 
  	}
  	\label{fig:polyp}
  \end{figure*}

Fig.~\ref{fig:RGB_SOD} - Fig.~\ref{fig:polyp} illustrate some visual comparisons on each sub-task. We summarize the advantages of the GateNet compared to others when facing some challenges:
\textbf{\uppercase\expandafter{\romannumeral1})} 
\textbf{Interference produced by complex.}
In camouflaged object detection and poly segmentation tasks,  foreground  objects usually share the similar appearance to the  background, which can easily deceive predictors.  But the GateNet can accurately capture the hidden objects and separate them from the surrounding environment (see the Fig.~\ref{fig:cod} and Fig.~\ref{fig:polyp}). The gated mechanism also plays an important role in RGB -D salient object detection. As shown in Fig.~\ref{fig:rgbd_sod}, the proposed two-stream GateNet can effectively utilize the guidance information provided by the high-quality depth map while suppressing the interference information from the low-quality depth map, thereby identifying the whole object precisely. 
\textbf{\uppercase\expandafter{\romannumeral2})} 
\textbf{Interference produced by adjacent objects.} In the real world, shadows often exist on the ground or desktop, and are closely adjacent to the original object. This characteristic requires shadow detection networks to have the ability to distinguish between adjacent objects. As shown in Fig.~\ref{fig:shadow}, most methods are disturbed by the surface or the original object, but our method can focus on the shadow regions.
\textbf{\uppercase\expandafter{\romannumeral3})}
\textbf{The foreground exists multiple or small objects.} On the one hand, glass-like objects are often present in groups in the real world, which poses a serious challenge to the perception capability of the network for the multiple objects. On the other hand, small objects usually appear in remote sensing images. Benefiting from the Fold-ASPP, both multiple and small objects can be localized accurately. Fig.~\ref{fig:glass} and Fig.~\ref{fig:transparent} show that our method can accurately distinguish each independent connected region without sticking to each other. GateNet is the only one can provide clean prediction maps and maintain the basic shape of the aircraft (see the $6^{th}$ - $8^{th}$ columns in Fig.~\ref{fig:eorssd}).
\textbf{\uppercase\expandafter{\romannumeral4})}
\textbf{Boundary and details.} Our GateNet has a mix feature aggregation decoder that a parallel branch by concatenating the output of the progressive branch and the features of the gated encoder, so that the residual information complementary to the progressive branch is supplemented to generate the final prediction. In this way,  the prediction can restore more details, therefore, the limbs and even tentacles of the insects are retained well (see the $3^{th}$ and $8^{th}$ columns in Fig.~\ref{fig:cod}).  
\textbf{\uppercase\expandafter{\romannumeral5})}
\textbf{Regional consistency.} In defocus blur detection task, the focused area usually has incomplete semantic information because the blurred region may also belong to the semantic part of the foreground.  Benefiting from the folded operation, our model can obtain more stable structural features to improve the intra-class consistency. 
From the results in Fig.~\ref{fig:DBD}, it can be observed that our method can segment the foreground well while the other methods more or less lose similar areas inside or around focused regions.

  
\end{appendices}



\end{document}